# Enhancing Academic Advisement with Case-Based Reasoning

Tanat Tonguthaisri

**RMIT University**
School of Electrical and Computer Engineering; institution at the time of the research (2003)

## Abstract

This research involved the study of an application of Case-Based Reasoning (CBR) in the field of academic advising. Academic advisement has been recognised as an important function of higher education. However, there is a widespread desire to improve the advisement services provided by many educational institutions (e.g., colleges and universities) around the world. In the last ten years, CBR expert systems have made important contributions in many domains by applying CBR methodology to a variety of areas of human problem solving. However, to date, expert advisement systems have not been successfully applied to academic advisement, which seems to be a very appropriate domain for this technology.

Academic advising is an important and time consuming task, and different tools and techniques must be used to make it an effective and efficient process. This thesis describes the design, development, and validation of a prototype case-based academic advisement recommender system that supplements the conventional advising process, and could eventually evolve to replace human academic advisors when its full potential is realised.

## AI disclosure - 2026 digital edition

AI tools were used in preparing the 2026 digital edition for transcription, proofreading, formatting and figure reconstruction. The research reported in the thesis dates from 2003. This digital preparation does not constitute a new validation of the historical findings; source uncertainties are documented in the separate correction log and unresolved source errors register.

# Enhancing Academic Advisement with Case-Based Reasoning

*Tanat Tonguthaisri*

## About this edition

This edition makes the supplied digitised thesis available as editable text with links back to its source pages. The research text remains the historical work. Contemporary explanations and research commentary belong to separately labelled companion material.

The PDF supplied by the author contains 201 pages, including binding and library pages. The source-preserving reconstructed PDF retains that complete sequence. Its single redacted figure has been replaced with the author-supplied 2026 redraw, which includes Watson’s extension and is not a facsimile of the original figure.

The text on all 201 supplied source pages has been compared page by page against the original images. Verified transcription and structure corrections are incorporated here. Original wording and source errors are retained. The separate correction log and unresolved source errors register describe editorial changes and remaining uncertainties. This comparison does not validate the historical claims, calculations, citations, or executability of the PHP listing.

Open the RMIT repository record

## Reading navigation

# Original front matter

## Original source PDF page 1

Enhancing Academic

Advisement with

Case-Based Reasoning

Tanat Tonguthaisri

Master of Engineering

2003

RMIT

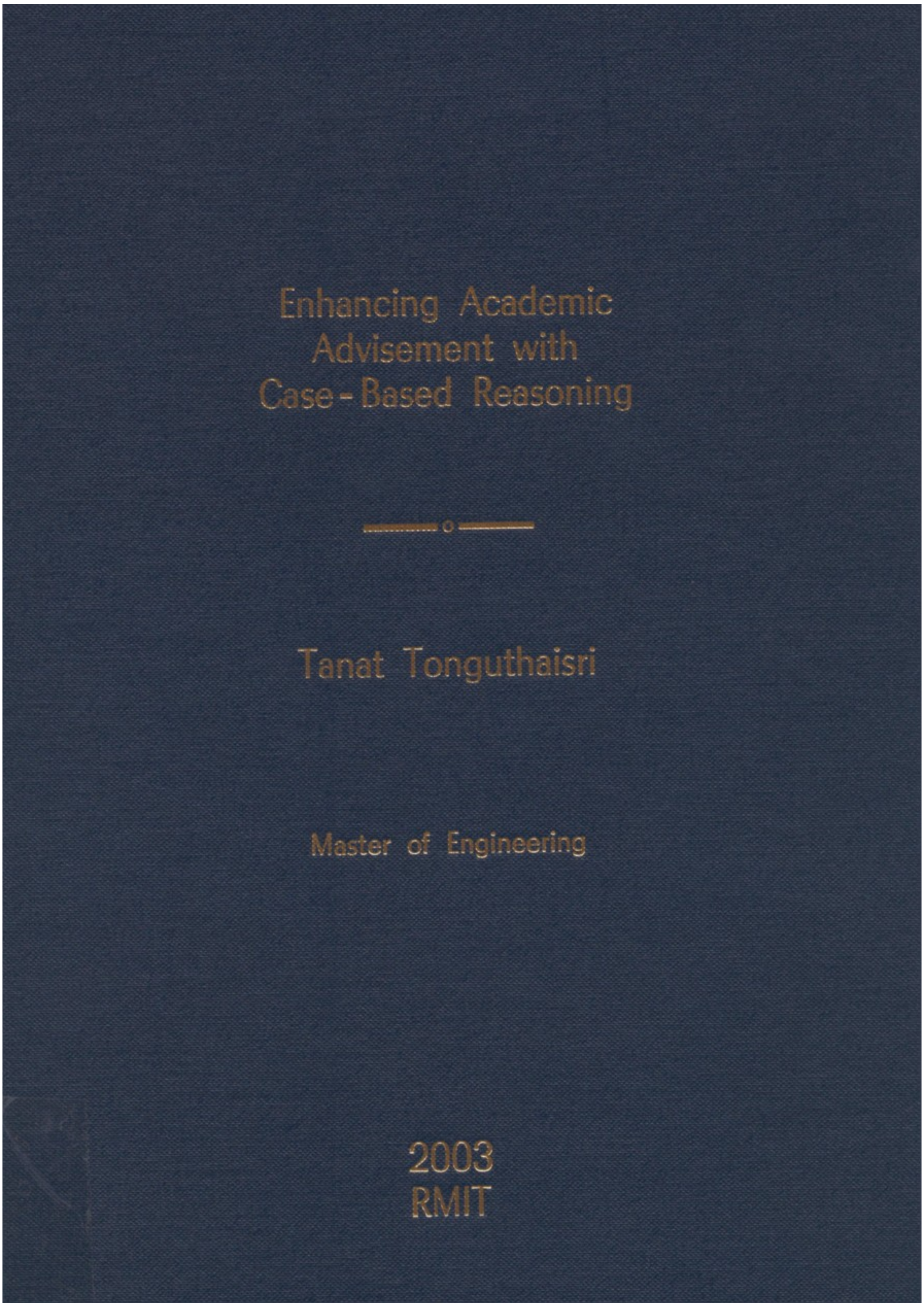

**Original source visual, PDF page 1**

## Original source PDF page 2

**ROYAL MELBOURNE INSTITUTE OF TECHNOLOGY**

**USE OF THESIS**

written work containing passages copied or closely paraphrased from this thesis, which passages would in total constitute an infringing copy for the purpose of the Copyright Act, he or she must obtain the written permission of the author to do so.

You are required to fill out the acknowledgement below that a volume/copy has been supplied to you on your undertaking to observe the above conditions.

NAME(in caps)

DATE

SIGNATURE

Special Collections Room

ROYAL MELBOURNE INSTITUTE OF TECHNOLOGY

## USE OF THESIS

This volume is the property of Royal Melbourne Institute of Technology, but the rights of the author must be respected. If any passage is quoted or closely paraphrased in a paper or written work prepared by the user, the source of the passage must be acknowledged in the work. If the user desires to publish a paper or written work containing passages copied or closely paraphrased from this thesis, which passages would in total constitute an infringing copy for the purpose of the Copyright Act, he or she must obtain the written permission of the author to do so.

You are required to fill out the acknowledgement below that a volume/copy has been supplied to you on your undertaking to observe the above conditions.

| NAME(in caps) | DATE | SIGNATURE |
| --- | --- | --- |
| | | |
| | | |
| | | |
| | | |
| | | |
| | | |
| | | |
| | | |
| | | |
| | | |
| | | |
| | | |
| | | |
| | | |
| | | |
| | | |
| | | |
| | | |
| | | |

Special Collections Room

**Original source visual, PDF page 2**

## Original source PDF page 3

# Enhancing Academic Advisement

# with Case-Based Reasoning

A thesis submitted in fulfilment of the requirements for the degree of

Master of Engineering

TANAT TONGUTHAISRI

B.Eng., B.AppSci.

School of Electrical and Computer Engineering

Faculty of Engineering

RMIT University

August 2003

# Enhancing Academic Advisement with Case-Based Reasoning

A thesis submitted in fulfilment of the requirements for the degree of

**Master of Engineering**

TANAT TONGUTHAISRI

B.Eng., B.AppSci.

School of Electrical and Computer Engineering
Faculty of Engineering
RMIT University

August 2003

**Original source visual, PDF page 3**

## Original source PDF page 4

**Original source visual, PDF page 4**

## Original source PDF page 5

### Declaration

I certify that except where due acknowledgement has been made, the work is that of the author alone; the work has not been submitted previously, in whole or in part, to qualify for any other academic award; the content of the thesis is the result of work which has been carried out since the official commencement date of the approved research program; and, any editorial work, paid or unpaid, carried out by a third party is acknowledged.

Tanat Tonguthaisri

August 22nd, 2003

# Declaration

I certify that except where due acknowledgement has been made, the work is that of the author alone; the work has not been submitted previously, in whole or in part, to qualify for any other academic award; the content of the thesis is the result of work which has been carried out since the official commencement date of the approved research program; and, any editorial work, paid or unpaid, carried out by a third party is acknowledged.

Tanat Tonguthaisri

August 22nd, 2003

**Original source visual, PDF page 5**

## Original source PDF page 6

### Acknowledgements

I would like to take this opportunity to express my utmost gratitude towards my senior supervisor, Roy Ferguson, for his insight, advice, guidance, support, understanding and patience, which has been given to me in numerous occasions during the course of this research. Without him, I would most probably be unable to complete this study and this thesis.

I would also like to acknowledge the very timely, professional, and high quality editorial work, provided by Chris Vedelago of WordsRU.

# Acknowledgements

I would like to take this opportunity to express my utmost gratitude towards my senior supervisor, Roy Ferguson, for his insight, advice, guidance, support, understanding and patience, which has been given to me in numerous occasions during the course of this research. Without him, I would most probably be unable to complete this study and this thesis.

I would also like to acknowledge the very timely, professional, and high quality editorial work, provided by Chris Vedelago of WordsRU.

**Original source visual, PDF page 6**

## Original source PDF page 7

### Contents

# Contents

iv

**Original source visual, PDF page 7**

## Original source PDF page 8

v

**Original source visual, PDF page 8**

## Original source PDF page 9

**Original source visual, PDF page 9**

## Original source PDF page 10

### Table of Tables and Figures

# Table of Tables and Figures

Original source visual, PDF page 10

# Abstract

**Original printed page 1 · source PDF page 11 · Open source page**

This research involved the study of an application of Case-Based Reasoning (CBR) in the field of academic advising. Academic advisement has been recognised as an important function of higher education. However, there is a widespread desire to improve the advisement services provided by many educational institutions (e.g., colleges and universities) around the world. In the last ten years, CBR expert systems have made important contributions in many domains by applying CBR methodology to a variety of areas of human problem solving. However, to date, expert advisement systems have not been successfully applied to academic advisement, which seems to be a very appropriate domain for this technology.

Academic advising is an important and time consuming task, and different tools and techniques must be used to make it an effective and efficient process. This thesis describes the design, development, and validation of a prototype case-based academic advisement recommender system that supplements the conventional advising process, and could eventually evolve to replace human academic advisors when its full potential is realised.

# Chapter 1 Introduction

**Original printed page 2 · source PDF page 12 · Open source page**

## 1.1 Background

In interacting with students, the academic advisor is on the “front line” of the university. It is a critical position, and the success or failure of a student’s education and growth is greatly influenced by the advising function.

The overall goal of academic advisement is to assist students in the development of meaningful educational plans that are compatible with their life goals. In essence, academic advisement is a fundamental part of higher education. The process helps students receive information, consider options, and make choices. Academic advisement assists students in achieving their educational potential by helping them understand themselves and learn to use the resources of the campus to meet their individual needs and aspirations.

Academic advisement should be viewed as a continuous process of clarification and evaluation. Academic advisement is a decision making process through which a student, aided by an advisor, effectively and efficiently plan their curricula according to the student’s goals.

It is important to remember, however, that the ultimate responsibility for making informed decisions about life goals and educational plans must reside with the student. The academic advisor helps to identify options for the student, each with its benefits and consequences. Therefore, advisors may also be called upon to help students discover what their goals are.

Ideally, the role of an Academic Advisor should be:

- To maintain up-to-date files containing the records of each student that is considered an advisee

**Original printed page 3 · source PDF page 13 · Open source page**

- To provide students with correct academic information concerning the university and the major school or department, noting their progress towards the fulfilment of general education, minor, and major requirements.
- To help foster in each advisee the ability to make informed choices concerning alternatives, limitations, and the consequences of academic decisions (selection of courses, major declaration or change, dropping courses, etc.)

However, these circumstances have changed in the past few years in Australia. As ever before, the ideal scenario remains an ideal; indeed, it is even becoming less achievable and less realistic. This is because Australian non-private tertiary education providers have seen gradually bigger cuts in higher education budgets.

The higher education system in Australia has been undergoing a significant change for the past few years, and this trend will continue for years to come. Australian universities and colleges are social institutions embedded in the general economy. Consequently, changes that are now reshaping Australian businesses and organisations inevitably influence higher education as well. Resource scarcity, increased competitiveness, downsizing and outsourcing, and a generally more turbulent, less predictable environment are the characteristics of the post-industrial environment. Modern technology has altered the expectations of consumers and the behaviour of service providers. Businesses that successfully take advantage of technological innovations are able to create efficiencies that allow them to recapture and sustain a high level of productivity and competitiveness.

While a greater number of students have access to tertiary education through increased quotas and full-fee places, staff-student ratios have decreased. In turn, this has adversely affected the availability of staff to provide academic advice to students.

Just as the higher costs of technology and human resources have affected business organisations, the steadily increasing cost of delivering higher education programs, coupled with enrolment increases predicted for the next decade, will create serious challenges for public education institutions. Many economists and higher education administrators believe that the economic

**Original printed page 4 · source PDF page 14 · Open source page**

and financial future of higher education is in jeopardy [6]. During the past two decades, per-student expenditures have increased while government student support and legislative funding for public higher education have declined. Tuition at public universities has soared to levels that alarm the public, and likely cannot be continued. Without changes in the delivery of higher education, aggregate budget shortfalls at Australian universities are expected to rise dramatically as we move further into this century [17].

At institutions of higher education, undergraduate enrolment will likely rise by 12% between 1996 and 2008. The traditional university age population of eighteen- to twenty-four-year-olds is projected to increase at a greater rate, by 18 percent, and is expected to comprise 60% of enrolments by 2008. It is estimated that students twenty-five and older will account for 40% of enrolments by that date [9]. Enrolment growth is a cause for concern, rather than a solution to educational finance problems, because tuition revenues cover only about one-third of the actual cost of public higher education. Declining federal and state government support will not make up the difference [27, 46]. Cost containment and finding new funding sources are clearly the primary challenges for higher education administrators.

The need to expand capacity to accommodate an increasing influx of students, and enhance the ability to provide sound and satisfying academic advice in an effective and efficient manner, is greater than ever before. With greater acceptance of automation technologies by the public, there is a better possibility of the successful provision of academic advice to remote students (e.g., via the Web). To achieve this goal, we require a system that “automates” the provision of academic advice as much as possible.

Web advising for students has recently been gaining in popularity at universities across the globe. Although only a handful of universities use this kind of system at present, human academic advisors are contemplating the various benefits associated with its introduction. The aim of this thesis is to develop a basic framework for automated academic advisement, and then to design a prototype system that integrates conventional academic advising and Web-based advising to form a hybrid advising model that can be easily applied to the advising process. Automated academic advisement reduces the workload of human academic advisors to some

**Original printed page 5 · source PDF page 15 · Open source page**

extent by giving information to students before approaching them. While the academic advisor assists them in this process, the final responsibility rests on the shoulders of the student. This means that students must actively become involved in their own education choices. At this point, the primary focus of the prototype automated academic advisement system is to provide educational planning and advising resources in support of students’ academic objectives.

The convergence in Artificial Intelligence (AI) and Cognitive Science gave birth to a new technique in Expert Systems (ES), namely Case-Based Reasoning (CBR). In short, CBR has the potential to capture the experience of academic advisors and use that experience as a guide to provide sound academic advice to students.

In CBR, systems expertise is embodied in a library of past cases, rather than being encoded in classical rules. Each case typically contains a description of the problem plus a solution and/or the outcome. The knowledge and reasoning process used by an expert to solve the problem is not recorded, but is implicit in the solution.

To solve a current problem, the problem is matched against the cases in the case base and similar cases are retrieved. The retrieved cases are used to suggest a solution that is reused and tested for success. If necessary, the solution is then revised. Finally, the current problem and the final solution may be retained as part of a new case.

Many people favour case-based reasoning because they are more comfortable with examples than with conclusions separated from their context. A case library can also be a powerful corporate resource, allowing everyone in an organisation to tap into the case library when handling a new problem.

Since the 1990s, CBR has experienced widespread academic and commercial interest. Mature tools and application-focused conferences exist. Case-based reasoning is often used as a generic term to describe techniques including, but not limited to, case-based reasoning as it is described (e.g., analogical reasoning is often referred to as case-based reasoning).

**Original printed page 6 · source PDF page 16 · Open source page**

### 1.2 Thesis overall goals

This thesis proposes a basic framework and outlines a prototype implementation of a case-based academic advisement recommender system with the following four overall goals:

1. To minimise repetitive tasks currently performed by academic advisors - The prototype case-based academic advisement recommender system demonstrates that its implementation can minimise the amount of time and energy spent by academic advisors on repetitive tasks such as answering frequently asked questions in person, by phone, or by email.

2. To extend the availability of official advising related information to distance online students - One of the most immediate and visible benefits of the proper implementation of the prototype case-based academic advisement recommender system is to make official and relevant advising information available on the Web, which would enable ubiquitous academic advising for distance online students.

3. To provide academic guidance in a consistent way - By having all of the advising guidelines stored electronically in one place, and using the same case-based academic advisement recommender system to advise students on which courses to take next, inconsistencies that used to occur with personal advising can be minimised.

4. To develop a backend reasoner that helps students to get a list of courses to take next - This is the “intelligent” portion of the system. Based on a student’s academic profile (courses already taken), the academic year requirements, and the database of existing and available courses and their prerequisites, the prototype case-based academic advisement recommender system provides the student with customised advice on which classes to register for the next semester(s).

**Original printed page 7 · source PDF page 17 · Open source page**

### 1.3 Organisation of the Thesis

The remainder of this thesis is organised as follows. Chapter 3 describes the history of expert systems and case-based reasoning. Chapter 4 describes the history of academic advisement and how information technology has been utilised to improve the efficiency of academic advisement to date. Chapter 5 explains why case-based reasoning is suitable for these types of problems. It also outlines the development of the prototype case-based academic advisement recommender system since its inception, validating the prototype system by means of test cases. Finally, Chapter 6 concludes the thesis with a summary of the general features of the prototype system and discusses the possible future extension of this system.

# Chapter 2 CBR background

**Original printed page 8 · source PDF page 18 · Open source page**

## 2.1 History of AI

The potential for computers to achieve “intelligence” equivalent to that of human beings was recognised by the pioneers of modern digital computing such as Alan Turing [42]. The goal of artificial intelligence (AI) scientists is to develop computer programs that can think and solve problems as intelligently as human experts. Expert systems are computer programs that use domain specific knowledge to emulate the reasoning process of human experts. It was not until the late 1970s that AI scientists realised that the problem solving power of a computer program mainly derives from the knowledge it possesses, rather than the inference mechanism it employs.

Developments in the early stages of the field of expert systems were aided by two major lessons learned from the work on AI and DENDRAL*. The first was the importance of encoding a rich source of knowledge within the program. General purpose reasoning techniques were recognised as too limited to solve real problems. The second lesson was related to the scope of

* From http://smi-web.stanford.edu/projects/history.html#DENDRAL

DENDRAL began as an effort to explore the mechanisation of scientific reasoning and the formalisation of scientific knowledge by working within a specific domain of science, organic chemistry. Another concern was to use AI methodology to better understand some fundamental questions in the philosophy of science, including the process by which explanatory hypotheses are discovered or judged adequate. After more than a decade of collaboration among chemists, geneticists, and computer scientists, DENDRAL had become not only a successful demonstration of the power of rule-based expert systems, but also a significant tool for molecular structure analysis in both academic and industrial research labs. Using a plan-generate-test search paradigm and data from mass spectrometry and other sources, DENDRAL proposes plausible candidate structures for new or unknown chemical compounds. Its performance rivals that of human experts for certain classes of organic compounds, resulting in a number of published papers in the chemical literature. Although no longer a topic of academic research, the most recent version of the interactive structure generator, GENOA, has been licensed by Stanford University for commercial use.

**Original printed page 9 · source PDF page 19 · Open source page**

the system’s knowledge. Systems that were designed for well-focused issues performed far better than those addressing broad problems.

MYCIN† was one of the first expert systems developed, and it is a benchmark for today’s rule-based expert systems, which utilize Rule-Based Reasoning (RBR). Many of the commonplace features in today’s systems were first developed during the MYCIN project. MYCIN [35] was developed at Stanford University to aid physicians in diagnosing and treating patients with infectious blood diseases caused by bacteremia (bacteria in the blood) and meningitis (bacterial disease that causes inflammation of the membrane surrounding the brain and spinal cord). These diseases can be fatal if not recognised and treated quickly. The system was developed during the mid 1970s, requiring approximately 20 person-years to complete. MYCIN is a rule-based expert system that uses backward chaining and incorporates approximately 500 rules. The system was written in INTERLISP, a dialect of the LISP programming language.

During the 1970s, the field of AI was moving toward knowledge-based systems - systems that contained and emphasised expert knowledge. Since the aid of an expert on infectious blood

† From http://smi-web.stanford.edu/projects/history.html#MYCIN

MYCIN is an interactive program that diagnoses certain infectious diseases, prescribes antimicrobial therapy, and can explain its reasoning in detail. In a controlled test, its performance equalled that of specialists. In addition, the MYCIN program incorporated several important AI developments. MYCIN extended the notion that the knowledge base should be separate from the inference engine, and its rule-based inference engine was built on a backward-chaining, or goal-directed, control strategy. Since it was designed as a consultant for physicians, MYCIN was given the ability to explain both its line of reasoning and its knowledge. Because of the rapid pace of developments in medicine, the knowledge base was designed for easy augmentation.

Furthermore, because medical diagnosis often involves a degree of uncertainty, MYCIN's rules incorporated certainty factors to indicate the importance (i.e., likelihood and risk) of a conclusion. Although MYCIN was never used routinely by physicians, it has substantially influenced other AI research. At the HPP, MYCIN led to work in TEIRESIAS, EMYCIN, PUFF, CENTAUR, VM, GUIDON, and SACON (described below), and to ONCOCIN and ROGET. The book, Rule-Based Expert System: The MYCIN Experiment at the Stanford Heuristic Programming Project, describes the decade of research on MYCIN and its descendants.

**Original printed page 10 · source PDF page 20 · Open source page**

diseases was sought to help solve the problem, this problem was a prime candidate for a knowledge-based approach.

In summary, some of the primary motivating factors for building MYCIN were:

- An expert was required to solve the problem
- Experts on the problem were scarce or unavailable because of time constraints
- Immediate expertise was needed in a possibly life threatening situation
- Time constraints required decisions to be made with limited or inexact information
- The computer solution needed to be accommodating to the user, who may have limited experience with computers
- Remembering the appropriateness and possible contraindications of a large number of drugs was a challenge for the physician

### 2.2 RBR background

The rule-based systems that are used today are a product of an evolutionary process that has taken place over the past 30 years. Earlier studies broke ground for the technology and pioneered techniques that evolved through trial-and-error into established methods for designing effective systems.

The idea behind the production (situation-action rule) system is simple, but effective. It offers the argument that humans solve problems by combining problem-specific information contained in their short-term memory with a set of productions representing problem solving knowledge contained in their long-term memory. The production system became the foundation of today’s rule-based expert system.

A rule-based expert system models the production system using the following modules:

- Knowledge base - Models a human’s long-term memory as a set of rules
- Working memory - Models a human’s short-term memory and contains problem facts both entered and inferred by the firing of the rules

**Original printed page 11 · source PDF page 21 · Open source page**

- Inference engine - Models human reasoning by combining problem facts contained in the working memory with rules contained in the knowledge base to infer new information

In the rule-based system, the rules contained in the knowledge base represent the productions contained in the long-term memory, and the facts contained in the working memory represent the situations in the short-term memory. The inference engine acts as the reasoning module of the production system model and compares the facts with the antecedents or premises of the rules to see which ones can fire. Those rules that can fire have their conclusions added to the working memory, and the process continues until no other rules have antecedents that match the facts contained in the working memory. Rule-based expert systems are not an exact match for human problem solving, but they provide a reasonable model for replicating this behaviour with a computer.

Traditional rule-based expert systems use exact matching which may not be the best approach for solving many real-world problems. An alternative approach is to use fuzzy rules.

### 2.2.1 Fuzzy Rules

There are two types of fuzzy rules: fuzzy mapping rules, and fuzzy implication rules. A fuzzy mapping rule describes a functional mapping relationship between inputs and an output using linguistic terms, while a fuzzy implication rule describes a generalized logic implication relationship between two logic formulas involving linguistic variables and imprecise linguistic terms. The foundation of a fuzzy mapping rule is a fuzzy graph, while the foundation of a fuzzy implication rule is the narrow sense of fuzzy logic. The two types of fuzzy rules are also related to different disciplines. Fuzzy mapping rules are related to other function approximation techniques in system identification and artificial neural networks, whereas fuzzy implication rules are related to classical two-valued logic and multivalued logic.

**Original printed page 12 · source PDF page 22 · Open source page**

#### 2.2.1.1 Fuzzy Mapping Rules

In many real-world problems, we are interested in finding the functional relationship between a set of observable parameters and one or multiple parameters whose values we do not know. For example, a financial trading broker is interested in the relationship between the history of various financial indices and future stock values. A control engineer is interested in finding an appropriate control law, which can also be viewed as a functional relationship between observable state variables and controllable parameters. Indeed, fuzzy logic controllers use rules of this type to approximate a mapping (typically nonlinear) from the observed state to a desired control action. In fact, most industrial applications of fuzzy logic use fuzzy mapping rules.

The needs to approximate a function of interest is often due to one or more of the following reasons, First, the mathematical structure of the function is not precisely known. Were the structure known, one could use various parameter identification techniques to find the parameters. Second, the function is so complex that finding its precise mathematical form is either impossible or practically unfeasible due to its high cost. Third, even if finding the function is not impractical, implementing the function in its previse mathematical form in a product or service may be too costly. This is particularly important for low-cost high volume products (e.g., automobiles, cameras, and other consumer products).

Even though human insights about the function can often be captured using fuzzy rules, the focus of this type of fuzzy rule inference is on exploring the trade-off between accuracy and cost. This actually echoes the motivation for developing fuzzy logic technology. Because each fuzzy rule approximates a small segment of the function, the entire function is approximated by a set of fuzzy mapping rules. Such a collection of fuzzy mapping rules will be referred to as fuzzy rule-based models or simply fuzzy models. The inference (i.e., mapping) for this type of rule is always in the forward direction. The main difference between fuzzy rules and non-fuzzy rules for function approximation lies in their “interpolative reasoning” capability, which allows the output of interpolative reasoning, can be viewed as a kind of partition-based modelling in system identification.

**Original printed page 13 · source PDF page 23 · Open source page**

Function approximation techniques can be broadly classified in three categories: global techniques, superimposition techniques, and partition-based techniques. The global techniques approximate a function globally using one mathematical structure (e.g., linear, second-order polynomial, etc.). A major issue of this technique lies in finding the suitable model structure for a given problem. The superimposition techniques approximate a function by superimposing functions of a given form (e.g., B-splines, Taylor expansions). The partition-based approximate techniques approximate a function by partitioning the input space of the function and approximate the function in each partitioned region separately (e.g., piecewise linear approximation). The latter two are also referred to as local modelling techniques, while the first is referred to as a global modelling technique.

Fuzzy rule-based function approximation is a partition-based technique. It generalizes classical partition by allowing a subregion to partially overlap with neighbouring sub-regions. This will be referred as a fuzzy partition. In the input space where subregions partially overlap with each other, the function is approximated using a kind of interpolation technique.

#### 2.2.1.2 Fuzzy Implication Rules

Fuzzy implication rules are a generalization of “implication” in two-valued logic. Their aim is to mimic human reasoning in its ability to reason with ideas or statements that are imprecise by nature. Even though it is obvious that human beings perform certain kinds of approximate reasoning, it is unclear how one can characterize such a reasoning process due to our present limited understanding of the human reasoning process to date. Consequently, there has not been posited a unique set of desired properties for fuzzy implication rules. Rather, several sets of desired properties have been developed. Even though the set of desired properties is not unique, they are useful for comparing different reasoning schemes using fuzzy implication rules. This “non-uniqueness” aspect of fuzzy implication should be pointed out, because it is in sharp contrast with many other techniques in science and engineering that were built on a set of well-defined axioms, properties, or principles.

**Original printed page 14 · source PDF page 24 · Open source page**

The inference of fuzzy implications generalizes two kinds of logic inference using the implications in classical logic: modus ponens and modus tollens. The logic inferences can e briefly illustrated here, using the following implication:

IF a person’s IQ is high THEN the person is smart.

Given the implication and the fact that “Jack’s IQ is high,” modus ponens enables us to infer that “Jack is smart.” On the other hand, if we are given the implication and the fact that “Jack is not smart,” modus tollens enables us to infer that “Jack’s IQ is not high.”

Even though classical logic can perform these two kinds of inferences, they are limited in two ways. First, these inferences insist on perfect matching. If Jack’s IQ is more or less high, for instance, no modus ponens inference can be made from the implication above because “more or less smart” is not identical to “smart.” However, our “common-sense reasoning” usually suggests that we can infer that “Jack is more or less smart.” Second, these inferences cannot handle uncertainty. For instance, if Jack told us his IQ is high but cannot provide documents supporting the claim, we may be somewhat uncertain about the claim. Under such a circumstance, however, logic cannot reason about the uncertainty.

These limitations motivated L.A. Zadeh [48] to develop a reasoning scheme that generalizes classical logic so that (1) it can conduct “common-sense” reasoning under partial matching, and (2) it can reason about the certainty degree of a statement. In particular, logic implications are generalized to allow partial matching. In the previous example, this means that we can infer “Jack is somewhat smart” from “Jack’s IQ is somewhat high” and the fuzzy implication “If Jack’s IQ is High, then Jack is Smart.” As shown below.

Given: Jack’s IQ is High → Jack is Smart.

Jack’s IQ is somewhat High.

Infer: Jack is somewhat Smart.

**Original printed page 15 · source PDF page 25 · Open source page**

The second limitation of logic (i.e., inability to deal with uncertainty) has motivated another extension to classical logic: multi-valued logic. Multi-valued logic extends the truth values in logic from binary (i.e., true and false) to multiple (i.e., more than two) so that different degrees of certainty are expressed by different truth values. Since fuzzy logic also generalizes the truth values in classical logic beyond true and false, it is related to multi-valued logic. However, fuzzy logic differs from multi-valued logic in that it also addresses the first limitation of logic (i.e., restricted to a perfect match). The reason that a fuzzy implication can address the first problem is that it uses linguistic variables in its antecedent (i.e., if parts). Consequently, the statement in the antecedent describes an elastic condition that can be partially satisfied.

In the previous example, “A person’s IQ” is a linguistic variable in the antecedent, and “A person’s IQ is high” describes an elastic condition about a person’s IQ. This enables a person with a somewhat high IQ to partially match the condition. Obviously, the degree to which the condition in a fuzzy implication is matched influences the inference that can be made from the fuzzy implication.

Other approaches fro reasoning under uncertainty include Bayesian probabilistic inference, the Dempster-Shafer theory, and non-monotonic logic. While these approaches have their merits for reasoning under uncertainty, fuzzy logic is unique in that it addresses both the uncertainty management problem and the partial matching issue.

The distinctions between fuzzy implication rules and fuzzy mapping rules are subtle, yet important. Prof. Zadeh mentioned that fuzzy logic has four principle facets: (1) the logical facet, (2) the set-theoretic facet, (3) the relational facet, and (4) the epistemic facet. Fuzzy implication rules are primarily in the logic facet, whereas fuzzy mapping rules are in the relational facet because functional mapping is a kind of relation.

**Original printed page 16 · source PDF page 26 · Open source page**

### 2.2.2 Disadvantages of Rule-Based Systems

#### 2.2.2.1 RBR require exact matching

The rule-based system attempts to match the antecedents of the available rules with the facts in the working memory. This is the primary characteristic of backward chaining expert systems. The match must be exact for this process to be effective, which in turn requires a strict adherence to consistent coding. For example, consider the following rule:

IF The motor is hot

THEN Shut the motor down

If we were to place in the short-term memory either of the following two statements:

- The motor is running hot
- The motor’s temperature is hot

We would not have an exact match and the rule would not fire. As humans, we can easily interpret these two statements as meaning the same thing as the antecedent of the example, but the computer is unforgiving of difference in the statement’s syntax.

#### 2.2.2.2 RBR have opaque rule relationships

Although an individual rule may be easy to interpret, it is often difficult to determine how rules are logically related through an inference chain. For example:

IF C THEN D, IF B THEN C, IF A THEN B

The first rule is dependent on the other two, and the second is dependent on the third rule. During testing and debugging, it is important to be able to locate these related rules. Since the rules can be placed anywhere in the knowledge base, and the number of rules can be large, it is often difficult to locate and trace the related rules.

**Original printed page 17 · source PDF page 27 · Open source page**

#### 2.2.2.3 RBR can be slow

Systems with a large set of rules can be slow. This difficulty occurs because the inference engine must scan the entire set of rules when deciding which rules to apply. This can result in slow processing times, which can be detrimental in real-time applications.

### 2.2.2.4 RBR are inappropriate for some problems

Another shortcoming of the rule-based system occurs when rules do not efficiently or naturally capture the representation of the domain's knowledge. Davis and King [8] comment on this deficiency: "Program designers have found that production systems easily model problems in some domains, but are awkward for others." Rules provide only one choice for representing the problem's knowledge in the expert system. Other techniques include case-based reasoning, frames, semantic networks, decision tables, etc.

Another difficulty is that the rules obtained from human experts are highly heuristic in nature. These heuristic rules tend to be "brittle" and cannot handle missing information or unexpected data values, although certainty factors may be introduced to handle missing data. Another aspect of this brittleness is a tendency to degrade rapidly near the "edges" of the domain knowledge. Unlike humans, RBR is usually unable to fall back on first principles when confronted with novel problems.

## 2.3 History of CBR

Case-based reasoning (CBR) is an alternative to RBR that approaches problem solving in quite a different manner. This technique uses an explicit database of problem descriptions and solutions to address new problem solving situations.

The modern development of CBR derived from research into cognitive science. In hindsight, it is also possible to find references of significance to CBR in Wittgenstein's observation (1953 Philosophical Investigations UK: Blackwell) that natural concepts such as tables and chairs are

**Original printed page 18 · source PDF page 28 · Open source page**

polymorphic and cannot be classified by a single set of necessary and sufficient features, but instead can be defined by a set of instances (i.e., cases) that have family resemblances.

The work of Schank and Abelson [34] is widely held to be the origins of CBR. They proposed that our general knowledge about situations is recorded in the brain as scripts, and these scripts allow us to set up expectations about what we will hear about, which in turn allow us to perform inferences (i.e., to infer the relationships between the things we hear). Case-based reasoning could also be derived from a view of understanding as an explanation process [33]. As we read a text or listen to a discussion, we use our knowledge about what is being written or talked about to help us tie together the pieces. Our knowledge helps us predict what we will hear or read next, establish a single semantic or grammatical interpretation for words, resolve pronouns, and make connections between disparate elements. Making connections or explaining the interactions of the individual elements we hear or read is particularly important in comprehension.

While many may lay claim the philosophical roots of CBR, it was the pioneering work of Roger Schank's group at Yale University in the early 1980s that produced both a cognitive model for CBR and the first CBR applications based on this model.

Scripts were proposed as a structure for conceptual memory, describing information about stereotypical events such as going to a restaurant or visiting a doctor. However, experiments on scripts demonstrated they were not a complete theory of memory representation – people often confuse events that have similar scripts. For example, a person might mix up room scenes from a visit to a doctor's office with a visit to a dentist's office.

In another word, scripts provide only one kind of knowledge; we also use knowledge about goals, plans, interpersonal relationships, the roles people play, character traits, and so on in our understanding. Understanding is seen as a knowledge application process in which the explanations that tie together the individual things we encounter are formed. [23]

**Original printed page 19 · source PDF page 29 · Open source page**

Roger Schank also explored the role that memories of previous situations (i.e., cases) and situation patterns, or memory organisation packets (MOPs), play in problem solving and learning. The theory of Dynamic Memory [33], proposed by Schank and his students in the late 1970s [23], presented indexing as the key to using experience in understanding. The premise was that remembering, understanding, experiencing, and learning cannot be separated. Our memories, which are dynamic, change as a result of the new things we encounter, the questions that arise in our minds because of these experiences, and the way we answer these questions. We understand by trying to integrate new things we encounter with what we already know. Thus, understanding causes us to encounter old experiences as we process new ones. Those experiences provide expectations that allow the understanding process to be primarily simple and top-down. Individual experiences and general descriptions of normative experiences are encountered as we understand - both provide the same sorts of information to understanding processes. The side effect of this process is that a dynamic memory never behaves exactly the same way twice because it changes with every single experience. Thus, reasoning processes never encounter exactly the same knowledge on two traversals through memory.

The means of selecting indexes and integrating general knowledge with experiences are primarily important in this theory of dynamic memory. The theory proposed that memory is organised by memory organisation packets (MOPs), which have two functions:

- They hold general knowledge
- They organise, in a complex hierarchy, specific experiences (cases) of that general knowledge

Janet Kolodner developed the first CBR system, called CYRUS, in 1983. CYRUS contained knowledge - as cases - of the travels and meetings of former U.S. Secretary of State Cyrus Vance, and allowed users to ask questions about these events. CYRUS was an implementation of Schank’s dynamic memory model. Its case-memory model later served as the basis for several other influential CBR systems.

An early alternative approach came from Bruce Porter’s work into heuristic classification and machine learning at the University of Texas in Austin, and resulted in the PROTOS system.

**Original printed page 20 · source PDF page 30 · Open source page**

PROTOS combined general domain knowledge and specific case knowledge into a single case-memory model.

In the late 1980s, the U.S. DARPA program funded a series of workshops on CBR and the development of a CBR tool. This tool became Cognitive Systems’ ReMind and marked the transition of CBR from purely academic research in cognitive science and artificial intelligence to the commercial arena. ReMind was joined in the marketplace almost immediately by several other tools.

Bergmann [4] pioneered a system called “Induction and Reasoning from Cases” or INRECA. The initial goal of this project was to develop innovative technologies that could help people make smarter and faster business decisions using cases, and to integrate these technologies into a single software platform that would allow the technology to be used more widely. In late 1995, these goals were achieved and the methodology of INRECA was proven very effective among CBR system designers and developers.

Goker and Roth-Berghofer [15] adapted this methodology to improve their old model for their existing help desk support system HOMER, which was originally developed by Goker, Roth-Berghofer, Bergmann, Pantleon, Traphoner, Wess, and Wilke [15]. The original CBR process model involved four main tasks: retrieve, reuse, revise, and retain. The improved model was broken down into two cycles, namely the application cycle and the maintenance cycle, with two newly added tasks: recycle and refine. The two process cycles allow the two cycles to work more autonomously and independently of each other. The flexibility for this help desk system to be used in a dynamic, corporate environment by a large group of users was the outcome.

The success of this adaptation was included in INRECA-II and highlighted by Bergmann [4]. Other highlights included the integration of induction and CBR, the evaluation of CBR systems, a methodology for developing CBR applications, and a reasonable number of INRECA applications.

Original printed page 21 · source PDF page 31 · Open source page

Similar to information retrieval, CBR relies heavily on algorithms to retrieve cases efficiently. A category of these algorithms is Nearest Neighbour, which is also well known and intensively studied for the solution of classification and pattern recognition problems. Short and Fukunaga [35] introduced the Value Difference Metric and the Euclidean-Hamming metrics, which were accepted and used in the field of information theory until recently. Blanzieri and Ricci [5] extended SF2 from the Short. and Fukunaga metric, with its definition to input spaces with nominal features, and introduced a different estimate for the density probability used in this metric. They also modified SF2 metric to be the Minimum Risk Metric (MRM) by optimising the risk of misclassification instead. The MRM proved was less restrictive than SF2, and thus seemed particularly suitable for CBR applications when a relevant classification of the cases is available. Further work was required to address in depth the relation between MRM and the base mathematical theory, as well as to explore the performances of the metrics in higher dimensions and their sensitivity to noisy data.

Another approach was to adapt relational CBR. Armengol and Plaza [3] developed LAUD, a distance measure that can be used to estimate similarity among relational cases. However, this added the need to include feature terms into each case. LAUD was proposed to estimate the similarity of relational cases as an aggregation of the similarities of the features that are common to both cases. It was also shown that in their extension of similarity assessment to relational cases, LAUD is not much more computationally intensive than a distance-based assessment for propositional cases. This characteristic was deemed cost effective enough for most CBR applications to switch to this new approach.

One variant of CBR that seems to be gaining popularity in the past few years is Conversational CBR (CCBR). CCBR used a dialogue similar to a Questions & Answers style to derive the user's preference. Goker and Thompson [15] developed the Adaptive Place Advisor. While most recommendation systems accepted keywords and produce a ranked list, the Advisor stood out by conducting a conversation with users to progressively narrow their options. Unfortunately, their current similarity calculation did not take into account the effects of a combination preferences and diversification. They play an important role in the user's approach to selecting an item, and they should be incorporated into the Advisor in the future.

Original printed page 22 · source PDF page 32 · Open source page

The history of CBR is not as long as other decision-support technologies such as rule-based system or neural networks. Ian Watson [43] believes that its rapid uptake is because of its intuitive nature and relative simplicity of implementation.

### 2.4 Case-Based Reasoning Approaches

Case-Based Reasoning involves learning from previous experiences. Given that this is a very general approach to human problem solving behaviour; it is natural that there are different approaches for implementing this process in computer systems. In commercial CBR systems, the three main approaches differ in the sources, materials, and knowledge they use.

#### 2.4.1 Textual CBR Approach

The textual CBR approach eases case acquisition. It is very useful in domains where large collections of expertise documents already exist and the intended user is able to immediately make use of the knowledge contained in the respective documents. This approach is well suited when there are not too many cases at a time (less than a couple of hundred) and each case has a short description (three sentences at most). Otherwise, textual CBR retrieves a large number of cases that are irrelevant. The cost for controlling the quality of textual CBR is high.

The developer of the case-based solution may decide to have cases recorded as free text. For example, this could be course descriptions or academic reports. The CBR retrieval engine uses various keyword-matching techniques to retrieve the cases. This can be based on string searches or on N-grams. N-grams is a technique that separates words into lists of N characters. For example, for tri-grams, the word

“academic” is turned into the list of three characters “aca”, “cad”, “ade”, “dem”, “emi”, and “mic”. Both the query and the case base are transformed, and case matching is performed on these. This technique is robust even when there are spelling mistakes. A word like “acdemic” would be recognised as similar to “academic”. The main disadvantage of N-grams is that almost anything matches everything when there are large blocks of texts with many words. The improvement of this is to use a proportional count, with many

**Original printed page 23 · source PDF page 33 · Open source page**

non-hits counting against case recall. That is if two cases had the same number of matches, the case with the longer text would appear after the shorter one.

Two parameters are used to assess the quality of a textual CBR system: recall and precision. Recall measures how many cases are retrieved from the set of existing relevant cases. Precision measures how many of the retrieved cases are relevant. As the case base grows, there is often high recall with low precision - the system retrieves many cases, but only a few of these cases are relevant.

Example: Frequently Asked Questions. An academic advisor has a collection of documents that represent the current set of frequently asked questions on his website. Each document contains a single problem and its solution. Navigating and searching the list of documents is difficult for students, as well as for a replacement advisor if the designated advisor is away. To avoid this unfriendly search process, students tend to call the advisor instead of solving the well-known problems on their own. As a result, consultation time is wasted on nonsense or trivial queries rather than being spent on real issues from disadvantaged or troubled students. The goal of CBR is to help students and advisors find solutions to problems that are already known and documented.

### 2.4.2 The Conversational CBR Approach

The conversational CBR approach is very useful for domains where a high volume of simple problems must be solved repeatedly. The system guides the agent and the customer with predefined dialogs. However, the case author organises the case base manually, which is a complex and costly activity when the cases are described by many attributes (questions). The conversational approach is well suited for applications in which only a few questions are needed for decision-making. Maintenance costs are high because the developer must manually position each new case in a decision tree-like structure and update the ordering of the questions.

In the context of academic advising, the principle is to capture the knowledge contained in student/advisor conversations. A case is represented through a list of questions that varies from

**Original printed page 24 · source PDF page 34 · Open source page**

one case to another. There is no domain model and no standardised structure for all of the cases. This may produce unreliable behaviour in the resulting CBR system, which may ask questions during the consultation that are irrelevant (e.g., questions about 4th year electives when the student has just completed 2nd year and is entering the 3rd year of the program). To overcome this problem, questions describing the cases are sometimes organised into tree-like structures. The developer creates groups of questions that are used to describe a certain subset of the case base. Each case is then positioned by hand in this structure, which must be maintained manually by the developer. This task can be compared to developing an expert system that relies on decision trees.

To index the cases, the case author must also define the order the user is asked to answer the questions during the consultation. Thus, the work required to develop a conversational approach CBR solution can be compared to building an expert system. The difference is that this “expert system” displays cases to the user during the consultation.

### 2.4.3 The Structural CBR Approach

The structural CBR approach relies on cases that are described with attributes and values that are predefined. The structural CBR approach is useful in domains where additional knowledge beside cases must be used to produce good results. The domain model ensures that new cases are high quality and the maintenance effort is low. This approach provides better results than the two others, but it requires an initial investment to produce the domain model.

The idea underlying the structural CBR approach is to represent cases according to a common structure called the domain model. The domain model specifies a set of attributes (also called features) that are used to represent a case. The domain model structures this set of attributes. In simple applications, the structure may be a flat table like those produced by a spreadsheet. In complex domains, the case description cannot be a flat table.

A relational or object data model can be used in such applications. In a relational model, several tables are indexed by a primary key. Each case has a unique key, and the case description is

**Original printed page 25 · source PDF page 35 · Open source page**

distributed in several tables. The relations of the database define the distribution. In an object-oriented database, objects are decomposed into sub objects. For example, in an academic advisement application, a course points to several sub objects, such as related fields of study and perceived difficulty. Each object has its own set of attributes, such as the list of all fields of study and the pass rate or high distinction rate.

Another important aspect of the structural CBR approach is that each attribute has a specific type. The attribute can be a symbol that takes one or several predefined values in a list. For example, the owning discipline of the course may be Software & Networks, Communications, Electronics, Electrical, and so on. The data type can also be a number (integer or real) within a certain range. For example, the pass rate of a course is a percentage between 55% and 65%. Some CBR systems provide additional powerful attribute types. For example, the course difficulty can be ordered as easy, moderate, or difficult. A taxonomy type can describe relations of generality among attribute values. For example, for the attribute discipline, sub discipline values; "Linux" and "Microprocessor" belongs to discipline software and networks; "Microwave" and "Satellite" belong to communications, and so on. The ability to represent and use this type of background knowledge can be used to discover additional similarities between two cases. For example, if a student wants to enrol in a microprocessor course, a course about micro assembler would be recommended over a course about C++.

To summarise, the domain model defines a standard way to represent all cases. Note that this is different from the textual CBR approach, which may contain a structure using languages such as Standard Generalized Mark-up Language (SGML‡), but where the content associated to each

‡ SGML is a standard for specifying a document mark-up language or tag set. Such a specification is a document type definition (DTD). SGML is not a document language, but a description of how to specify one. It is metadata. SGML is based on the idea that documents have structural and other semantic elements that can be described without reference to how such elements should be displayed. The actual display of such a document may vary depending on the output medium and style preferences. Some advantages of documents based on SGML are:

- They can be created by thinking in terms of document structure rather than appearance characteristics (which may change over time).

**Original printed page 26 · source PDF page 36 · Open source page**

header is not standardised (it contains free text). Even if the content of these headers is stored in a database (using what is called text blobs), this is not sufficient to qualify for the structural approach. In the structural CBR approach, each attribute must be a field in the database and each value must be standardised. Table 2-1 in the next page compares between different types of CBR.

- They will be more portable because an SGML compiler can interpret any document by reference to its document type definition (DTD).
- Documents originally intended for the print medium can easily be re-adapted for other media, such as the computer display screen.

The language that this Web browser uses, Hypertext Mark-up Language (HTML), is an example of an SGML-based language. There is a document type definition for HTML (and reading the HTML specification is effectively reading an expanded version of the document type definition). SGML is based somewhat on earlier generalised mark-up languages developed at IBM, including General Mark-up Language (GML) and ISIL.

**Original printed page 27 · source PDF page 37 · Open source page**

**Table 2-1: Comparison of the different CBR approaches**

| Criteria | Textual CBR | Conversational CBR | Structural CBR |
|---|---|---|---|
| Case Base | A case is represented in free-text format. The cases may be structured according to headers, but the content of the headers is in free-text form. The case base is the collection of free texts that may be, for example, in electronic documents that are accessed on the Internet. | Cases are represented by a list of question/answers. The list of questions and answers may vary from one case to another. There is no common data structure. The questions may be stored in a database but each question is not a separate field of the database. | The case base is made out of records in a database. The records contain field whose values are standardised. Specific cases are recorded by assigning values to the attributes that have been defined. The initial case base may easily be extracted from an existing database. |
| Query Case | A query is represented by question in free-text form whose content is similar to the description part of the stored cases. | A list of questions and answers following a dialog with the user. | The query consists of a set of attribute values. These may be obtained by asking questions. In an academic advisement application, attributes are fields of interest and academic history. |
| Results | A list of documents that might be useful | A list of possible actions attached to the cases that have been retrieved | A list of database records that describe cases |
| Background Knowledge | Dictionaries of similar terms like “High Distinction: competent” or “Microwave: Wireless” and information about the relationships between different words | Ordering of questions, hierarchy of questions in a tree-like structure, possible answers for the questions that have been defined | Domain model that defines the case structure. Rules to deduce values. Information on how to compute the similarity among attributes values |
| Initial Effort | Defining terms, synonyms and stop-words. Analysing and setting up relationships between different words | Storing dialogs within a case base, structuring the questions to organise the case base manually, and ordering the questions manually. This can be done by entering existing decision trees | Defining the domain model. Importing the database or collecting cases according to the structure of the domain model. This can be done, for example, using an electronic questionnaire |
| Case Creation | Writing new documents | Adding new dialogs. Depending on the structure of the dialog, new questions must be entered. | Adding a new record to the database. A questionnaire (forms) can be used in order to use the same vocabulary. |
| Maintenance | Maintaining the dictionary of controlled terms and sentences. Analysing user queries and results. | Maintaining the list of questions and answers. Eliminating doubles, combining answers and questions. Reordering the questions by hand for the consultation. | Maintaining the case base and the domain model. The CBR system must keep up with these modifications. It can automatically compute the order of the questions during consultation. |
| Multilingual Case Bases | Almost impossible. Must have different case bases. | Difficult, no standardised thesaurus of terms. | Easy to implement with a single case base. |
| Advantages | Existing documents can be used as cases. No initial investment is required for modelling the cases. | The approach is intuitive and easy to understand. The system asks questions which are answered by the user and this dialog is stored in the case base. | An existing database can be used to produce the initial case base. Creating additional cases can be done without any knowledge about how the CBR system will use the cases afterward. |
| Drawbacks | The user is not guided. The CBR system must cope with spelling | No explicit knowledge represented in the system for | It may be difficult to create a predefined case structure. This |

| Criteria | Textual CBR | Conversational CBR | Structural CBR |
|---|---|---|---|
| | mistakes. The quality of retrieval depends on the syntax and not on the real content of the cases. High costs for quality control of the CBR system. | computing similarity. The interdependencies among different dialogs cannot be predicted. High maintenance costs. | can usually be achieved within most technical domains, but it becomes harder in softer domains. High investment costs. |

**Original printed page 28 · source PDF page 38 · Open source page**

**Original printed page 29 · source PDF page 39 · Open source page**

**Original printed page 30 · source PDF page 40 · Open source page**

### 2.5 Case-Based Advisory Systems

Case-based advisory systems are also often called case-based aiding systems. They use interactive CBR engineering models to support human reasoning. Pragmatic CBR systems could be divided into two categories: automated CBR systems and interactive CBR systems. Automated CBR systems solve problems without human intervention. Most of the CBR systems that implement CBR cognitive models are automated CBR systems. In contrast, interactive CBR systems work with people to solve problems. Some interactive CBR systems require human support for their reasoning tasks. For example, some CBR systems need people to help with their adaptation, decision-making, or evaluation to successfully complete their tasks. Some interactive CBR systems are designed to support human reasoning. For example, some such systems act as case retrievers for a user who employs the cases to reason, while some provide other forms of support, such as help with adaptation or evaluation. The degree of automation in interactive CBR systems varies. However, no matter what degree of automation is used, those interactive CBR systems designed to support human reasoning are usually case-based advisory systems.

**Original printed page 31 · source PDF page 41 · Open source page**

Case-based advisory systems provide users with the opportunity to learn on the job. The more automated systems can act like an expert looking over the shoulder of the user, providing timely support on an as-needed basis. The simpler systems use their case libraries to augment the memory of a user solving a problem, and perhaps provide other forms of support as well. With the support of case-based advisory systems, users can make decisions about which cases could be used, which adaptations to apply, and which potential problems to address.

#### 2.5.1 Ways in which case-based advisory systems support user reasoning and learning

Case-based advisory systems typically support users in three ways: inform users of potentially useful cases, provide adaptation support, and provide evaluation support. First, case-based advisory systems can augment the user’s memory. The case library of a case-based advisory system is usually built to cover the range of reasoning tasks the system will be responsible for supporting, as well as cover the range of well known solutions and well known mistakes [23]. When users are trying to understand and assess situations, such representative sets of cases allow them to interpret or understand a situation in the context of other similar situations. Those similar situations can not only point out to users what to focus on and what outcomes might arise, but it also allows them to argue and justify the pros and cons of interpreting a situation in a certain way. It is especially helpful when users are interpreting open-ended and ill-defined concepts. In the process of problem-solving, similar cases retrieved from the case library of the case-based advisory systems allow users to:

- propose solutions to problems quickly, avoiding the time necessary to derive those answers from scratch;
- propose solutions in domains that they do not completely understand;
- be aware of the potential for problems that have occurred in the past and take actions to avoid repeating past mistakes

- focus their reasoning on the important parts of a problem; and
- propose solutions even when the logic behind the recommendations of the case are weakly presented or absent.

**Original printed page 32 · source PDF page 42 · Open source page**

In addition, the variety of cases included in case-based advisory systems expose users to multiple situations, multiple decisions, and multiple perspectives. This multiplicity helps to build awareness of knowing what skills to apply and when to apply them. A beneficial side effect of using the case-based advisory system to augment the user's memory is that users could acquire a representative set of cases for their own case libraries (memories) in the process of using the systems to support their reasoning.

Second, case-based advisory systems can support the adaptation process of reasoners by suggesting adaptation strategies, the applicable conditions of adaptation strategies, and past adaptation processes that could be usefully recorded as part of the case in their case libraries. Some case-based advisory systems, which integrate CBR method with semantic networks or other reasoning methods such as rule-based reasoning or model-based reasoning, can also provide users with relevant domain knowledge to support their adaptation. Adaptation support from case-based advisory systems can help users identify what to change, find out the available adaptation or repair strategies and their conditions, determine the appropriate strategies and, in cases where several appropriate strategies suggest different adaptations, they can help users choose among several appropriate strategies.

Third, case-based advisory systems can also help users with evaluation by showing them the factors that should be considered during evaluation, as well as clustering cases to make comparison easy. This could be done by:

- retrieving and presenting cases that have similar situations and similar solutions to those being evaluated, and
- suggesting which factors the users should focus on when comparing and contrasting new situations and their solutions to old ones in order to determine if an old outcome can apply to a new situation.

The outcomes of the old situations can be projected on the new ones, allowing the outcome of the new solutions to be predicted. When different old cases can be used to predict different outcomes, the factors suggested by case-based advisory systems can also help the users to do argumentation (i.e., to determine which of the old cases are similar enough to the new one for the projection to make sense). This process has the potential to help the users in several ways. First,

**Original printed page 33 · source PDF page 43 · Open source page**

it can point out the potential problems with a proposed solution and point the way towards repairing the solution. Second, it can point out what might be changed to make a good solution better. Third, scenarios used during critiquing can serve as a justification for whatever solution is selected. Fourth, it can help the users realise what is important to pay attention to in future problem solving.

Because case-based advisory systems have the potential to support reasoning and learning in so many ways, some case-based advisory systems have been built especially for instructional purposes.

### 2.6 Goal-based Scenario's (GBS) Top-down Approach to Design

The moderated Dynamic Memory CBR model implies that a "learning" academic advisement system should revolve around the target knowledge, skills, and experiences of the expert human academic advisors that it aims to teach by helping the CBR system accumulate the kind of cases that can lead to the acquisition of knowledge, generalisation of skills, and representation of past experiences. In essence, the design of a learning system should take a top-down approach starting with the higher level knowledge (skills and strategies of academic advisors), then going down to the bottom level to choose specific cases that would facilitate the generalisation of the higher level knowledge, and finally incorporating the conceptual knowledge to be taught into those cases. The cases chosen should be ones that are directly linked to academic advising skills, and the set of cases should be diverse in context and include both success and failure cases.

Correspondingly, in the design of GBS, the structure (mission structure) and content (scenario operations, mission, mission focus, and cover story) of the learning environment is determined by the target knowledge and skills. Advising experiences that do not directly contribute to the generalisation of case attributes are minimised. Advising experiences that directly contribute to the abstraction of case attributes are not only maximised, but they are also designed to be diverse in context and include both experiences of success and failure. Strategies and conceptual

**Original printed page 34 · source PDF page 44 · Open source page**

knowledge are presented either as part of a scenario or as part of the expert academic advisors' stories, which are presented to the CBR system when they experience expectation failures.

### 2.6.1 Analysis of Case-Based Advisory Systems and Knowledge-Sharing Systems

Case-based advisory systems and knowledge sharing systems are different from Goal-Based Scenario (GBS) and Learning by Design (LBD) not only because they are CBR systems, but also because the majority of them are designed to be job aids rather than educational systems. Most are built with engineering and economic concerns, and do not strictly follow any CBR cognitive model. Factors such as the nature of the subject matter, the target audience, and the resources available often determine what the framework of a specific case-based advisory system or case-based knowledge sharing system would be. This characteristic presents a great potential in the academic advisement application, with factors such as students' fields of interest, academic performance, and preferences towards technical field specialisation.

### 2.6.2 Present cases as well as overall knowledge representation

Kolodner's Category and Exemplar model, and some integrated CBR models, suggest there are semantic networks of general domain knowledge in people's memory. These CBR models acknowledge that overall knowledge representation of an expert academic advisor's semantic networks could affect his/her reasoning capacity. Therefore, these CBR models might suggest that case-based advisory systems and knowledge sharing systems should make the overall knowledge representation of the academic advisement domain available to students, who are essentially users. For example, if the CBR systems are structured according to the structure of the academic advisement domain, it might be beneficial to make the structure of the CBR system and indices of cases available to students.

**Original printed page 35 · source PDF page 45 · Open source page**

### 2.6.3 Make explicit important features, deep structure, and rules, models or strategies used

Kolodner's CBR cognitive model suggests that making important features, deep structure, and rules, models, and strategies explicit to users helps them focus on the important factors and make more thoughtful decisions. Her model implies that case-based advisory systems should present cases to users in such a manner and, in the case of case-based knowledge-sharing systems, those systems should also have features to help users input their experiences into the system in a similar structured manner.

Schank's moderated Dynamic Memory CBR model also implies that presenting cases in such a structured way would facilitate reasoning because it helps users to create a more effective indexing scheme. In addition, the Dynamic Memory CBR model suggests that it is desirable to present strategies in the context of cases, because:

- Strategies could only be acquired from cases
- After the users acquire strategies from those cases, their overall reasoning capability will be enhanced and they would be better prepared to reason effectively and efficiently.

### 2.6.4 Explain cases with generalised domain knowledge

The Category and Exemplar model, and several integrated CBR models, emphasise rich explanatory links between case knowledge and generalised domain knowledge. It proposes that when explained with generalised domain knowledge, cases are easier to retrieve, easier to adapt, more predictive, and more memorable. Therefore, it suggests that case-based advisory systems and knowledge sharing systems should include explanations that refer to generalised domain knowledge in their cases, and case-based knowledge sharing systems, in particular, should include features or guidelines to encourage users to explain their experiences with generalised domain knowledge whenever possible.

**Original printed page 36 · source PDF page 46 · Open source page**

### 2.6.5 Avoid reasoning bias by presenting exceptions

If the case-based advisory systems and knowledge sharing systems use the approach of explained cases with generalised domain knowledge, the Dynamic Memory cognitive model would suggest that exceptions to the generalised knowledge should also be made available to users. The Category and Exemplar model also suggests that information should be provided to users to help them become aware of similar situations in which the generalised domain knowledge would apply.

## 2.7 Knowledge supporting the CBR process

Learning by acquiring new cases is an integral part of the CBR process: each problem-solving episode provides a new case to save for future reuse. However, these cases form only one pool of knowledge that supports learning in this framework. CBR systems rely on at least four types of knowledge:

- The case base: the stored solutions known to the system
- The indexing scheme: the features used to retrieve cases from the case base. These features are often predefined and fixed within the system
- Similarity criteria: the method used for the selection of candidates retrieved during the retrieval process. Like indexes, the criteria used have traditionally been predefined by the user
- Case adaptation knowledge: the stored problem-solving episodes provide a rich source of knowledge that enables many types of learning

These four types of knowledge exist in support of various aspects of the CBR process. For example, the indexing scheme supports the retrieval of relevant cases from the case base. While the case base changes as new solutions are accepted by the system, the other three knowledge sources are generally unchanging and defined from the outset of processing in the system. With the acquired knowledge, there is a potential to affect the other knowledge sources already in the system. Kinley [23] claimed that the stored plan cases could be reapplied more effectively if better case adaptation knowledge is available.

**Original printed page 37 · source PDF page 47 · Open source page**

Case-based reasoning can be further improved by augmenting the retrieval similarity criteria with the stored case adaptation knowledge. Stored cases can be selected for reuse based on how easily the system can subsequently adapt them. The interaction of different knowledge sources to augment and overcome the deficiencies of one another has previously been established and documented (Richter, 1995). However, the types of knowledge interactions available with the addition of learned case adaptation knowledge have not been examined.

## 2.8 The ASK systems

The ASK systems [12] are hypermedia case-based advisory systems designed to simulate a conversation between a novice (the user) and an expert (the system). This series of systems (e.g., ASK-Tom, ASK-Michael, Advise the president, Trans-ASK, Engines of Education) are essentially CBR systems that consistently predict what a user might be interested in and retrieve cases for her/him to review.

A user starts using an ASK system by first “zooming” from the top-level structure of a domain to a specific case. This is done by identifying an area of interest, then choosing a theme from that area of interest, and subsequently choosing a case of interest from that theme. Afterwards, the user starts the “browsing” process in which they go from case to case in a manner similar to conversing with an expert. A user starts browsing with the first case he/she chooses at the end of the zooming process. The case summary and follow-up questions related to that case are presented in a case screen, and the user can either:

- Choose to view the full case
- Choose to follow up a question to go to another case screen

If he/she chooses to view the full case, a story in the form of a short video is presented. When the user finishes viewing or reading the first case, they are brought back to the case screen. He/she could then follow the links to another case screen and repeat the cycle.

For example, ASK-Tom [12] teaches novice bank consultants at Anderson Consulting about trust bank consulting. When ASK-Tom is started, it displays a big picture diagram of trust bank

**Original printed page 38 · source PDF page 48 · Open source page**

consulting with several topic areas. Clicking on the nodes or links in this diagram causes the screen to display either a more specific big picture diagram describing, for example, the temporal layout of a typical consulting engagement at a trust bank, or a screen showing a set of themes and stories that they organise. Big picture diagram screens ultimately lead to a theme screen. When the users reach the theme screen, they find several listings representing individual stories that correspond to various aspects of the theme. Once a choice is made, the system presents the story screen of the story chosen. The clip named in the centre of the screen is a link to the video clip of the main story. The questions linked to the main story are questions that the users might ask in a conversation with the expert following their previous conversation. These questions are predicted by the CBR system using a conversation model. Choosing one question linked to the current story brings the user to another story screen, and the cycle repeats.

The tree of all big picture models in the system form a conceptual map of the domain organised around its most important agents, relationships, and processes. These big picture models not only mediate the users’ choices of areas of interest, but also help the learners understand the structure of the domain and prompt them to see individual cases as illustrations of general principles. By using case-based reasoning algorithms, these case-based advisory systems are able to provide users with a coherent model of the subject matter. This coherent model, in turn, enables the users to assimilate the answers to their questions in a way that they can then use those answers to solve problems.

ASK systems use a story telling approach to instruction implied by the Dynamic Memory cognitive model. The ASK systems are designed with the assumption that expertise can be represented as a large, diverse bank of cases organised by a knowledge structure similar to the structure of the subject domain, and that novices could learn to become experts in a certain domain by becoming familiar with the experts’ case library. The ASK system is an example of a case-based advisory system.

**Original printed page 39 · source PDF page 49 · Open source page**

## 2.9 Case-Based Knowledge Sharing Systems

Case-based knowledge sharing systems are case-based advisory systems extended to augment users’ memory and to be augmented by users’ memory. Users’ experiences of reasoning with the support of case-based knowledge sharing systems are selectively collected and recorded as new cases for the systems’ case libraries. Such CBR systems are usually used as corporate memory. A case-based knowledge sharing system starts with a representative case library, and is constantly augmented by new cases to update and perfect its library. The idea is to gradually build a system that could act as a large-scale case-based reasoner that consists of the case-based knowledge sharing system and all of the users of the system. This large-scale case-based reasoner would solve problems by:

• Assessing the problem situation on the users' side using the case library of the system

• Retrieving similar cases from the case library either automatically or through special requests from users

• Interpreting the current case using past experiences and adapting available cases to solve new problems, usually done by the users with the support of the system

• Evaluating the solution and recording useful lessons as new cases, which is done by the users but recorded in the system

The valuable experiences of all reasoners who use the system would be incorporated into the large case-based reasoner's memory. In turn, this "memory of the community" would use the collective experiences and expertise of the system and the entire user community to support the reasoning of each reasoner.

In addition to all the benefits offered by case-based advisory systems, a case-based knowledge sharing system also offers a forum for building and sharing wisdom and insights derived from diverse sources. They encourage users to contribute their expertise through cases to a community of users within an organisation, thereby enabling the best practices of the community to be disseminated quickly. Since contributions from all users will not only serve as a source of information, but also serve as a source of inspiration and criticism, the collective experiences of members in the community can generate ideas that no individual would have developed alone,

**Original printed page 40 · source PDF page 50 · Open source page**

which leads to higher quality reasoning and more learning gains. In time, this will also result in a higher quality case library in the case-based knowledge sharing system.

Another feature that is often included in case-based knowledge sharing systems, but not case-based advisory systems, is the collaboration tools. The collaboration tools allow users to discuss certain cases or reasoning tasks, to annotate cases in the system, and to provide other forms of support to one another that compliment the kind of support the system provides. The communications, which carry these "other forms of support", are sometimes also recorded in the knowledge sharing system for later reference.

## 2.10 CBR Operation

### 2.10.1 Case Structure

The case structure for a given application is defined by the "advice" specifications used by the expert human advisors. For academic advising, the case structure is well known in advance since it relies on an existing set of student cases whose characteristics are known (This is contrary to a diagnosis application where one must make an intensive effort to "discover" the domain model, as in Hamilton & Gomes, National Semiconductor, in Watson 2003)

### 2.10.2 CBR components

The four components - retrieve, reuse, revise, and retain – are described in detail below. Each component is listed alongside the key issues involved in its implementation, as well as the unique knowledge necessary for that component's successful use.

The conventional set of CBR principles are more fully defined as a cycle comprising four activities, as shown in Figure 2-1:

• Retrieve similar cases to the problem description

• Reuse a solution suggested by a similar case

• Revise or adapt that solution to better fit the new problem if necessary

**Original printed page 41 · source PDF page 51 · Open source page**

• Retain the new solution

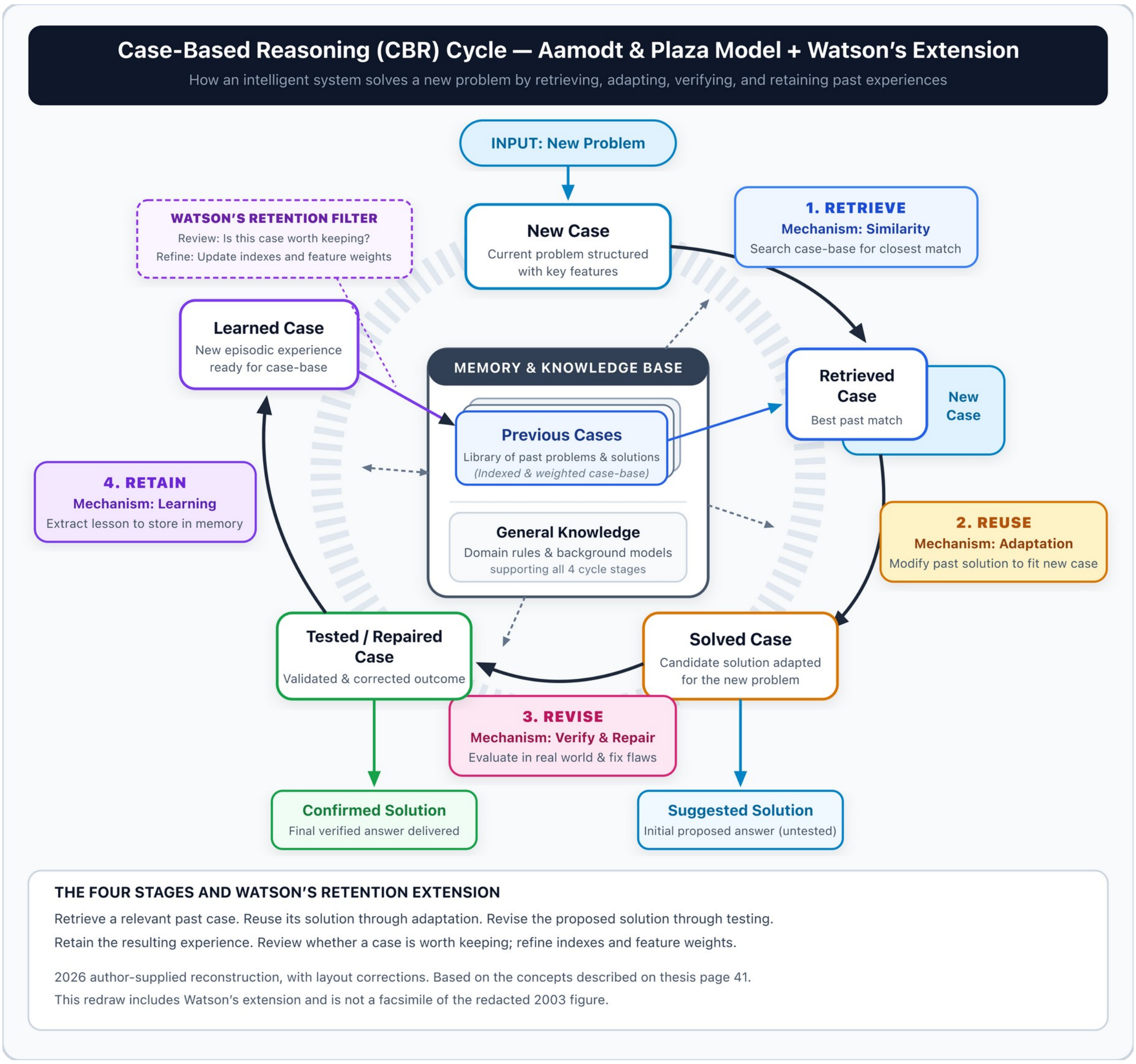

**2026 author-supplied reconstruction of the CBR cycle with Watson's extension; not the original 2003 figure.**

Figure 2-1: Aamodt & Plaza's CBR cycle [1]

Ian Watson proposed an improved version that contains an additional two activities:

- Review the new problem-solution pair to see if they are worth retaining as a new case
- Refine the case base index and feature weights as necessary

The case is retained in a case-base, but it typically undergoes some pre-processing and filtering (i.e., indexing and feature weighting). The reasoner retrieves the case to solve some problem and attempts to reuse it. This may result in some revision or adaptation of the case's solution, resulting in new episodic data and the creation of a new case. The new case is reviewed (this

**Original printed page 42 · source PDF page 52 · Open source page**

involves comparing to cases already in the case-base) and, if it is judged useful, it is retained. In addition, the use of the case (successfully or otherwise) may result in the case-base's indexing scheme or the feature weights being refined [44].

### 2.10.2.1 Retrieval

The retrieval process receives a description of the current problem as input. When first encountering a new problem, a "ballpark" solution needs to be located in the memory of past cases. This process assumes two factors: first, that a usable case exists somewhere in memory, and second, that once located, cases can be evaluated based on their similarity to the current problem. Once a case has been successfully retrieved, the next step is to apply the matching case to the new situation. This is the task of the reuse phase of case-based reasoning.

#### 2.10.2.2 Reuse

With the retrieved case, the system attempts to reuse the case in the current context. This requires it to identify any potential problems with the reuse of the plan. An evaluation sub component attempts to identify these potential problems so they may be repaired later. Several different approaches to evaluation have been used by CBR systems. These methods include: testing the proposed solution in the real world or a simulated world [17], asking a user for feedback [33], examining the outcomes of previous cases [23], or accessing a rule base that provides information about well known problems. If the evaluation identifies no relevant problems, the solution can be used and the new case is retrained. Otherwise, the solution is sent with a list of identified problems for revision. In practice, the boundaries between the reuse and revise phases are blurred until an acceptable solution is built.

#### 2.10.2.3 Revise

The revision component performs case adaptation as necessary to repair the problems identified during the reuse evaluation. However, case adaptation can be difficult. To repair a problem, the

**Original printed page 43 · source PDF page 53 · Open source page**

problem must first be understood. For example, a student failing an exam could have many potential causes: the student may not have studied, the exam was surprisingly difficult, or the teacher graded the student unfairly. In order to minimise the chance of the student failing in the future, it is necessary to assess blame [23] to a cause so that an appropriate solution can be generated.

However, it is non-trivial to assign blame for the problems identified by the evaluator. One solution could cause a host of other problems, some far more difficult to solve than the initial problem. In fact, some problems may require knowledge that is outside the scope of the original problem domain. These types of problems often require that a separate knowledge base be created exclusively for the use of the case adaptation module.

Due to these difficulties, case adaptation is often performed using static adaptation rules or by user intervention. Automated case adaptation has been considered by many to be an intractable problem. The problem is so acute that experts in both CBR research [23] and applications [4] agree that it is not currently practical to deploy CBR applications with automatic case adaptation.

As repairs are made, the case adaptation component passes the repaired solution back to the reuse phase to ensure that other negative side effects do not occur. If other problems are found, control is returned for additional revisions. This sub process continues until no additional problems are found. The final solution now can be used, and it is passed to the retain phase.

#### 2.10.2.4 Retain

The original four REs model did not provide a phase where an evaluation of the usefulness of the new solution was performed. Thus, we assume that all solutions reaching this retain phase are worth keeping. When the final solution is sent to be retained, it is ready to be placed in the case base. A key issue for retention is where to store the final solution in the case base. A well-organised case base can reduce the retrieval time and thus decrease the total overhead of the CBR process. The retention process provides the mechanism for the system to learn from the

**Original printed page 44 · source PDF page 54 · Open source page**

problems that have been solved from scratch, and provides the opportunity to reuse the solutions in future similar problem episodes.

### 2.10.3 CBR classification

A case consists of available information about a module, including its class and all measurements. The case library that represents the experiences of the development organisation is known as the fit data set. Each case in this library has known attributes and class membership. The new cases whose class membership is to be predicted are known as the target or test data set. Given a case with an unknown class, its class is predicted to be the same as the class of the most similar case in the library, where similarity is defined in terms of case attributes.

Let the case library or fit data set to be used be selected by the user.

The case library consists of the fit data set, having n modules. Let $c_j$ be the vector of independent variable values for case j, and let $c_{jk}$ be the value of the $k^{th}$ independent variable of case j. Similarly, a test data set is also selected by the user. Let $x_i$ be the vector of attributes of the $i^{th}$ unclassified module, and $x_{ik}$ be the $k^{th}$ attribute of that target module i.

The aim of academic advising CBR is to predict the dependent variable with minimal error. Since test data is strictly quantitative, similarity is thought of as the distance between cases. Consider each attribute to be a dimension of a multi-dimensional space. Each case is represented by a point in this space, and the unclassified module by another point. Similarity is thought of as the inverse of the distance between modules. Several similarity functions are provided to calculate a distance, $d_{ij}$, between the target module, $x_i$ and every case, $c_j$, in the case library.

### 2.10.4 Case Similarity Functions

For the purpose of explanation and similarity assessment, it is important for an “interpretive CBR” program to know not only that a similarity or difference is significant, but also why it is significant in terms of abstract knowledge.

**Original printed page 45 · source PDF page 55 · Open source page**

Although knowledge that a similarity or difference is significant can be represented by numerical weighting schemes, knowledge of why it is significant must be represented symbolically in terms of abstract knowledge. The symbolic representation should capture the general purposes and strategies for drawing analogies and distinctions among cases as a guide to decision making. A reasoner may, for instance, analogise a problem to positive case examples and contrast it with negative examples as a way of arguing that a problem is governed by a general principle, policy, or normative concern. Without this kind of abstract knowledge, a program cannot explain why a similarity or difference matters, and thus cannot adequately justify or explain its conclusions [2].

Several measures of similarity functions are described in the literature according to the problem domain, the availability of attribute data, and whether the data types are categorical, discrete, real, etc. [23] The Euclidean distance and City-Block distance [22] are commonly used for quantitative attributes. The Mahalanobis distance [10] has the advantage of explicitly accounting for correlations among attributes. The raw metrics (i.e., the independent variables) usually have vastly differing measurement units and highly varied ranges. Standardisation is a technique that converts all of the metrics to a uniform system of coordinates so they will all have the same unit of measure.

For each metric, $X_i$, the standardised metric, $Z_i$ is given by

$$Z_i = \frac{x_i - \acute{x}_i}{s_i}$$

Where $\bar{x}_i$ is the mean, and $s_i$ is the standard deviation of the $i^{th}$ metric, $x_i$.

All independent variables in the data set are standardised to a mean of zero and a variance of one. However, standardisation is not required when the Mahalanobis distance measure is used.

Original printed page 46 · source PDF page 56 · Open source page

#### 2.10.4.1 Absolute Difference

Absolute Difference is also known as “City block distance” or “Manhattan distance”. The distance is the weighted sum of the absolute value of the difference in the corresponding independent variables between a target module and a case in the library. The weights are provided by the user. Since the direction is not taken into consideration, the absolute value is used. The distance is given by the following equation:

$$d_{ij} = \sum_{k=1}^{m} w_k \left| c_{jk} - x_{ik} \right|$$

Where m is the number of independent variables and $w_k$ is the weight of the $k^{th}$ independent variable.

#### 2.10.4.2 Euclidean Distance

Euclidean distance considers each independent variable as a dimension of an m-dimensional space. A module is represented by a point in that space. The Euclidean distance between a fit module from the case library and a target module is their weighted distance in this space. The weights are those provided by the user. The distance is given by the following equation:

$$d_{ij} = \left( \sum_{k=1}^{m} \left( w_k \left( c_{jk} - x_{ik} \right) \right)^2 \right)^{1/2}$$

#### 2.10.4.3 Mahalanobis Distance

Mahalanobis distance is an alternative to Euclidean distance. Mahalanobis distance is a very good alternative to other distance measures when the independent variables are highly correlated and/or vary on vastly differing scales. The distance is given by the following equation:

$$d_{ij} = \left( c_j - x_i \right)' S^{-1} \left( c_j - x_i \right)$$

Original printed page 47 · source PDF page 57 · Open source page

S is the covariance matrix of the independent variables over the entire case library, and $S^{-1}$ is its inverse. S becomes the identity matrix and Mahalanobis becomes the Euclidean distance squared when the independent variables are uncorrelated and have a one unit variance. The independent variables do not need to be standardised or normalised when the Mahalanobis measure is used.

#### 2.10.4.4 Fuzzy Relations

**Table 2-2: PROFILE Relation**

| Fname | Lname | JobType | Expertise |
|---|---|---|---|
| Bob | McLedon | Academic | AI |
| Rob | Mucker | Industry | Expert System |
| Nancy | McCay | Government | Statistics |
| John | Hunt | Government | Robotics |

##### 2.10.4.4.1 Relational Data Models

A relational model organizes data into relations, which can be viewed as tables with meaningful names. The columns of this table are called attributes of the relation, while the rows are called tuples. For example, table 2-2 shows a relation whose attributes are Fname (First Name), Lname (Last Name), JobType, and Expertise. Conceptually, the relation describes the preferred careers and the expertise of job candidates in a job search agency’s database. The relation is named PROFILE.

To formally define a relation, we need to first introduce the notion of domains. A domain is a set of values. A domain can be discrete (e.g., the set {Academic, Industry, Government}) or continuous (e.g., the real numbers between 0 and 100). Each attribute in a relation has a domain, which specifies all possible values for the attribute. Let $D_1$, $D_2$, ..., $D_k$ be the domains of k attributes in a relation R. The Cartesian product of these domains $D_1$, $D_2$, ..., $D_k$ forms all

**Original printed page 48 · source PDF page 58 · Open source page**

possible combinations of values of the k attributes. The relation R, hence, is a subset of the Cartesian product $D_1$, $D_2$, ..., $D_k$. A tuple is a member in a relation.

Alternatively, a relation can be viewed as a set of mappings, each of which maps each domain to a value. The main advantage of this formalism is that changing the order of a relation's attributes does not change the relation. In other words, a relation is defined solely by the content of its table, independent of the order of its columns.

In the original relational model, the domains for each attribute have a set of predefined values. Using the similarity matrix approach, the imprecision is implicit in the similarity matrix and so the representation is still the same. Thus, this approach is very close to the conventional relational model. The possibility approach requires changing the data representation to a representation of fuzzy attribute values. For example, if we have an age attribute in a given database, then originally this attribute will have numeric values. Using possibility distributions, we will introduce terms like Young, Old, Middle-aged that are characterized by corresponding possibility distributions. (This could be applied to student's academic performance in Academic Advisement system, e.g., Excellent, Good, Fair, Poor.) In the next two subsections, a more detailed description of the two approaches will be presented. The table 2-3 below can be applied to the Specialization Characteristics, i.e., High specialization, Some specialization, Minimal specialization.

**Table 2-3: a similarity matrix for expertise areas**

| | Robotics | Expert Systems | AI | Statistics |
|---|---|---|---|---|
| Robotics | 1.0 | 0.6 | 0.6 | 0.2 |
| Expert Systems | 0.6 | 1.0 | 0.9 | 0.2 |
| AI | 0.6 | 0.9 | 1.0 | 0.2 |
| Statistics | 0.2 | 0.2 | 0.2 | 1.0 |

### 2.10.4.4.2 Similarity-based Fuzzy Relations

**Original printed page 49 · source PDF page 59 · Open source page**

In a similarity-based fuzzy relation, the impreciseness of attribute values is primarily in their meaning. For example, the values of the attribute EXPERTISE in the relation shown in PROFILE Relation often partially overlap. The overlapping degree of these areas of expertise is specified by a similarity matrix such as the one shown in table 2-3. In addition to imprecision due to semantics, a similarity-based fuzzy relation could also describe imprecision due to incomplete information. For example, we may wish to describe that the age of a suspect is 19 or 20. To do this, an attribute in a tuple is allowed to have a set of values, which describe all possible values for the attribute in the tuple. A similarity-based fuzzy relation involving domains $D_1$, $D_2$, ..., $D_k$ is thus a subset of the Cartesian product $2_1^D \, x \, 2_2^D \, x \ldots x \, 2_k^D$. Furthermore, as we discussed earlier, the values in domains $D_1$, $D_2$, ..., $D_k$ are related by a similarity relations $S_1$, $S_2$, ..., $S_k$ that map each pair of values in a domain to the interval [0,1]. A similarity measure of 1 means that the pair is identical, whereas a similarity measure of 0 means the pair is entirely different.

A similarity relation S for a domain D is a fuzzy relation that maps pairs of domain values in D to the unit interval, i.e., $S: D \times D \rightarrow [0,1]$, with the following three properties for all x, y, z in D:

1. Reflexive

$$S_D(x,x)=1$$

2. Symmetric

$$S_D(x,y)=S_D(y,x)$$

3. Transitive

$$S_D(x,z)\geq \max_{y\in D}\left(\min\left(S_D(x,y),S_D(y,z)\right)\right)$$

The first two properties are intuitive. The usefulness of the third property will become clear in the context of redundancy issue in a fuzzy database.

An example of the similarity relation is given in table 2-3 in a matrix form. The matrix describes the similarity within domain values of a given attribute. In the example, for the attribute Expertise, the domain values are {Robotics, Expert Systems, AI, Statistics}. For each pair of expertise areas, the similarity matrix describes one degree to which each expertise area is similar to the others.

**Original printed page 50 · source PDF page 60 · Open source page**

A similarity relation cannot be freely constructed. It has to satisfy the three properties listed above. In fact, whenever we specify two different similarity measures s(x,y) and S(y,z) , the similarity measure between x and z must be the smaller of S(x,y) and S(y,z).

#### 2.10.4.4.3 Possibility-based Fuzzy Relations

The possibility-based approach is, as the name implies, based on the possibility distribution theory. The possibility/necessity approach is more general than the similarity approach in the sense that it handles all types of information. The similarity based approach uses domains with discrete values. Even though an extension was added to it to handle the fuzzy number, the similarity approach still heavily depends on discrete values.

A possibility-based fuzzy relation generalizes a relation by allowing the values of an attribute A to be a possibility distribution $\Pi_{A(t)}$ of the attribute's domain. Let $D_1$, $D_2$, ..., $D_k$ be the attributes of a fuzzy relation R. A tuple in a possibility-based fuzzy relation is denoted ($v_1$, $v_2$, ..., $v_k$) where $v_1$, $v_2$, ..., $v_k$ are fuzzy subsets of $D_1$, $D_2$, ..., $D_k$. The interest of such an approach is its ability to represent, in a unified manner, precise values (singletons), NULL values, as well as fuzzy values.

It should be noted that since the data are imprecise, then the result of the query will also be imprecise. In theory, both possibility measures and necessity measures can be used for processing queries in fuzzy databases. In practice, however, the necessity measures are rarely used.

### 2.10.5 Case Acquisition

The cases for this thesis can be obtained from the existing student information database. As part of creating the case base, it needs to be checked for accuracy and coverage. Some of the data from the database may not be correct, so any “bad” cases that were created from incorrect data

**Original printed page 51 · source PDF page 61 · Open source page**

should be removed. It is recommended that the previous matches for all of the profiles that we would want to match in the future are migrated. At this stage, we may have to manually check for the accuracy by requesting help from expert advisors or devise an aid from rule-based reasoning. In academic advisement, it is impossible to define a set of rules that is very accurate in predicting the advice, but these rules can still point out the obviously correct cases. A set of test rules can be used to point out cases where the advice does not match the student profile within a set threshold, and then evaluate all of the cases with a large error by hand (or by expert advisors).

#### 2.10.6 Case Retrieval

This section describes a method to evaluate the quality of a specific matching algorithm. A selection process that uses this method to evaluate an algorithm can be used to find the algorithm that will reproduce sound advice and meet all of the desired attributes for the application of the specified advice. Nearest neighbour retrieval is used. However, the nearest neighbour must be determined by evaluating the degree of match in all of the attributes. This evaluation needs to provide a consistent meaning of an attribute's similarity throughout all attributes. Apart from the preference in fields of interest, the consistency is achieved using fuzzy linguistic terms such as Absolute, Partial, and Minimal (in Specialisation preference), and Difficult, Moderate, and Easy (in Course's Perceived Difficulty), which are associated with measured differences in an attribute. Any number of linguistic terms can be used. A fuzzy preference function is used to calculate the similarity of a single attribute of a case with the corresponding attribute of the subject.

The fuzzy preference function may be further improved by using the difference in the values of that attribute for the subject and comparable. For example, a difference of 1 unit would be considered excellent, 2 would be good, 3 would be fair, and 4 would be poor. This rating is then transformed into the fuzzy preference function.

The result of using fuzzy preference functions is a vector, called the fuzzy preference vector. The vector contains a fuzzy preference value for each attribute. The values in this vector can be

**Original printed page 52 · source PDF page 62 · Open source page**

combined, through weighted aggregation, to produce a robust similarity value. The use of fuzzy preference functions allows for smooth changes in the result when an attribute is changed, unlike the large changes that are possible when step functions are used.

A fuzzy preference function is used to transform a quantifiable value for each attribute into a qualitative description of the attribute that can be compared with the qualitative description of other attributes. A fuzzy preference function allows a comparison of properties that are based on entirely different scales, such as student's GPA and the rate of success in a course. Based on discussions with experts and work to classify previous matches into various sets of linguistic terms, it was found that there was enough precision in the evaluation of the similarity of the attributes to have four linguistic terms [44].

To assess the similarity of a problem case and a knowledge case, the similarities of all corresponding parameter values are calculated by applying local similarity functions to each pair of corresponding parameters. These local parameter similarities are then used to calculate an overall similarity value for the two cases. From three aggregated attributes (derived from seven raw measures) describing the cases, the preference in the fields of interest have complex similarity measures that represent a part of customisation to support a student's individuality.

The overall similarity is computed as the weighted average of the individual similarities. For our initial testing purpose, the weight factor in the aggregated preference in the fields of interest is a half, whereas the other attributes have a weight of a quarter each. The future improvement may allow students to adjust them according to their priorities. The local similarities for discrete and vector values are calculated in different ways. Discrete similarity measures can be defined according to the fuzzy linguistic terms above.

##### 2.10.6.1 Case Retrieval: Catalogue Search

Intelligent Catalogue Search can be incorporated into CBR in the Case Retrieval phase (or similarity matching).

**Original printed page 53 · source PDF page 63 · Open source page**

Two techniques that may be applicable:

1. Parametric Search is common and often combined with keyword or stepwise (hierarchical) searches to improve functionality. In this approach, the user enters one or more values for attributes of the product. The attributes of the parameters can generally be numeric or symbolic. A parametric search acts like a filter and is, in effect, the same as a database query - it meets all of the search criteria exactly or it fails. Some parametric search systems provide ranges and sets of values to overcome the limitations of the database approach, but “No match found!” is still a common occurrence. Some implementations of this approach use the product code and manufacturer’s code as the parameters, meaning that the user must know the product he/she wants before starting the search.

2. Another approach is the Nearest Neighbour method of searching, also called an Intelligent Parametric Search. This approach is a key output from CBR research and is the basis for the INRECA Catalogue Search applications (INRECA, 99). The characteristics of the problem domain that make “nearest neighbour” so effective are as follows:

- Student needs to search the knowledge base (a.k.a. product space) in its entirety
- Cases in the knowledge base will be similar in terms of their characteristics and their structure
- Student generally does not need an exact match as each student’s profile is quite individually unique - the optimum match will suffice
- Student needs the ability to negotiate with the search system
- No match found is an unacceptable outcome from a search

Human assessment of similarity or closeness is a normal thinking function and, in many cases, an automatic one. Given that other constraints such as academic performance and program structure do not hinder the request, an academic advisor interacting with a student in a real scenario will not take the student’s requirements as absolute unless specifically requested to do. The advisor will treat these requirements as a goal or ambition for the student, and will try to establish the student’s priorities before making a recommendation. This allows the student to

**Original printed page 54 · source PDF page 64 · Open source page**

give a precise statement of requirements, but not have it interpreted as such. The intelligent parametric search does exactly this, and uses weights to represent user priorities.

Retrieval is the essence of the CBR paradigm. Case retrieval consists of Identify Features, Initial Match, Search, and Select (Weber-Lee et al., 1996). All of these subtasks are equally important in order to perform an efficient case retrieval. The subtask Select chooses the solution for the input case. The Select step may combine the cases returned by Initial Match and produce the solution or simply choose the best match. Besides, when case retrieval is oriented by scores, one needs to set a threshold to limit the amount of cases to be returned. This threshold determines the domain of the Select subtask.

The selection of the best match may happen during the Initial Match step, but usually the Initial Match returns a set of cases. However, there are other types of retrieval that search for matching indexes, and the outcome of this retrieval can be of any size. The retrieval may yield either a score meaning the degree of match or assign a yes or no to determine whether the case matches sufficiently or not.

# Chapter 3 Academic Advisement

**Original printed page 55 · source PDF page 65 · Open source page**

In any academic program, one important staff role is to provide academic guidance to help students progress in their academic careers with a minimum of difficulty. Staff who assume this role are known as academic advisors, and usually have an in-depth knowledge of program structures, as well as access to individual student academic records. Based on this information, advisors can then provide useful academic advice to aid in a student's academic pathway selection or their future professional careers.

In the current information technology (IT) generation, many operations and processes have been automated by computers. Academic advisement has the potential to migrate to this kind of automation as well. In the author's opinion, a future where students can obtain highly intelligent academic advice over the Internet from comprehensive computer systems - without any human intervention - is not farfetched.

To reach this goal, we have to concentrate on how to build intelligence and smartness into computer systems. Human experts are able to perform at a high level because they know a great deal about their areas of expertise. This simple, seemingly obvious observation is the underlying rationale for the design of knowledge-based problem solvers. An expert system, as these programs are often called, uses domain specific knowledge to provide "expert quality" performance in a problem domain. Generally, expert system designers acquire this knowledge with the help of human domain experts, and the system emulates their methodology and performance. As with skilled humans, expert systems tend to be specialists, focusing on a narrow set of problems. Like humans, their knowledge is also both theoretical and practical: the human experts that provide the system's knowledge have generally augmented their own theoretical understanding of the problem domain with tricks, shortcuts, and heuristics for using the knowledge they have gained through problem-solving experience. Unlike a human, however, current programs cannot learn from their own experience: knowledge must be

**Original printed page 56 · source PDF page 66 · Open source page**

extracted from humans and encoded in a formal language. This is the major task facing designers of knowledge intensive problem solvers.

### 3.1 The Problem

One educational application that seems highly amenable to the use of an expert system is academic advisement. However, based upon a review of the literature published to date (February 2003), it appears that few expert academic advisement systems exist. Yet, this application is well within the scope of problems that expert systems are well suited to solve. No detailed published account of an expert advisement system, which had been submitted for a formal assessment of the quality of its output, was found.

### 3.2 Expert Advisor

An expert can be defined as an individual who is widely recognised as being able to solve a particular type of problem that most other people cannot solve as efficiently or effectively [18]. Academic advisors have been perceived to possess differing degrees of effectiveness [19, 24, 40]. Experienced faculty and staff with prior experience in advising have been regarded as better suited for this role than novices [28, 36]. Because different levels of competence exist within this domain, it seems plausible that the most competent of the advisors could be considered experts.

In contrast to some other domains, many aspects of expertise in academic advisement are related to a thorough knowledge of a specific institution. Although global advisement principles and practices exist [39], an academic advisor's expertise is somewhat limited to a local institution. A highly capable advisor in one setting could be incompetent in a different, autonomous institution due to unfamiliarity with the school mission, curriculum requirements, articulation agreements, institutional policies, and other related factors. Accordingly, the operational definition of an expert academic advisor in this study was a faculty or staff member who was recognised by local peers as being extraordinarily skilled in providing academic advisement to students.

**Original printed page 57 · source PDF page 67 · Open source page**

### 3.3 Existing advising practice

Many universities, especially in the United States, offer students one-to-one access to advisors; in Australia, this is usually not the case. Different approaches have been developed to open up different communication channels between advisors and students. Some provide advising via email and assign each student to a specific advisor. Others develop an advising pool with one or more advisors "on call" at any time to respond to students' email questions. In some places, advisors have established online chat rooms to provide opportunities for live or asynchronous discussions of students' questions.

#### 3.3.1 Current Web-based advising systems

Universities have recently started to use the Web to help in advising. Although some only use the Web to link to helpful sites or post the contact information of the advisors, an important trend for using the Web in academic advising does exist. This section shows the advising tasks some universities fulfil on the Web, and while they still have some limitations, these examples demonstrate that Web-based advising is very popular these days. Some examples to be reviewed are shown below.

North Dakota State University

(http://www.ndsu.nodak.edu/instruct/mcclean/ag_www/advise.htm#useful)

This is typical of the most common approach. North Dakota simply posts all requirements and relevant information on the Web. Students still have to make their own decision on what courses to take each semester. All general education and major requirements are defined clearly and concisely on the Web.

INSITE (Indiana Student Information Transaction Environment), Indiana University

http://insite.indiana.edu/

INSITE is a Web-based service for Indiana University students, faculty, and staff. It provides secure access to student information. For students, access to information is protected by a Student ID number (SIDN) and PIN. The following information is available through INSITE:

**Original printed page 58 · source PDF page 68 · Open source page**

- Produce an advising report for a student's current major
- Produce an advising report for a different major
- Produce an advising report for a special purpose program
- Produce an advising report for a program under development
- Produce an advising report for a special purpose program under development
- Generate a student transcript (course history)
- See how in-progress courses apply to a student's advising report

Student Planning System – Brigham Young University

http://ar.byu.edu/dept_academ_advise/sect3.htm

Academic Advisement at BYU combines the benefits of College Advisement Centres and other academic services including the use of the AIM system, Advisement by Computer (ABC) report, and faculty advising to provide direction and support to each student's academic progress. These resources are specifically geared towards addressing specific academic needs, and contribute to the quality of each student's overall university experience. Route Y is the BYU's campus intranet. It is designed to be a secure and personalised information source for current BYU patrons. Patrons include current BYU students, faculty, and staff. To help monitor students' general education and major progress, BYU has created the Advisement by Computer (ABC) report. Students can view and print their ABC report by using the Route Y AIM system.

"PLANNER WEB", California State University, Monterey Bay

http://csumb.edu/student/services/registration/

Planner Web for student advising is the first step in the Advising and Registration process for continuing and newly admitted CSUMB students. All undergraduates and second BA/BS candidates must submit their proposed schedule in order to register.

"Pre-major academic advising", Duke University

http://pmac.aas.duke.edu/

The website doubles as a question site and an appointment site. Students can access advising FAQs, and they can make appointments with their advisors.

**Original printed page 59 · source PDF page 69 · Open source page**

"Academic Advising Centre", Western Washington University

http://www.acadweb.wwu.edu/advising/gur_humanities.html

At West Washington University, academic advising is one of the most important services that students receive. The process of academic advising is a shared responsibility where a student works with faculty and staff academic advisers. Freshmen are assigned a faculty advisor at orientation who serves until a major is declared.

"Advising Central", North Carolina State University

http://www.ncsu.edu/advising_central/advisers/preprof.html

Advising Central at NC State is designed to provide Web and Internet-based advising to NC State undergraduate students. The goals of Advising Central are to:

- Make academic policies clear and meaningful for the students
- Help students navigate through NC State's human resources to locate and secure advice from the most knowledgeable person in a particular field or college, and help students identify sources that will aid them in clarifying their academic direction and strengthening their academic skills.

The above survey concluded that there are currently many universities with Web-based advising systems at work. Most of the pages entitled, Web-based advising, are typically a bulletin board with advising-related announcements, a repository of official documents in PDF or HTML format, a collection of useful links that help students get official advising related information off the Web, or a combination of these. They do not include any server-side scripts to process specific student information or produce customised advice. Indiana University is an exception, but the Web system is only for advisors to get their advisee's information - it is not designed for students.

### 3.4 Expert Systems and Academic Advisement

Educators have been comparatively slow in applying expert system technology to instructional and management problems in schools [20]. Nevertheless, excitement about these systems has

Original printed page 60 · source PDF page 70 · Open source page flourished in recent years, as it has in business and industry. The first reference to expert systems in the Current Index to Journals in Education as a separate category appeared in 1983. Other areas have been cited that are closely related to academic advisement, and have been demonstrated to be suitable domains for expert systems. For example, at least three expert systems have been developed for college admissions advisement [11, 13, 14]. The results from their experience indicated that transcript evaluation and admissions counselling were appropriate problem areas for expert systems. Edwards and Bader [11] concluded, "Expert systems have been shown to be of value in the university undergraduate-admissions process, by improving the consistency of decisions and reducing the workload of both the admissions tutor and the admissions assistant."

Another area closely related to academic advisement is the identification of "high-risk students" [41]. The Dallas Community College District developed an expert system in the student support area called the Computer Analysis System for Educational Success (CASES). The CASES system provided an important ancillary function for the academic advisor. CASES had the ability to analyse the records of all students and, based on a set of indicators, identify students who could benefit from a variety of student support services. Data was input from a scan-form and, in seconds, CASES printed a specific report on each student. Advisors were then able to utilise the research and recommendations provided by CASES to offer more intelligent guidance for the student. The developers indicated that they planned to incorporate articulation information and degree planning into CASES later.

Although this has shown that computers have been used to assist advisement for nearly 30 years, the use of expert systems in this specific domain is a unique and recent phenomenon. At the beginning of the last decade, there were only a few examples of the direct application of expert systems to academic advisement.

The earliest examples of expert academic advisement systems were published in 1984. The Graduate Course Adviser (GCA) was an expert system that simulated "a faculty adviser in suggesting the schedule of courses offered by the Department of Computer Science at Duke University" [43]. The GCA system apparently "knew" about computer science courses, degree

Original printed page 61 · source PDF page 71 · Open source page requirements, and departmental regulations at Duke. At the time of publication in 1984, the GCA had advised students on an experimental basis for two semesters. Technically, GCA was based upon MYCIN, written in the Prolog programming language, comprised 400 production rules, and was implemented in four modules. It is significant that the GCA researchers recognised that advisement was a complex task that involved both diagnostic and planning skills. A monolithic structure was inadequate to solve such a problem. Two of the GCA modules were required to determine the needs of the student, while the other two modules planned the student's schedule.

Another advisement tool was developed at Simon Fraser University in 1984 [8]. The Automated Academic Advisor (AAA) was described as "a large practical artificial intelligence system". AAA was designed to offer faculty, students, and academic advisors accurate information on degree requirements and course offerings. AAA was unique in that the performance of this system was to be evaluated by an automatic and interactive subsystem that would gather users' evaluation data as the application was running. The emphasis of the evaluation was placed on the system's general acceptability, rather than on a more traditional performance measurement criterion. The researchers concluded, "Much work remains to be done before a robust prototype can be tested by students, faculty, and academic advisors." This system appeared to be very promising, but no further reference to AAA can be found.

A later instance of an expert system used for academic advisement was described by Chan and Cochran [9]. They described a program called the Graduate Student Advisor (GSA) at Arizona State University. The GSA was used to help industrial engineering students get answers to routine questions about academic matters. The program was a rule-based system implemented in Personal Consultant, a pre-programmed expert system shell for personal computers. More than 100 rules were contained in the GSA. It was divided into four modules:

- Requirements for an M.S. or Ph.D. program
- Options for major fields of study and associated course and research structures
- Information on available courses
- Information on the faculty

**Original printed page 62 · source PDF page 72 ·** Open source page

In their article, Chan and Cochran argued for using expert system shells to implement such programs rather than using expert system languages like LISP or Prolog. They claimed the GSA not only helped graduate students, but also used the advisors' time more efficiently. "The system has proved successful in a university administration environment, and may benefit other organizations since the method is not necessarily limited to advising graduate students" [9].

Another example of an expert advisement system was created at the University of Arkansas to advise computer science majors [15]. This system, called Mentor-I, was an experimental prototype that contained expert knowledge embodied in rules. The rules were used to suggest the number of hours and courses to take to satisfy college degree requirements. Additional expertise was contained in Mentor-I in the form of suggestions for elective courses based on student interests and career goals. Like the earlier GCA system, Mentor-I was both a diagnostic and planning system and it was implemented in four modular components. Two diagnostic components were implemented in the Personal Consultant expert system shell. C-Prolog was used to create the other two planning modules. Although very few details were provided in the report by Frank, Duffield, and Swearingen [15], an evaluation of Mentor-I had yielded positive results. In two informal evaluations, Mentor-I's output was comparable to human advisors' advice in over 65% of the cases. At the time of their report in 1988, the authors had begun implementing a second system (Mentor-II), which would be generalised for the entire college.

Several expert systems were applied to tasks closely related to academic advisement, including the selection of college admission candidates, transcript evaluation, and the identification of high-risk students.

### 3.5 Degree Audits

Simply stated, an automated degree audit matches completed coursework with sets of degree-program requirements, tracking student academic progress from the declaration of a major to the completion of a degree. In addition to assessing the progress of completed coursework, the system applies various university policies and regulations – such as those related to residency, course repetition, and pass-fail – in an automated fashion, negating the need for advisors to

**Original printed page 63 · source PDF page 73 ·** Open source page intervene manually. Furthermore, some degree-audit systems equate transfer credit to local institution courses and provide for exception record data (course substitutions, waivers, and individualised requirements).

Automated degree audits have become prevalent in colleges and universities in the past decade, but they had their beginning in the early 1960s with the "Bingo Sheet" created at Purdue University. In 1968, Georgia State University introduced its Programmed Academic Curriculum Evaluation (PACE) degree audit, followed by North Carolina State's table-driven system in the early 1970s. Brigham Young University released its Advising by Computer reporting system in 1978, and Miami University (Ohio) implemented its Degree Audit Reporting System in 1984 [30]. Other institutions that subsequently developed degree-audit systems on their respective campuses include Southern Methodist University, Texas A & M University, and Clemson University.

Because of the potential profitability of degree-audit systems, several software firms entered the marketplace, some with a degree-audit module incorporated within a student information software package. Among these corporate vendors are Campus America, CARS Information Systems, Computer Management & Development Services, Computing Options Company, Datatel, Degree Navigator, Iron Soft, Peoplesoft, Quodata Corporation, Software Research Northwest, and SCT Corporation [30].

According to the most recent ACT survey on academic advising, 48% of faculty advisors among departmentally based advising units are routinely provided with computerised degree audits as informational reference materials. Similarly, 31% of advisors of centralised advising units are routinely offered degree audits as an academic reference [17].

Because the academic advisor is frequently the university official who interprets complex academic requirements, enforces a multiplicity of academic policies and regulations, and communicates changes to academic programs, it is crucial that the degree-audit system be capable of executing many diverse transactions. Using the degree-audit system enables institutions to achieve many vital and important objectives.

**Original printed page 64 · source PDF page 74 · Open source page**

First and foremost, a properly developed and implemented audit system can significantly reduce the amount of advisor time expended in the clerical duties typically associated with the advising enterprise. The degree audit accurately reports student progress towards the completion of degree requirements, and does so comprehensively, consistently, and in a timely manner. These systems provide for instant feedback from clientele, whether the users are students, faculty, administrators, or support personnel. Degree audits enhance advisor morale and effectiveness by reducing the likelihood of litigation typically associated with advising, and by virtually eliminating perfunctory and redundant record keeping chores. In addition, degree audits can provide a comfortable academic environment for students by clearly and concisely reporting their progress toward graduation, proper course choices, and minimum grade point average stipulations.

Some degree-audit systems have the capacity to support tasks peripheral to typical advising functions. In some institutions, degree-audit systems assist with the management of course offerings, monitor “four-year guarantee” compacts, evaluate academic honours (cum laude, summa cum laude, magna cum laude, Phi Beta Kappa), and execute the complex National Collegiate Athletic Association (NCAA) Division 1 and 2 Continuing Academic Eligibility Compliance Checks. Many degree-audit systems improve the accuracy of some transcript information, especially with cross-listed courses and converted courses (for which the prefix or number has been changed), as well as course repetition policies.

Finally, these systems provide a fiscally prudent way to accomplish vital and necessary advising tasks. Reducing the labour-intensive aspects of the advising function by using an effective and efficient degree-audit system demonstrates fiscal responsibility in an era of constrained budgets and downsized operations [30].

In order for degree audits to be of paramount support for academic advising, the systems must posses a large number of diversified features. These features, when invoked with little or no manual intervention, save students and advisors time in monitoring academic progress and interpreting academic policies. One mandatory feature permits students to shop for alternative

**Original printed page 65 · source PDF page 75 · Open source page**

curricular choices. This “major-shopping” feature particularly assists students who have not yet decided on an educational program, as well as those who wish to supplement a major with a minor, option, or a specialisation area.

The audit should be available to clients in both online and batch modes, and it should reflect real-time data. Audits produced via batch runs and distributed to students later may not reflect data that has changed since the initial run. Thus, it is imperative that an online system providing up-to-the-minute information be available to the advisor.

Degree audits assess degree program progress by evaluating courses according to the number of courses completed, the number of credit hours completed, the minimum grade earned, the minimum grade point average achieved, and the best grades earned from a group of courses used to satisfy a requirement. The system must correctly identify cross-listed courses, repeated courses, repeatable courses, and converted courses, and apply them appropriately to each requirement without manual intervention by the advisor or a school official.

All components of a degree program must be clearly identified and monitored correctly. University requirements (such as minimum credit hours, minimum grade point average, and residency policies), as well as general education, upper or lower-division, college- or department-specific, professional school (licensure), and major and minor requirements must be evaluated collectively and correctly.

A critical feature of a computer assisted advising degree-audit system is the treatment of individual exceptions such as courses substitutions, course or credit-hour waivers, and so forth. Caution should be exercised when making exceptions and exception-tracking procedures must be in place to ensure the academic integrity of the report.

Few individuals other than advisors recognise that two types of course substitution exceptions – global and individual – exist. A global course substitution treats the substituted course in the same manner every time the course appears on the audit. An individual course substitution

**Original printed page 66 · source PDF page 76 · Open source page**

permits a course to be substituted only once. Security is the primary issue with exceptions, and external as well as internal auditing of exception entries is necessary.

By providing accurate and comprehensive curricular information, the degree audit supports academic advising. It promotes a feeling of ownership in the student and requires the student to engage in decision-making activities. It relieves the advisor of record keeping duties so that appropriate time can be expended assisting the student with the developmental aspects of academic advising.

### 3.5.1 PeopleSoft Academic Advisement

Academic Advisement is the application within PeopleSoft Student Administration that is used to track the requirements and policies that a student must satisfy in order to graduate. As a student progresses towards graduation, Academic Advisement analyses those courses completed by the student - both successfully and unsuccessfully - and ascertains what requirements are still outstanding. Using data from student records and requirements, the application automatically tracks a student's degree progress. Once the requirements are entered into the system, the student's data is analysed against the requirements to report the degree progress. What-if scenarios can also be performed for the student to see what courses they might need to complete for a particular major.

### 3.6 Technology in the Self-Help Advising Age

In the Information and Learning Ages, we have been discovering new ways to access information and services. In a sense, technology is providing a tool that fosters the developmental advising process and promotes students’ responsibility for their academic careers. Through technology, academic advising information and services are readily available at any time, any pace, and any place. Students are becoming better informed about basic academic information, and thus advisors have the opportunity to provide more detailed, in-depth advising when they meet face-to-face with students.

Original printed page 67 · source PDF page 77 · Open source page

### 3.6.1 World Wide Web

The most active and prolific use of technology to deliver academic advising is through the World Wide Web (WWW). Some features are listed as below.

#### 3.6.1.1 Basic Services – Text Information

Basic text information includes:

- Policies and procedures
- Frequently asked questions
- Address, building location, room number
- Phone numbers
- Hours of operation
- Name and/or e-mail contact
- Links to other resources
- Assessment or practice tests

#### 3.6.1.2 Midlevel Services

Mid-level services include:

- Online forms – applications and requests for information
- Search capability
- Order forms (such as for textbooks)
- Audio and/or video

#### 3.6.1.3 Advance-Level Services

Advance-level services include:

- Web registration
- Degree audit
- Schedule planner

Original printed page 68 · source PDF page 78 · Open source page

- Articulation and transfer evaluation
- Access to student records and profiles
- Grade point average calculator
- Interactivity with student profile
- Artificial intelligence

Advising through the WWW is self-help advising. Students may interact with databases and their personal profiles, but there is no one-to-one interaction. Most higher education institutions have WWW home pages that provide at least basic information about the institution. College catalogues, schedules of classes, policies and procedures, addresses, phone numbers, and so on are readily available at the click of a mouse. Institutions that have the resources are developing interactive web pages. Online applications, request forms, and the capability to search for information are appearing on more college websites. The most sophisticated college web pages provide online registration, degree audits, electronic evaluation for articulation, schedule planners, grade point average calculators, and artificial intelligence, all based on a student's personal profile.

Anyone who has been on the Web has found a plethora of easily accessible and readily available information. Yet, many have also gotten lost as they wander around the Web. Looking for information and not finding it can be time consuming and frustrating. To make the search for academic advising information and services more convenient, colleges are creating virtual advising centres. A virtual advising centre is a central point of electronic access for advising services and information. Web links both within and outside the college are organised so that students and advisors can quickly find information about student services.

Virtual kiosks are another way of providing academic advising information. Virtual kiosks are touch-screen information centres that integrate access to the Web with text or multimedia information stored on the local computer's hard drive. Virtual kiosks are centrally located where students gather, such as campus student centres. Sophisticated virtual kiosks and virtual advising centres enable interactivity with the college's databases, allowing students to register for classes, calculate degree audits, and change personal information.

**Original printed page 69 · source PDF page 79 · Open source page**

### 3.6.2 Advising over the Internet

In addition to obtaining advising services over the Web, students can also find listservs, e-mail newsletters, and electronic bulletin boards on the Internet. Listservs and e-mail newsletters are accessed via e-mail accounts and do not require web browsers. Advisors can mass distribute information about upcoming events, calendar deadlines, and other timely titbits. E-mail newsletters can provide students with helpful articles, including study hints, note-taking guides, and test tips. Listservs provide students with opportunities for interchange with other students and college staff. Lively discussions on hot topics can provide a sense of group connectedness. Electronic bulletin boards provide students with a means of viewing announcements and posting messages. Examples of the innovative use of advising on the Internet can be found at Academic Advising Resources on the Internet (http://www.psu.edu/dus/ncta/linkacad.htm).

## 3.7 Internet-based expert system for selecting an academic major

In the late 1990s, Fritz H. Grupe [16] developed an Internet-based expert system (http://www.MyMajors.com/), which provides advice to high school students or college freshman seeking assistance in selecting a potential major. It emulates a professional academic advisor. The on-demand, approximately 15-minute consultation gathers information from the student on his or her grades, degree of enjoyment of traditional courses, standardised test scores, interests, and aptitudes. It assesses student qualifications for a variety of majors. The expert system recommends six majors from among 60 widely diverse majors for the students to consider, and produces a report that fully describes the students' responses in such a way that the output can be used by a human advisor to further extend and strengthen the advisement process. However, this system was not designed to advise on the list of subjects that students should enrol in, and the system has been implemented using Rule-Based Reasoning as the main expert system technique.

# Chapter 4 Applying CBR to automate Academic Advisement

**Original printed page 70 · source PDF page 80 · Open source page**

## 4.1 Needs for Automated Academic Advising

### 4.1.1 The Role of Technology in Higher Education

Technology has clearly been the primary transforming factor as we move into a post-industrial society. In industry, technology accounts for much of the impetus to downsize firms, the popularity of outsourcing, the surge in value-adding partnerships, and for the flattening of hierarchies [33]. Modern information technology has dramatically changed the way scholars and students access information in higher education, and it is beginning to alter current forms of teaching and learning. Faculty and students were traditionally bound to the physical space of universities because libraries were the central depositories of information. Information technology now allows universities to shift from investing in the physical storage of information to creating electronic databases. Ready access to discipline-based information through electronic means increasingly frees scholars and students from the limitations of physical space and time. Faculty can now easily interact with their colleagues at distant places and pursue their research in a global environment.

In the future, technology will also dramatically change the delivery of instruction and the way students learn. Instead of coming to a campus, students will have access to a wide variety of electronic curricula. They will be able to communicate electronically with their tutors and learn by utilising interactive reading materials and study exercises [29, 31]. Distance learners will be credentialed through the competency-based certification of degrees. For-profit organisations are expected to take the lead in providing electronic educational programs, and this will increase the latitude available to students as they choose their educational paths. Electronic curricula will be

**Original printed page 71 · source PDF page 81 · Open source page**

especially appealing to the growing number of students who, because of work-related factors or family commitments, cannot pursue a traditional, campus-based education [31].

Cost reduction is one of the primary reasons that traditional and for-profit institutions of higher education are interested in moving into the electronic education market. After the initial investment in hardware and software and the creation of electronic curricula, courses and programs are accessible to thousands of students around the world for relatively small further investments.

### 4.1.2 Strategies for Addressing Resource Constraints

Over the next decade, higher education institutions will continue to experience declining resources, making staff additions in advising offices unlikely. At the same time, undergraduate enrolments are expected to increase. As a result, advisors will be called on to serve larger numbers of students with existing or, in some cases, even reduced resources. The obvious challenge is to advise more students with no additional staff while maintaining high quality services. Reconciling these seemingly contradictory goals will require advisors to seriously examine their traditional ways of doing business – the current advising structure and practices – and identify strategies for increasing efficiency. These demands closely resemble the recent experience of Australian businesses that have had to reinvent themselves through rethinking and restructuring their business practices in order to function effectively in changed circumstances.

### 4.1.3 Non-traditional Access to Advising Services

Diversity with respect to race, ethnicity, age, and purpose makes it essential for advisors to design programs and services directed to the differing needs of their clientele. They must also consider their availability to provide that advice. Many students with full-time jobs can take classes only in the evening and advising offices are often closed at that time. In addition, with the expansion of coursework and degree programs that are delivered electronically, an increasing proportion of students will rarely set foot on a campus. Those students will come to expect that their advising needs will also be met in non-traditional ways.

**Original printed page 72 · source PDF page 82 · Open source page**

Current practices in business and health care provide some idea of how these new advising needs might be met. Interactive electronic conference sessions are already employed in industry because they are convenient and cost effective. For example, nursing services to elderly outpatients are delivered in some locations through electronic interactivity, allowing both parties to see each other on a monitor. This technology can increase the contact between nurses and patients, as well as increase the number of patients served because of the substantial savings in travel time. In turn, the elderly are able to continue living independently, which many want to do as long as possible [21].

The same technology will allow universities to provide electronic advising services. Even now, students on many campuses have ready access to their academic records, automated degree-audit reports, and web pages that provide important academic information. These provide the high-tech, but not the "high-touch" aspect that comes with individual guidance by an academic advisor. However, it is also true that advisors increasingly use e-mail to assist students who cannot come to campus. Interactive electronic advising sessions in which advisors sit in their offices and can see their distant students on a video screen are well within our current technical capacity, and will be conducive to an educational environment where a substantial number of students find it inconvenient or impossible to come to a campus. The marriage of high-tech and high-touch may provide the setting for enhanced efficiency without the loss of the human relationships and expert guidance that is important for students who present complex educational issues. For example, Rio Hondo Community College in Whittier, California (http://www.riohondo.edu/students/counseling/) has already pioneered a video advising program for high school students in the Southern Los Angeles area. College counsellors use interactive video technology to advise high school students in real-time, answering their questions about admission, registration, academic programs, and careers.

**Original printed page 73 · source PDF page 83 · Open source page**

#### 4.1.3.1 Subjective needs of students

If students can be thought of as customers of a technical support service [21], student requirements are relatively clear:

- Students want advice according to their academic profile and preference
- If students ask to consult an academic advisor or a CBR system counterpart, either directly or indirectly, they expect advice. If students do not get sound advice based on their circumstances, they get an impression of incompetence, either of the advisor or the CBR system
- Students want satisfactory advice for their academic profile and preference
- Obviously, students are additionally interested in getting sound or satisfying advice. It might not be necessary to provide the most satisfying advice, but it is essential to offer satisfying or relatively suitable advice
- Students want fast advice to their academic profile and preference
- During enrolment/re-enrolment, students are impatient and want advice fast. Although some student cases may be deemed higher priority (e.g., international students may get deported if they fail to fulfil the Department of Immigration and Multicultural & Indigenous Affairs (DIMIA) requirement of full-time study load), students perceive their own case as the most important and urgent case
- Students want confident advices to their academic profile and preference

- In addition to satisfying and fast advice, students often expect the advisor has a high level of confidence in their given advice. They insist that the first advice is suitable. Alternatively, if the advisor is not able to provide satisfying advice, students want to be confident that the advisors are in no doubt of their decisions

**Original printed page 74 · source PDF page 84 · Open source page**

### 4.2 RMIT and Different Perspectives Approach

At RMIT, the term “program” or, more specifically, “academic program” refers to a program of study leading to the award of a degree or certificate. The term “course” refers to a unit of study, often called in Australia a “subject” or “unit”. RMIT grade systems classify five grades, in order of highest achievement, as follows: High Distinction (HD), Distinction (DI), Credit (CR), Pass (PA), and Fail (NN).

In the School of Electrical and Computer Engineering (ECE) at RMIT, the multi-major program concurrently consists of five engineering majors: Communications (CO), Computer Systems (CSE), Electronics (EL), Electrical (EE), and Software Systems Engineering (SSE). The students study a common 1st year and start branching out via their choices of electives from the 2nd year onwards.

#### 4.2.1 A review of program structure

Courses are placed into blocks, depending on their fields and whether they are compulsory or elective. This block structure enables they easy tracking of student progress towards the fulfilment of program requirements and program completion.

The normal load for each year is 96 credit points for full-time students. Given that most courses carry 12 credit points, this equates to eight courses per year or four courses per semester under normal circumstances. Overloading courses must be approved by authoritative program coordinators (which is essentially an RMIT role for academic advisors).

**Original printed page 75 · source PDF page 85 · Open source page**

##### 4.2.1.1 1st year

Every student shares a common 1st year structure and every course is compulsory:

- Core engineering foundation courses
- Mathematics
- Physics
- Design course

##### 4.2.1.2 2nd year

Students begin to narrow towards a specialisation by selecting compulsory electives from two streams:

- Design course
- Mathematics
- Core engineering foundation courses
- Compulsory technical electives (chosen from two out of four streams)
- Free elective

##### 4.2.1.3 3rd year

Students decide on which of the two streams will be their major for the degree outcome. From this year onwards, specialisation within majors becomes possible:

- Professional practice courses

- Core technical courses (based on the stream of preference)
- Technical electives (from other streams, effectively a minor if chosen wisely from the same stream)
- Free elective

**Original printed page 76 · source PDF page 86 · Open source page**

#### 4.2.1.4 4th year

Most courses in this year will be electives from within their major, and students can choose to specialise in one area or to spread their electives across other relevant fields to gain broader knowledge, rather than depth:

- Vacation employment
- Design courses
- Stream technical electives
- Business elective
- Other stream technical elective
- Free elective

### 4.2.2 Governing Rules (applicable to all courses)

There are four governing rules as follows:

Students should finish within a minimum time of four years

Normal load per year is 96 credit points, or four courses per semester. Overloading or over-enrolling requires approval from the relevant authority

Students must fulfil corresponding prerequisites before being permitted to enrol in each course

Failed compulsory courses must be repeated and have the higher priority than other courses (compulsory or elective)

## 4.3 Aiding in electives selection

Choosing electives can be daunting because the decision making process is often based on many influencing factors. This is the characteristic of the problem that is most suitable for CBR. As a proof of concept, CBR will be applied in the prototype system to recommend elective selections based on some influential factors. Please note that the elective structure used is a “cut down” version and does not reflect the full complexity of the actual program structure.

**Original printed page 77 · source PDF page 87 · Open source page**

### 4.3.1 Why select electives?

According to http://web.mit.edu/catalogue/ch3-major.html:

“Electives can be used for several different purposes. For example, many students who are undecided about their eventual majors will use some part of their elective time to get more information about the various departments or fields they are considering. Students more certain of their professional goals will use electives to explore areas of secondary interest. Still other students will choose to begin work on departmental or General Institute Requirements, deferring subjects of a more supplemental nature until a later year.”

According to http://www.umass.edu/larp/AcademicPrograms/mlaelectives.html:

“Some students may wish to select electives in several of these areas to broaden their education. Others may wish to take several of these electives in one of the concentrations, to develop a special strength in a given concentration.”

The CBR system can be most useful when aiding students to select their final year electives or free electives. As per stream electives, the system may be more complex since a hybrid model between RBR and CBR may be required.

**Original printed page 78 · source PDF page 88 · Open source page**

#### 4.3.1.1 Decision Influences

The following factors usually affect a student’s individual decision when selecting electives:

- Personal interest (e.g., fields of expertise/specialisation/concentration)
- Peer preference, popularity (e.g., can be measured by the number or percentage of students enrolled in any given year)
- Perceived degree of difficulty (e.g., word of mouth from peers or the program coordinator). This can also be measured by either the pass rate or chance of getting a good grade (course may be easier to pass but not to get HD). Thus, this degree will vary according to a student’s individual profile
- Relevance to the current job market or industry demand
- Reputation of the lecturers - teaching and in-charge (e.g., expertise in the field of the course, past history in teaching/assessment difficulty, presentation skills)
- Type of assessment (e.g., no exam?)
- Student’s personality (e.g., expert in a particular field, or an all-rounder with broad general knowledge in many fields)

#### 4.3.1.2 Factors Quantification

To construct a test case structure to reflect the above factors, each factor can be quantified as follows:

- Personal interest: DSP (1), Mobile Comm. (2), Power (3), µController (4), Internetworking (5), Software (6)
- Popularity: this can be calculated from either the figures of the previous year or a pre-re-enrolment survey (which students must complete before the re-enrolment period).
- Both the pass rate and High Distinction (HD) rate should be calculated from the previous year and stored as two separate sub-attributes.
- In collaboration with the Careers and Employment Service, job market trends should be researched and manually entered before the re-enrolment period. The attribute values

**Original printed page 79 · source PDF page 89 · Open source page**

should be the same as or similar to those of “Personal interest”, with associated ranking based on industry demand.

- Reputation of lecturers: sub-attributes can be grouped as either a factual (calculated or obtained from other databases) or an opinionated (derived from student survey results in previous years) type. An example of factual sub-attributes is the number of year since teaching the first class or a particular course. Examples of opinionated sub-attributes are field expertise, presentation skills, and tendency towards difficult assessments.
- Type of assessment: have exam/no exam (E1/E0), number of assignments (e.g., A= 0 – 4).

- Individual personality: student prefers to have broadened general knowledge in diversified fields, or to become an expert in a specific field (broad/expert).

One option, at the re-enrolment session, is for the Conversational CBR (CCBR) system to ask students a set of questions to determine the values of these attributes. These values will form a problem case. In turn, this problem case will be matched with the CBR knowledgebase, in which the same indices and data structure are implemented. “Index Structures for Data Warehouses” by Marcus Jürgens, LNCS 1859, provides a good pointer for this process.

### 4.4 Indexing Cases

The INRECA developers found that a flat structure is sufficient to meet the search needs of most end-users. By flat, it means that all attributes of the student case of interest are used to describe every student case in the knowledge base. The alternative to a flat structure is an object structure. For an academic advice case, an object structure might see its attributes as fields of [interest, academic performance, or specialisation preference], where the fields of interest would contain the preference order of [DSP, fibre optics, microwave, networking, satellite, etc.], and academic performance would contain (GPA, WAM, peer ranking, etc.), while a flat structure would contain [DSP, fibre optics, microwave, networking, satellite, GPA, WAM, peer ranking, etc.], all at the same level. The object model allows us to describe a structure for a more complex advice case representation. However, an object model structure may require a more complex interface for the student and this, unless necessary, should be avoided. A flat

**Original printed page 80 · source PDF page 90 · Open source page**

representation can simplify the student’s view of the catalogue search and, in most cases, give the same level of functionality as the object representation. A flat model can also be stored easily in a standard database system, of which many are available.

In order for the case-based reasoner to retrieve “similar” cases, labels (known as indexes) must be assigned to a case to “identify” it as an appropriate case to be retrieved given a particular student’s circumstances. Hence, indexes signify which cases are relevant (i.e., useful) and are potentially able to provide academic advice to a student case. Deciding on the appropriate index requires a good understanding of the academic advisement domain and the advice-giving goal. Since indexes represent an interpretation of a student’s circumstances, a good index (appropriate for the advice-giving situation) would allow relevant cases (with potential solutions) to be retrieved quickly and accurately. However, choosing a good index may not be so simple since there are no guidelines or established methods that exist for choosing good indexes.

#### 4.4.1 Four qualities of good indexes

As a guide, there are four qualities of good indexes (Kolodner, 1993) described below.

##### 4.4.1.1 Predictive features

Indexes should address the advice-giving goal of the cases. In the author’s prototype system, the chosen indexes are fields of interest (INT), specialisation characteristics (CHRT), and previous academic performance (PRFM). At the most basic level, these three indexes describe a student’s circumstances.

##### 4.4.1.2 Abstractness of indexes

Indexes should be abstract enough so that a case can be used in a variety of future situations. As a proof of concept in the prototype system, the three indexes (INT, CHRT, PRFM) have been chosen so that the case can be used in a broad and general collection of student circumstances. Indexing the case according to an abstract description that captures what is important about the

**Original printed page 81 · source PDF page 91 · Open source page**

case allows a retriever to focus immediately on the features that point out its applicability. To contrast this feature, take the PRFM index for example. The author could have used a particular GPA or WAM (Weighted Average Marks) calculation instead. This would have constrained the determination of “academic performance” to one aspect without taking into account other influential factors. Choosing PRFM allows the calculation of academic performance to include these so-called influential factors (such as course loads in each semester). These indexes or descriptors point out the range of applicability of the case and make the case generally applicable.

#### 4.4.1.3 Concreteness of indexes

Indexes should be obvious and recognisable in future situations without further interpretation. The INT index contains sub-indexes of DSP, OPTIC, MWAVE, NTWRK, and SATEL as fields of interest. These fields can be recognised immediately by students. The specialisation characteristic (CHRT) assumes three values: absolute specialisation, some specialisation, and minimal specialisation (broad generalisation). CHRT is also recognisable to students with little inference. PRFM is obtained from a uniform algorithm, so it is also concrete.

#### 4.4.1.4 Usefulness of indexes

Indexes should be able to differentiate cases in useful ways and be able to give guidance about the decisions that the case-based reasoner deals with. INT, CHRT, and PRFM have also been chosen to make the kinds of predictions that will be useful in later reasoning. However, we can only determine this after the actual implementation of the system. Then, we can evaluate the usefulness of the indexes we have chosen.

In brief, the indexing problem is “the problem of retrieving applicable cases at appropriate times – problem of assigning labels (called indexes) to cases that designate under what conditions each case can be used to make useful inferences” [23].

**Original printed page 82 · source PDF page 92 · Open source page**

### 4.4.2 Matching Calculation

#### 4.4.2.1 Defining Similarity

The assessment of the similarity between one advice case and another, and between one value for a parameter and another, is always likely to be subjective. The expert’s view of similarity will differ from the student’s and, as a result, the effectiveness of similarity calculations is best seen in the appropriateness of the results. In each of the intelligent catalogue search applications developed to date, it was necessary to go through a process of experimentation with the similarity functions (INRECA, 99). In other words, standard (simple) similarities did not work, and a set of tailored similarity functions had to be defined.

By prioritising similarity during retrieval, a standard case-based approach will implicitly ignore the importance of retrieval diversity and this may reduce the quality of the retrieval results. It is often not good enough for a recommender (an academic advisor) to return only the most similar cases. It should also return a diverse set of cases in order to provide the student with optimal coverage of the information space in the vicinity of their query (Smyth & McClave, 2001).

A degree of benefit can be expected when a system lacks any knowledge of a customer’s preferences - that is, when the system does not perform any learning. An important variable in determining the importance of customer feature-weight accuracy is the actual distribution of each customer’s feature weights. Intuitively, one would expect a minority of features to be of high importance, and a larger number of features to be less important (e.g., in academic advising, field interest may be very important, whereas peer preference is less important, though not inconsequential) (Branting, 2001).

This distribution of feature importance and four alternatives can be represented by the following n-dimensional weight vectors:

Two-Level. For each feature, there is 25% probability that the feature weight is in the interval [0.75..1.0] and a 75% probability that it is in the interval [0.0..0.25].

Exponential. For 0 <= i < n, exactly one feature has weight

$$\sqrt{\frac{0.75}{2^{i}}}.$$

Linear. For 0 <= i <= n-1, exactly one feature has weight

$$1-\frac{1+i}{1+n}.$$

Random. Feature weights are uniformly distributed on the [0..1] interval.

Equal. All features are weighted equally.

Consider courses A, B, C, with the following values for the study field:

A: {DSP}

B: { DSP, Optics, Networking }

C: { Optics }

This asymmetric overlap metric looks at the size of the feature set intersection relative to the size of the set in the query case. For example, the similarity between A and B is maximal because B overlaps with everything found in A. The similarity between B and A is less because A only contains 1/3 of what is found in B. This metric is intentionally asymmetric under the assumption that a student looking for DSP knowledge would be satisfied by a course that covered both DSP and Optics, but one looking for a course that combined Microwave and Satellite materials would not necessarily be happy with a course that focused on one or the other (Burke, 2001).

#### 4.4.2.2 Early Attempts

##### 4.4.2.2.1 How to determine overall matching percentage

The database system should maintain a catalogue of what courses are contained in each field of study (COURSES_IN_FIELD_OF_STUDY table).

There should be a table that maintains the fields of study and their unique primary keys (FIELDS_OF_STUDY table).

*Exponential*. For 0 <= $i$ < $n$, exactly one feature has weight $\sqrt{\frac{0.75}{2^i}}$.

*Linear*. For 0 <= $i$ <= $n$-1, exactly one feature has weight $1-\frac{1+i}{1+n}$.

*Random*. Feature weights are uniformly distributed on the [0..1] interval.

*Equal*. All features are weighted equally.

Consider courses A, B, C, with the following values for the study field:

A: { DSP }

B: { DSP, Optics, Networking }

C: { Optics }

This asymmetric overlap metric looks at the size of the feature set intersection relative to the size of the set in the query case. For example, the similarity between A and B is maximal because B overlaps with everything found in A. The similarity between B and A is less because A only contains 1/3 of what is found in B. This metric is intentionally asymmetric under the assumption that a student looking for DSP knowledge would be satisfied by a course that covered both DSP and Optics, but one looking for a course that combined Microwave and Satellite materials would not necessarily be happy with a course that focused on one or the other (Burke, 2001).

#### 4.4.2.2 Early Attempts

##### *4.4.2.2.1 How to determine overall matching percentage*

The database system should maintain a catalogue of what courses are contained in each field of study (COURSES_IN_FIELD_OF_STUDY table).

There should be a table that maintains the fields of study and their unique primary keys (FIELDS_OF_STUDY table).

**Original source visual: printed page 83, PDF page 93.**

**Original printed page 84 · source PDF page 94 · Open source page**

Then retrieve a list of courses from the COURSES_IN_FIELD_OF_STUDY table whose fields of study are in the student's top four choices.

For each course, first determine the matching weight for FLD/INT (field/interest).

The top choice would have a weight of 1.00 with a decrease of 0.25 for each of the subsequent choices (i.e., the 4th choice would have a weight of 0.25).

Depending on the importance weight of 7, 8, 9, or 10, the above value would be normalised. If the weight is to be from 1 to 4, then the difference will be too significant. For example, 4 is actually 400% of 1. I propose to have a discrete weight between 7 and 10. In this instance, 7 is actually only 30% less than 10.

For assessment preference attribute, simply construct a vector of (EXAM, PROJ, ASGN).

Then, for each of the courses in the above temporary list use a simple IR matching algorithm to find a matching percentage, and normalise this value according to the discrete importance factor between 7 and 10.

Not taking into account the specialisation attribute and the grade expectation attribute there will be two factors, one with an importance factor of 10 and the other of 9. Thus, add the two normalised weights, and divide the sum by 10+9=19 to get the overall matching percentage.

#### 4.4.2.2.2 Explanation of algorithm

The first step is to define a description of the new problem to solve. This is actually the problem section in a case, and resides in the problem space. The description can be a set of “distinct” attributes that can characterise the features or the “personality” of each case. The description of the new problem is then used to find the best possible match from the description of solved problems.

Then retrieve a list of courses from the COURSES_IN_FIELD_OF_STUDY table whose fields of study are in the student's top four choices.

For each course, first determine the matching weight for FLD/INT (field/interest).

The top choice would have a weight of 1.00 with a decrease of 0.25 for each of the subsequent choices (i.e., the 4th choice would have a weight of 0.25).

Depending on the importance weight of 7, 8, 9, or 10, the above value would be normalised. If the weight is to be from 1 to 4, then the difference will be too significant. For example, 4 is actually 400% of 1. I propose to have a discrete weight between 7 and 10. In this instance, 7 is actually only 30% less than 10.

For assessment preference attribute, simply construct a vector of (EXAM, PROJ, ASGN).

Then, for each of the courses in the above temporary list use a simple IR matching algorithm to find a matching percentage, and normalise this value according to the discrete importance factor between 7 and 10.

Not taking into account the specialisation attribute and the grade expectation attribute there will be two factors, one with an importance factor of 10 and the other of 9. Thus, add the two normalised weights, and divide the sum by 10+9=19 to get the overall matching percentage.

#### *4.4.2.2.2 Explanation of algorithm*

The first step is to define a description of the new problem to solve. This is actually the problem section in a case, and resides in the problem space. The description can be a set of "distinct" attributes that can characterise the features or the "personality" of each case. The description of the new problem is then used to find the best possible match from the description of solved problems.

**Original source visual: printed page 84, PDF page 94.**

**Original printed page 85 · source PDF page 95 · Open source page**

Once the best match is found, retrieve the corresponding stored solution in the solution space. With the stored solution, the system has to perform adaptation or reparation to create a new solution for the given problem. This adaptation can be rule-based. For example, removing the courses whose prerequisites are not fulfilled and applying the rating/ranking based on secondary attributes. Actually, the adaptation cycle can be thought of as "personalisation" based on a student's individual preferences: specialisation (expert or all-rounder), assessment preference, grade expectation versus grade distribution in each course listed in the preliminary resultant course lists (stored solutions), and academic history (to determine student's potential) versus the grade distribution of courses.

#### 4.4.2.2.3 Nearest Neighbour Algorithm

The trivial nearest neighbour matching formula, which is the correlation between the two vectors (vector 1 times vector 2), divided by the size of vector 1 plus the size of vector 2, seems to be too trivial for matching the academic advice cases.

Nearest neighbour matching, as the name implies, should be ranked on the distance between these two vectors. The closer they are, the better match they will make. The algorithm is simple. Pick one of the vectors to be the reference point (it does not matter which one). For each of the case attributes, find its difference by subtracting the reference value from the corresponding case attribute. Then, for each of the attributes, square it to eliminate the effect of signs (positive/negative). Sum each of the squares up and the distance is simply the square root of this sum.

However, the best possible match will return zero distance, and we want this to be 100% matching value. To do this, we set the initial value to 1. Then, for any non-zero distance, we penalise the matching percentage by subtracting a ratio from it. The ratio is merely the distance between these two vectors to the size of the problem vector (used as reference, as this will remain constant throughout the whole matching process with this particular problem).

**Original printed page 86 · source PDF page 96 · Open source page**

But, again, the INT vector has unique characteristics in itself. The highest preference is denoted by a higher numerical value. Because of this convention, we can make use of these weights (within INT vector only). Thus, it makes sense that the INT vector should be matched using correlation between them. In the improved formula, we then calculate the matching percentage from the correlation of the INT vector and the distance of the (PRFM, CHRT) vector. The distance calculation is also improved by using the distance from the mean value instead. For the time being, this mean is derived from each pair of vectors being matched. But, there is a possibility that using mean values from the Knowledge Base may increase the reliability of the formula even further, although this needs to be investigated.

Although INT contains five sub-attributes, it is essentially an attribute for the fields of interest. PRFM and CHRT attributes can be deemed lower priority than the INT attribute. Thus, a weight of 50% is assigned to the correlation percentage of INT vector, and 50% weighting is assigned to the mean distance of (PRFM, CHRT) vector. The sum of these two components constitutes the overall matching percentage, which is then used for ranking in the next step.

#### 4.4.2.2.4 Details of early algorithm

1. Create description of new problem to solve

a. Obtain the top four choices of fields of interest from a student. This can form an "INT" (or interest) vector with the top most choice assigned a value of 4, then decrement by a step of 1 down to a value of 1 for the last choice. For non-selected field, a value of 0 is assigned. The vector has a form of (DSP, OPTIC, MWAVE, WLESS, NTWRK, SATEL) to reflect the study fields in Digital Signal Processing, Fibre Optics, Microwave, Wireless Communication, Networking, and Satellites.

b. To capture the academic performance of students, calculate a GPA and a WAM. (Would the inclusion of a difference of GPA-WAM be useful here?)

**Original printed page 87 · source PDF page 97 · Open source page**

i. GPA excludes calculation of NN results and assigns value of 1 for PA, 2 – CR, 3 – DI and 4 – HD. Divide the total sum with number of "passed" courses to get the GPA.

ii. WAM includes NN results and assigns value of -1 for NN, 0 – PA, 1 – CR, 2 – DI and 3 – HD. Then, divide the total sum with number of attempted courses to get the WAM.

c. To capture the current progress level of students, record the Chrono-Year Level (CHRON), Effective Year Level (EFFYR), and the Percentage of "passed" courses towards the overall requirement for graduation (PGRSS).

2. Use Nearest-Neighbour Matching technique to find the best match to the new problem in the problem space. This returns the description of a solved problem.

3. Retrieve the stored solution of the solved problem from the solution space. This returns academic advice previously given to the student whose profile is described in the solved problem.

4. Perform case adaptation or reparation. For each course in the preliminary resultant list, do the following:

a. Filter out the courses whose prerequisites are not fulfilled. This is likely to be RBR.

b. Rank courses by calculating:

i. Assessment preference in the form of (EXAM, PROJ, ASGN) vector

ii. Grade Expectation versus Grade Distribution in each course. Ensure the grade distribution is shown as a cumulative percentage. For example, if a course has 10% HD, 20% DI, 30% CR, 20% PA and 20% NN, the cumulative percent of DI is 20% and for CR is 60%.

**Original printed page 88 · source PDF page 98 · Open source page**

iii. Academic History (to indicate potential performance) versus Grade Distribution of courses (refer to point ii above.)

c. Apply the Specialisation Preference or CHRT parameter.

Values for CHRT

i. An all-rounder with broad knowledge across diversified fields (-2)

ii. Broad knowledge in many fields but with some small specialisation (-1)

iii. No preferences at all (0)

iv. High specialisation in a field, but with broad knowledge in few other fields (1)

v. Absolute specialisation in one field with minimal knowledge in other fields (2)

Weighting assignment

i. Value -2 assigns weight of 10, 9.9, 9.8 and 9.7

ii. Value -1 assigns weight of 10, 9.5, 9 and 8.5

iii. Value 0 assigns weight of 10, 9, 8, 7

iv. Value 1 assigns weight of 10, 8, 6 and 4

v. Value 2 assigns weight of 10, 4, 2 and 1

These weights are then applied for the top four choices of the fields of interest.

d. Use point b and point c above to rank the courses.

5. Display the ranked course list as the academic advice (the solution) for the current student (the description of new problem to solve).

Original printed page 89 · source PDF page 99 · Open source page

Assuming an attribute value based case representation, a corresponding similarity measure consists of three basic elements (Stahl, 2002):

1. Local similarity functions used to compute the similarity between query and case attribute values. Describe why statistical correlation is suitable for INT and “Distance of Means” suits PRFM & CHRT.

2. Attribute weights representing the relevance of each attribute for the similarity assessment (Refer to the Nearest Neighbour Algorithm above)

3. An amalgamation function responsible for the computation of a final similarity value for a particular query-case pair based on the local similarities and attribute weights (Refer to the Nearest Neighbour Algorithm above)

### 4.5 The “Prototype System”

The main advantage of CBR over other knowledge engineering techniques is that typical users should be able to read a case and understand it without much background information. Unfortunately, due to time constraint, this main feature was not exploited, and the prototype system was developed as a proof of concept to show how CBR could work in general in this academic advisement context.

#### 4.5.1 Decision Breakdown

Academic advisement can be broken down into three independent decisions:

1. Core courses that must be taken to finish the program requirement

2. Compulsory electives that dictate the amount of elective credit points that must be reached, although students have a certain level of freedom to choose the exact courses to satisfy this group of electives

a. Technical electives that are highly related to engineering, either as foundation or specialisation

i. Stream electives that essentially determine the major can also make up the minor as well

Original printed page 90 · source PDF page 100 · Open source page

ii. Non-Stream electives that include physics in the 1st year and mathematics throughout the entire program

b. Non-Technical electives (e.g., context curriculum courses, business courses)

3. Free electives that allow students to pursue other non-technical interests in an academic environment

Given the rather complex nature of stream electives selection, CBR can be used here to develop an automated academic advising system. For example, certain attributes such as fields of interest (INT), student academic performance (PRFM), and specialisation characteristics (CHRT) can be used to uniquely identify and index cases. This makes up the problem space in CBR terminology.

The actual choice of courses that a student with a particular profile (i.e., problem) selects can then be stored as the solution for this problem. This is a part of the solution space in CBR terminology.

#### 4.5.2 3-pass process of Stream Elective decision

Given that academic advice on stream electives can involve a complex decision making process, this process should be broken down into small parts involving a set of simpler decision to be made one at a time. To handle decision making for stream electives, there should be multi-pass processes that can subsequently filter out irrelevant cases.

Suppose that each existing case in the database (knowledge base) contains the following information:

- The history of attempt for each course

**Original printed page 91 · source PDF page 101 · Open source page**

- Academic history or results obtained for each course in the program
- Preference of a major stream and a minor stream. During each re-enrolment period, students can choose to either verify/confirm or alter their current stream preference

Consider the different passes when obtaining a problem case:

The first pass should be rule-based and this should filter out courses that fail prerequisite tests. The output of this pass should store the eligible courses and stream preference in a structure of data (to be defined).

The initialisation of second pass converts this structure into generic attributes and left specific streams or courses from consideration. For example:

- Has the student passed the 1st year course in the major stream?
- Has the student passed the 1st year course in the minor stream?

This approach would enable the system to capture the behaviour/characteristics of students in terms of the significance of the courses (e.g., major electives/minor electives).

The output of the initialisation should return a vector containing these attributes. This vector is then fed into a case-based system, which contains similar attributes. The advice or the solution part of each of the existing case in the CBR knowledge base should be in the following format:

- Student should enrol in the 1st year courses corresponding to their major stream of choice
- Student should change the minor stream into the major stream by enrolling in the 2nd year courses of the new major stream

An attribute of how popular each past advice is students should be stored in the knowledge base and be updated after the current user/student finalises his/her re-enrolment. This attribute can be used as part of the ratings.

**Original printed page 92 · source PDF page 102 · Open source page**

The CBR system in this second pass should - quite safely - be able to utilise the basic information retrieval (IR) algorithm, which involves a dot product divided by the product of two vectors' sizes.

The third pass would take into account that some courses might be harder than others. The initialisation of the third pass would create a temporary "problem" vector, which is essentially a combination of the structure in the first pass and the solution vector from the second pass. This "temp" vector is then fed into the CBR system in the third pass. This would enable us to observe a student's reaction/response to the specific advice (i.e., specific courses to be taken), the given student's individual performance in his/her choice of courses, and preference and advice based on the significance of the courses.

### 4.5.3 Common Rules for Compulsory Elective decision

Within compulsory electives decision making for full-time students, there are some governing rules that must be enforced:

- The advised list of courses should maximise the chance for students to complete the program within the minimum possible time
- The total credit points allowed in any given enrolment year should be the maximum possible number of points, which is less than or equal to the credit points prescribed for full-time study load (e.g., 96 credit points for current program structure in RMIT)
- Technical electives must not be undertaken where their prerequisite courses (core and/or technical electives) have not been successfully completed. Electives should automatically be readily available to students where there are no prerequisite courses

- Core courses should always take precedent over compulsory electives, which in turn should always take precedent over free electives
- Within compulsory electives, technical electives should always take precedent over non-technical electives
- Within technical electives, stream electives should always take precedent over non-stream electives

**Original printed page 93 · source PDF page 103 · Open source page**

- Students should have successfully completed all compulsory electives from earlier academic years before attempting the compulsory electives belonging in the following or later academic year
- If a grade of Pass (PA) or better has been achieved for a course, this course must not be taken again. This is specific to Australia, where government funding will not financially support students who attempt the same course for a possibly higher grade, as may be allowed in the USA. International students are similarly restricted by visa rules in Australia.

Students are allowed to take more compulsory electives if fewer than the usual core courses are being attempted in the same enrolment year. However, the maximum limit based on Rule 2 must be imposed.

#### 4.5.4 External input for Compulsory Elective decision

To customise academic advice on an individual student basis, some personal preferences are required. An obvious example would be personal likings in particular fields of technology. Another factor, which may be deemed indirect or too vague, is influence from friends, colleagues, and study groups. Often, students tend to stick to their study group, which is likely formed during the 1st year of the program. Although it was hard to capture, quantify, or represent this aspect, it would have a significant effect on students that were not scholarly sound and had to rely on aid from their friends most of the time.

#### 4.5.5 Substructure of course

The structure of each stream elective course should comprise of its own sub-attributes (e.g., course ID, stream it belongs to, academic year level it belongs to, and its prerequisite course – assuming there is only one immediate prerequisite course for each course entity). Thus, based on Matrix A-L example, each course entity can be represented as (course ID, year level, stream ID, prerequisite), for example, (1C, 1, C, 0), (2A, 2, A, 1A). Please note that 0 in the prerequisite

**Original printed page 94 · source PDF page 104 · Open source page**

attribute denotes NULL value, meaning there is no prerequisite for that particular course entity. For simplicity, this representation ignores co-requisite courses in the meantime.

##### 4.5.5.1 Excel Prototype 3-year program

Initially, a Microsoft Excel spreadsheet prototype was developed to gain a basic understanding on the overall operation of CBR.

###### 4.5.5.1.1 Test case structure using fixed number of measures in a matrix.

Program X (fixed program, for simplicity) has three academic years, each of which contains five courses, as outlined below.

There are three series of core courses, namely A, B, and C.

Series A and B go all the way to final year, whereas Series C is compulsory for 1st and 2nd years only.

There are also three series of stream electives, namely D, E, and F.

Two stream electives can be chosen in 1st year to allow students to explore their preferences. From the 2nd year onwards, only one stream must be selected from the two streams undertaken in the 1st year.

There are three 2nd year free electives (namely G, H, and I) and three 3rd year free electives (namely J, K, and L).

Students must pick one from each set when they are enrolled in the relevant academic years.

Year 1:

Core courses: 1A, 1B, 1C

**Original printed page 95 · source PDF page 105 · Open source page**

Pick 2 stream elective from: 1D, 1E, 1F

Year 2:

Core courses: 2A, 2B, 2C

Pick 1 stream elective from: 2D, 2E, 2F

Pick 1 free elective from: 2G, 2H, 2I

Year 3:

Core courses: 3A, 3B

Pick 1 stream elective from: 3D, 3E, 3F

Pick 2 free electives from: 3J, 3K, 3L

Given that the advice on core courses take precedence over advice on electives, and that they can be enforced using rules in RBR, CBR should only be applied to the advice on stream electives and free electives only.

Thus, academic history can be extracted from the transcripts and then fed into the CBR as inputs. The first pass should also feed the history to a RBR system to determine the courses availability.

This will form another set of inputs for the CBR system.

The output part of each case should then show the strongly recommended electives that students should be taking in the following enrolled year.

Other important information that is needed in each case is a student's preference on the major stream. This measure can assume a value of D, E, or F.

The last measure would be the academic year of each student. This can be determined by the majority of courses this student undertook in the previous year.

Thus, the case structure would look something like this:

**Original printed page 96 · source PDF page 106 · Open source page**

- Input part (indices should be built for this part only):
    - Results of 1D, 1E, 1F, 2D, 2E, 2F, 2G, 2H, 2I, 3D, 3E, 3F, 3J, 3K, and 3L
    - Availability of 1D, 1E, 1F, 2D, 2E, 2F, 2G, 2H, 2I, 3D, 3E, 3F, 3J, 3K, and 3L
    - Stream Preference
    - Previous Academic Year
- Output part:
    - Strong recommendation list of 1D, 1E, 1F, 2D, 2E, 2F, 2G, 2H, 2I, 3D, 3E, 3F, 3J, 3K, and 3L
- History transcripts are recorded as Pass (1) or Fail (0) only.

- Available courses are recorded as “can be taken” (1) or “cannot be taken” (0). This simply shows every possible course that students can take, regardless of academic level, although prerequisites are kept in mind.
- More complex CBR can then sort the available courses according to the rankings/ratings. The limit on the maximum number of enrolled courses each year will be enforced with highest priority.
- Academic advices are recorded as “should be taken” (1) or “should not be taken” (0).

Assumptions:

- Student Number is not recorded as a measure to ensure simplicity. But, this ignores a student’s individual scholastic capability.
- 1A is a prerequisite of 2A, which is in turn a prerequisite of 3A. The alphabet denotes the sequence of prerequisites.
- For stream electives (D, E, F), assume that stream D is the most popular and stream F is the LEAST popular.

Students must also be enrolled in five courses each year unless they are in their final year and have fewer than five courses to complete before graduation.

#### 4.5.5.2 Test Structure of the Rich Matrix

- FLD denotes field of study and assumes the following predefined test values:

**Original printed page 97 · source PDF page 107 · Open source page**

  - Mobile Communication
  - Microcontroller
  - Internetworking Protocols
  - Software Engineering
- POP denotes Popularity or Peer Preference as obtained during the pre-re-enrolment survey.
- DIFF denotes perceived degree of difficulties and is measured in two ways:
- PASS denotes the pass rate
- HDDI denotes percentage of getting good results (e.g., Grade of Distinction and above, DI & HD, for RMIT)
- LECT denotes the experience and reputation of teaching lecturers and can be broken down into two groups, then into sub-attributes.
- Factual information can be calculated:
  - YRSCRR denotes the number of years since starting their teaching career
  - YRSCRS denotes the number of years since lecturing in this particular course
- Opinionated data are obtained from End-of-Course student survey in the immediate past:
  - EXPT denotes the lecturer’s perceived competency as an expert in the field of study of the course
  - PRST denotes the effectiveness of the lecturer’s presentation skills and styles
  - TOUGH denotes the lecturer’s tendency towards authoring tough and difficult assessments
- ASSM denotes the type of assessments of the course:

- ○ EXAM denotes the presence of a final examination and assumes values of “Have exam” (1) and “No exam” (0)
- ○ ASGN denotes the number of assignments. For simplicity, the maximum number of assignments is set to two. Thus, ASGN can assume values of 0, 1, or 2

#### 4.5.5.2.1 Student Dialogs

Dialog with the student is the key to a successful implementation of an intelligent catalogue search application. Some questions need to be answered before a developer can prepare the

**Original printed page 98 · source PDF page 108 · Open source page**

dialogs to support the student in using the search system. Do students need to weight the parameters for which they specify values? (To answer: What validation is required for inputs to the search? What explanation of the search parameters is needed? What explanation of the results is needed by the student? What extra features will the student need to make use of the query or results featured in the system?) The answers may need to be obtained by consulting students directly or by consulting academic staff that deal with student queries on a daily basis. Consequently, the catalogue application should be open to validation based on feedback from its use in the “field”. This step is important to ensure the catalogue’s long-term acceptance and success (INRECA, 99).

#### 4.5.5.3 Conversational CBR

CCBR needs to find out the following:

- INT denotes personal interest and takes the same values as FLD
- MKT denotes the current job market or industry demand, as researched by external Careers Consultants, and assumes the same predefined test values as FLD
- CHAR denotes a student’s character, personality, or learning styles, whether it is to become an expert in a particular field (1), to be an all-rounder with broad knowledge across diversified fields (-1), or no preference at all (0)
- PEER denotes whether a student wants a popular course (1), or non-popular (0).
- LECT denotes a profile of lecturers that a student wants to study under
- ASSM denotes the type of assessment that a student prefers

The Course Set forms the advice or the solution that the CBR system returns.

For test purposes, each field as denoted in the INT attribute has two electives for students to choose from. For example, Course C1 and Course C2 belong to Internetworking Protocols or Field C. For simplicity, we assume that the prerequisite has been checked during the RBR pass.

**Original printed page 99 · source PDF page 109 · Open source page**

Once CCBR obtains the necessary preference values for each attribute, this is calculated against the attributes of each course. The simple IR algorithm may suffice in providing the ranking. This is how they are going to be used:

- INT and MKT are calculated against FLD
- CHAR is to be investigated further
- PEER is calculated against POP

Sub-attributes of LECT and ASSM are to be calculated against their counterparts, respectively.

The next step is to find how the CHAR attribute fits into the calculation quantitatively, and to investigate how efficiently the simple IR algorithm can provide the ranking/rating of each course.

#### 4.5.5.3.1 CCBR Questions for students

Students can tick the checkbox for the questions they think are relevant or important to them.

Note: Ranking arranges them so they have a position in relation to one another, or to give the order of importance among all factors. Rating is to give weighting of each factor and is in the form of a scale.

Please rank the top four choices in the fields of study (FLD or INT) that you are interested in (4 being the most interested, and 1 being the least interested)

- Mobile Communication
- Microcontroller
- Internetworking Protocols
- Software Engineering
- Fibre Optics •
- Digital Signal Processing
- Power Engineering
- Biomedical Devices

**Original printed page 100 · source PDF page 110 · Open source page**

Please select your preference towards the range of knowledge (CHAR).

- An all-rounder with broad knowledge across diversified fields (-2)
- Broad knowledge in many fields but with some small specialisation (-1)
- No preferences at all (0)
- High specialisation in a field, but with broad knowledge in few other fields (1)
- Absolute specialisation in one field with minimal knowledge in other fields (2)

Please choose your preferred methods of assessment (ASSM):

- Would you like to have an exam (EXAM)? Yes (1) or No (0)
- Would you like to have a major project? Yes (1) or No (0)
- How many assignments would you prefer to have (ASGN)? 0, 1, or 2

Which of the following profiles best describes your expectation? This is used to aid the academic advisement only, and not to incriminate students in anyway.

- I expect to get HDs in all courses.
- I expect to get DIs or HDs for most of my courses.
- I expect to get CRs or DIs for most of my courses.
- I expect to get Pas or CRs for most of my courses.
- I am happy to get a pass (PAs) in most courses.

After students pick their questions and answers, collate each question to make a list of important factors and then ask students to rank them. Each of the four factors above will have to be assigned a weight. If the weight is to be from 1 to 4, then the difference will be too significant. For example, 4 is actually 400% of 1. I propose to have a discrete weight between 6 and 10. In this instance, 6 is actually only 40% less than 10.

Now, look at the matching technique.

**Original printed page 101 · source PDF page 111 · Open source page**

A student's interest is matched using the simple IR algorithm. A course can be classified in a few different fields. Its tendency in each field can be represented using a weighting value. When matching, the student's interest rank vector is calculated against the course's field vector to determine the matching percentage. This percentage is then normalised to have the weight (6-10) according to the overall rank above.

Expertise factor has a discrete value between -2 and 2. Exactly how this will be calculated to a matching percentage is still to be determined.

This is similar to factor 1. Methods of assessment can be represented using a vector. This vector is then matched with a student's preferred method of assessment vector using a simple IR algorithm. The matching percentage is then normalised accordingly to a value between 6 and 10. We have to quantize the grade range. For example:

HDs → 86-100%

DI/HD → 77-85%

CR/DI → 67-76%

PA/CR → 58-66%

PAs → 50-57%

Then, the course's history grade statistics are compared to determine where in the above ranges each course's highest concentration of grades lies.

For those students who do not tick any questions (i.e., do not really care or cannot decide), the system may suggest they follow the popularity of courses or peer preference.

#### 4.5.5.4 Proof of Concept - Test Cases

Legends of Test Cases using MS Excel

Courses in INT (interests)

- DSP: D1, D2, D3

**Original printed page 102 · source PDF page 112 · Open source page**

- OPTIC: O1, O2, O3, OM2, OM3
- MWAVE: OM2, OM3, M1, M2, MN3
- NTWRK: MN3, N1, N2, NS3
- SATEL: NS3, S1, S2, S3

Note: Courses with a single alphabet denote a specialised course in its field.

Note 2: The number in the course code denotes that year level.

Sequence of Prerequisites

D1 → D2 → D3

O1 → O2 → O3

O1,M1 → OM2 → OM3

M1 → M2

M1,N2 → MN3

N1,N2,S2 → NS3 (note: N1 is NOT a prerequisite for N2)

S1 → S2 → S3

PRFM (performance or aptitude level): 1-weak, 2-average, 3-good

CHAR (specialisation preference): 1-broad knowledge, 2-some specialisation, 3-absolute specialisation

The test cases raise an issue of indexing.

Given that each case will have multi-dimensional indices, a special type of DBMS will be required to efficiently retrieve (or match) cases.

For the illustration and prototype, we may have to go through each record (i.e., case) in DBMS, calculate the matching percentage using the simple IR formula, and then retrieve the case whose percentage is the highest.

**Original printed page 103 · source PDF page 113 · Open source page**

When trying to match the PRFM (student's performance) with DIFF (difficulty of courses), the expert would probably be giving advice according to these "rules":

- If PRFM=3, student can do any courses, and thus INT and CHAR play an important role.
- If PRFM=2, student should be doing courses with DIFF of 2 or 1. Under some circumstances, they can probably be allowed to enrol into courses with DIFF=3. INT and CHAR may have less effect here.
- If PRFM=1, student must really be doing courses with DIFF=1. Under no circumstances will they be allowed to do courses with DIFF=3, but the probability of enrolling into courses with DIFF=2 remains. In this case, DIFF probably overrides INT and CHAR.

Without using DIFF for eight courses, there are three possible outcomes for each of DIFF, CHAR, AVG, and EFFYR, and thus, 3^4 = 81 cases.

For INT, given five choices, we are to pick three choices in order. This is ${}_5P_3$ or $\frac{5!}{(5-3)!}=60$ cases.

Thus, without DIFF, there are 4,860 distinct cases.

Inclusion of DIFF creates three possible cases for each course and, thus 3^8=6,561 cases, i.e., 4860*6561=31,886,460 or 31 million distinct cases.

For simplicity, DIFF will remain constant between semesters, so 6561 factor will be kept as 1 instead. The course list is proposed to contain two difficult courses (DIFF=3), three average courses (DIFF=2), and three easy courses (DIFF=1).

AVG will only be useful when real student data is available. With a real data set, correlation between the average number of attempts and student performance can then probably be investigated. For the time being, we can ignore this attribute.

If we deal with students with a clear pass, we can also ignore the effective year level (EFFYR).

**Original printed page 104 · source PDF page 114 · Open source page**

Thus, we would have 60 (INT) x 3 (PRFM) x 3 (CHAR) = 540 cases. However, because there are no year levels stored as an index, we may have to break up the CBR down to three sub-systems, one for each of the year level.

#### 4.5.6 Description of the prototype Case-based Reasoning Academic Advisement system

This prototype system has been developed to demonstrate the application of CBR in academic advising. The system can be test-driven at http://cbr.tanat.net/. The system was developed using PHP scripting language and MySQL database.

Upon entering the front page of the system, links to three different test sets of knowledge base are provided. On the same page, a table of sample test problems are shown. Users can click on provided links to launch the academic advisement page. Each link has its own designated test set knowledge base.

Once the academic advisement page is loaded with a specific test problem and a designated knowledge, users are presented with three tables outlining the top ten ranking of the retrieved knowledge cases. Each table uses a different algorithm to calculate the ranking percentage. The three algorithms are Trivial IR, Euclidean Distance, and Improved Combination, as described earlier.

Once the retrieved cases are ranked, these top ten cases were fed into a rule-based system to filter out courses that were deemed to be more difficult than the advised student's perceived past academic performance. For each course, this pass also recorded the highest-ranking percentage of the case in which the course first appeared in the advised set.

This stage also took into account the frequency of occurrence of each course in the top ten cases, as well as the weighting according to the case order that they appeared. This value was referred to as a "weighted sum" of advised course occurrence. The role of weighted sums is to break the tie between two or more courses with the same ranking percentage.

**Original printed page 105 · source PDF page 115 · Open source page**

The final stage shows the actual academic advice against each available course with simple reasons to support its argument.

The system can currently show courses under four categories:

- The RED text denotes courses whose prerequisites have not yet been fulfilled.
- Courses that are not recommended due to their perceived difficulty being greater than the student's demonstrated academic performance. This is denoted in ORANGE text.
- The GREEN text denotes a recommended course with its rating (ranking percentage) and a sub-rating (the tie-breaking weighted sum).
- Other possible courses that students may consider taking if they prefer are denoted in BLUE text.

### 4.5.7 Initial Direction

Initially, the author was concentrating on the CBR application, which is driven by the student profile. The algorithm is as follows:

- Obtaining student's profile and make this a "problem" profile
- Find the best possible match of the "problem" profile in the database of "knowledge" profiles
- Once the best match or matches of profiles are retrieved, show the pattern or the list of courses undertaken by previous students of such profiles

#### 4.5.7.1 A note about initial direction

In regards to grade results, the student's expectation should be obtained. In turn, this is matched with the historical academic performance level of each student in the previous terms. In order to calculate the performance level, higher significance (i.e., higher weighting) should be applied to the recent terms to determine the current student's performance, as well as to capture any potential performance pattern in the past.

**Original printed page 106 · source PDF page 116 · Open source page**

The RMIT full-time load for one academic year is 96 credit points (CP). This consists of eight courses of 12 CP each. Other universities (e.g., Swinburne) make their subjects (which are equivalent to RMIT courses) have a weighting of 12.5 CP each. Thus, one full-time year load has a round figure of 100 CP.

Perhaps this would make it a bit easier to quantize a range of grades - that is, we could calculate the percentage of each of the grades (HD, DI, CR, PA, NN).

An interesting question arises when comparing the profile of a student who has obtained four HDs and four Pas, and another student with four DIs and four CRs. Given that the credit points of each course are equal (e.g., 12 CP), the Grade-per-Average will be apparently the same, but which student should be deemed more academically competent?

#### 4.5.8 Change in Direction

The direction was changed at a later stage to investigate the course profile driven approach. The algorithm:

- Determine a list of all currently eligible courses that a student can enrol in using RBR with the student's Academic History/Transcript
- For each possible course, determine the profiles of students previously undertaking the course
- Calculate the matching percentage, a.k.a. ranking, between these profiles and the student's profile
- Show the order of courses based on the likelihood that this student profile would be taking each course.

##### 4.5.8.1 A note about change in direction

Some real sets of courses should be used to demonstrate the statistically skewed pattern in the past. There may be a possibility of using historical data from the ECE student admin database. The only catch is to map the student numbers in the database to non-existing values, to ensure the anonymity of students remain intact.

**Original printed page 107 · source PDF page 117 · Open source page**

The CBR should be storing the profiles of the students that previously attempted each course. In each course, calculate the exact number of students choosing this course, as well as the percentage of this profile among all of the other profiles associated with this course. The percentage gives a good indication of the popularity of this course for students in this profile. This can be made a criterion for ranking.

A good CBR should have a good variety/diversity/distinction. This can be achieved from a high number of students, which in turn will result in a large number of cases. For the prototype system, this should be controlled in order to limit the number of test cases to a reasonable quantity.

Given that there are only two main elements involving enrolment (i.e., students themselves and courses available), other distinct perspectives for using CBRs cannot be determined at this time. Other possibilities may involve a hybrid model of these two perspectives with varied factors of importance. The factors of importance can probably be used to determine the execution sequence of the algorithms.

**Original printed page 108 · source PDF page 118 · Open source page**

#### 4.5.9 System Modelling

##### 4.5.9.1 System Architecture

Figure 4-1: below shows the major modules in the prototype system. It also shows which components interact with each other.

Figure 4-1: System Architecture

## 4.5.9 System Modelling

### 4.5.9.1 System Architecture

Figure 4-1: below shows the major modules in the prototype system. It also shows which components interact with each other.

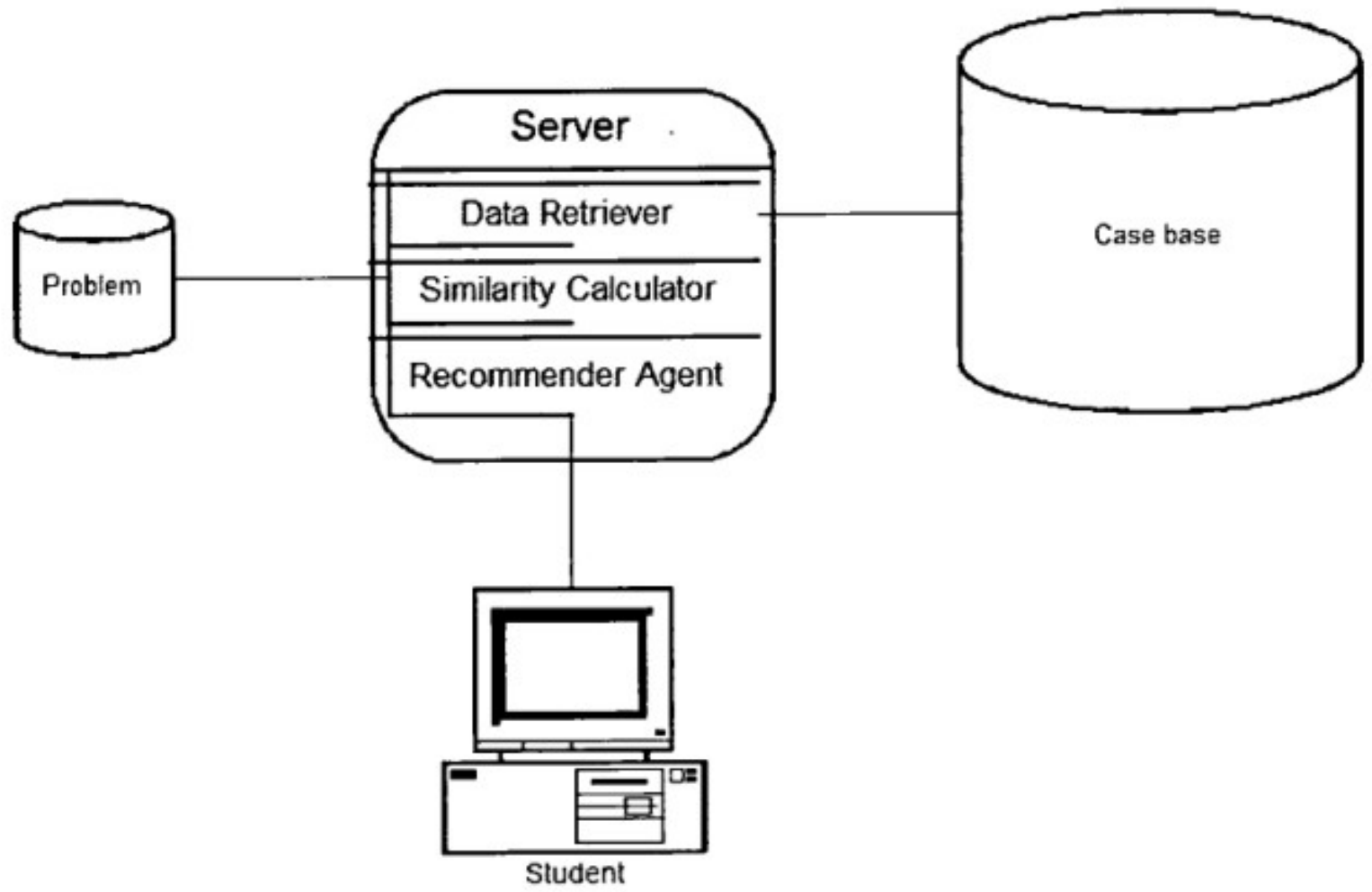

**Figure 4-1: System Architecture**

**Original source visual: printed page 108, PDF page 118.**

**Original printed page 109 · source PDF page 119 · Open source page**

### 4.5.9.2 Object Model

Figure 4-2 below shows the aggregation between knowledge base and smaller objects that make up one case.

Figure 4-2: Object Model

#### 4.5.9.2 Object Model

Figure 4-2 below shows the aggregation between knowledge base and smaller objects that make up one case.

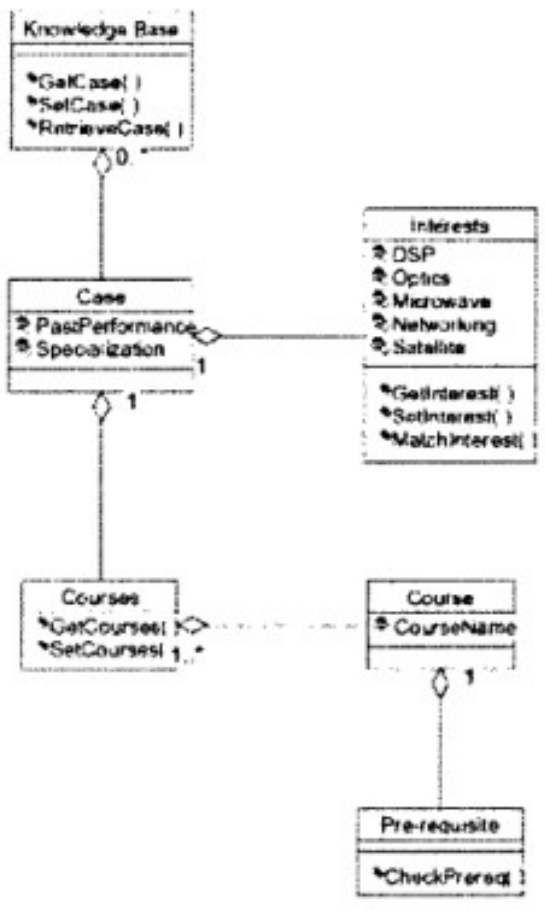

**Figure 4-2: Object Model**

109

**Original source visual: printed page 109, PDF page 119.**

#### 4.5.9.3 Entity-Relationship Diagram

Figure 4-3 below shows the Entity-Relationship (ER) diagram of the database, as the knowledge or case base is stored in MySQL database software.

Figure 4-3: ER diagram

#### 4.5.9.3 Entity-Relationship Diagram

Figure 4-3 below shows the Entity-Relationship (ER) diagram of the database, as the knowledge or case base is stored in MySQL database software.

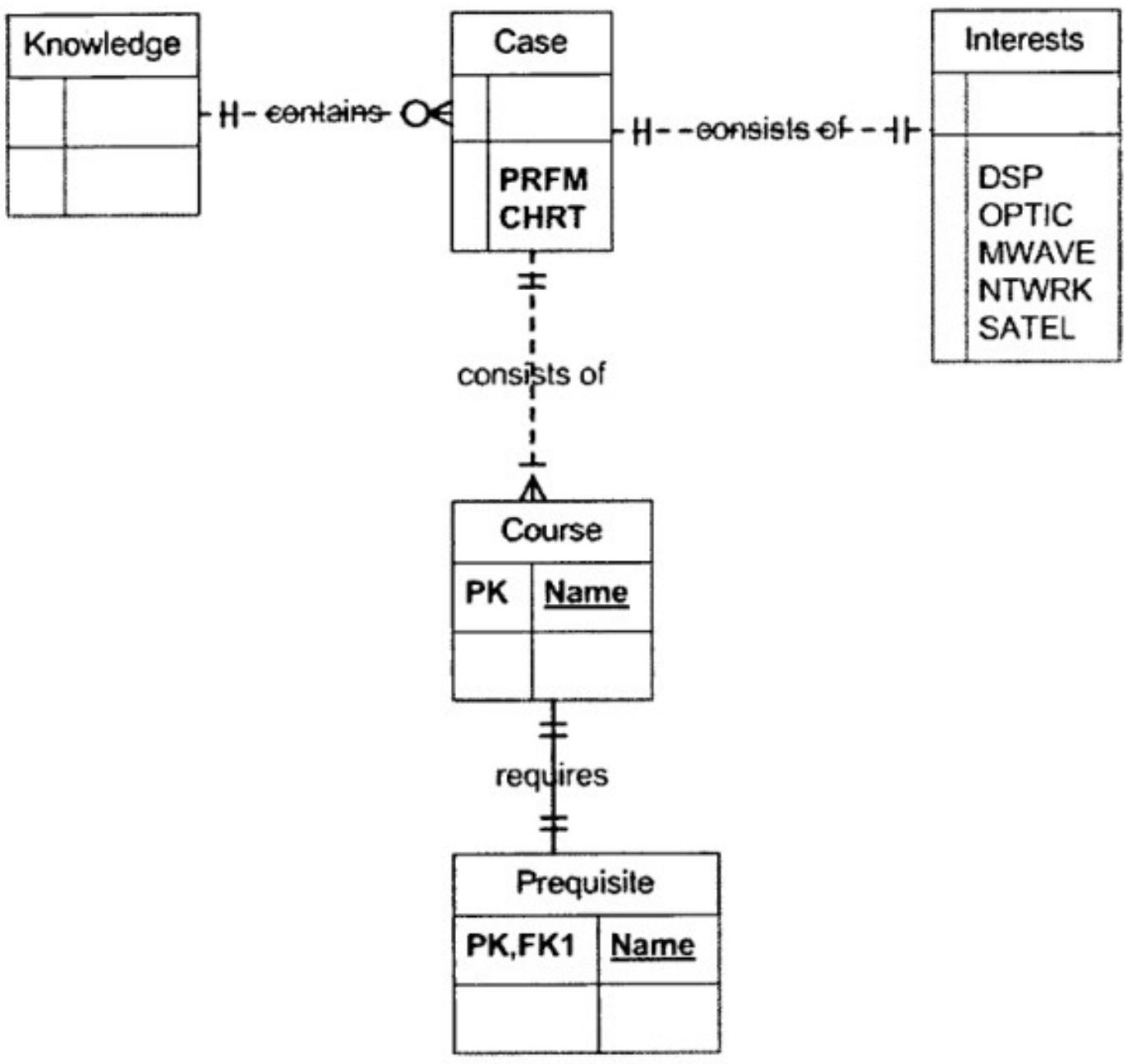

**Figure 4-3: ER diagram**

**Original source visual: printed page 110, PDF page 120.**

**Original printed page 111 · source PDF page 121 · Open source page**

#### 4.5.9.4 System Operation

Figure 4-4 below gives the system flow of control in the form of a UML Activity Diagram. This is the high level flow and shows the sequence of major operations and the two main loops in the prototype system.

Figure 4-4: Flow of Control

### 4.5.9.4 System Operation

Figure 4-4 below gives the system flow of control in the form of a UML Activity Diagram. This is the high level flow and shows the sequence of major operations and the two main loops in the prototype system.

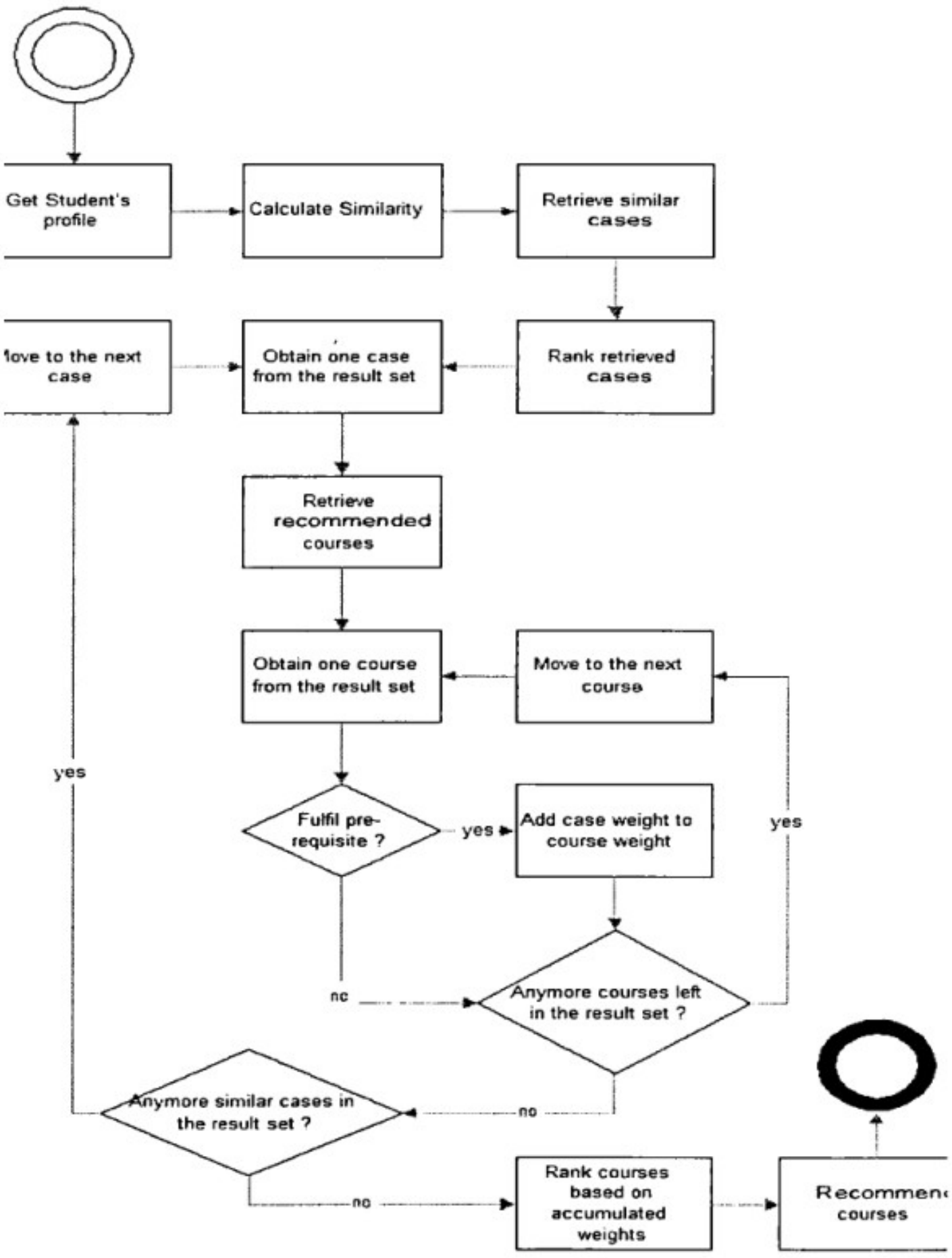

**Figure 4-4: Flow of Control**

**Original source visual: printed page 111, PDF page 121.**

Original printed page 112 · source PDF page 122 · Open source page

#### 4.5.10 Prototype Versions

The prototype versions merely serve as a proof of concept. Only the pure numerical calculations are used for similarity matching. The obvious drawback here would need the cases be coded and decoded between textual attributes and their respective numerical representation.

##### 4.5.10.1 Trivial IR

This calculation uses Trivial Information Retrieval technique, implementing a simple Nearest Neighbour formula. This is particularly useful for the measures that demonstrate statistical correlation behaviour.

##### 4.5.10.2 Euclidean Distance

This algorithm is merely a calculation of distance between two vectors. This is an adaptation of the Pythagoras triangle theorem, but with more than three dimensions. It is suitable for a set of measures that are not interrelated to one another.

##### 4.5.10.3 Improvement Calculation

The improved algorithm is actually a combination of statistical correlation of Student Fields of Interest (INT) measures and a calculation of Vector Distance from a Reference Point. This algorithm is the combined and improved version of the first two algorithms. Given the Student Fields of Interest (INT) needed to capture the relative preference order among them, the statistical correlation method seemed to be the most appropriate in such a situation. As for the other two measures, Student Specialisation Preference (CHRT) and aggregated Student Academic Performance (PRFM), the vector distance method seemed appropriate, as they are mutually exclusive. A slight improvement in the vector distance formula was to calculate a reference point so the ranking became more consistent. This reference point was calculated as a mean value of all the existing values in the knowledge base.

Original printed page 113 · source PDF page 123 · Open source page

##### 4.5.10.4 Fuzzy Matching (Future Improvement)

###### 4.5.10.4.1 Possibility Measures and Necessity Measures

If the value of a variable is described using a possibility distribution, it often needs to determine to what degree it matches a given condition. Such needs arise in fuzzy logic control as well as in fuzzy databases. In fuzzy database systems, imprecise information may be represented using possibility distributions. Retrieving information from a database involves posting a question (called a query) to the database management system. The question typically involves conditions about the kind of information we are interested in retrieving. Hence, processing these database queries for retrieving imprecise information requires the capability to match conditions in a query with possibility distributions in a fuzzy database.

In general, matching a possibility distribution of a variable x with a condition “x is A” involves asking the following two related questions:

(Q1) Given the possibility distribution of x (denoted $\Pi(x)$), is it possible for x to be A?

(Q2) Given the possibility distribution $\Pi(x)$, is it necessary for x to be A?

These two questions will be first illustrated using a non-fuzzy example. Suppose we know the following:

Jack’s GPA is between 2.00 and 2.50

Consider the following condition:

Whether a student’s GPA exceeds 2.20

Does the possibility distribution of Jack’s GPA match the condition? Due to the imprecise information about Jack’s GPA, a black-and-white answer to the question cannot be found. More specifically, it is possible for Jack's GPA to exceed 2.20; however, this is not necessarily the case. Hence, the answer to the first question listed above (Q1) is "yes," but the answer to the

#### 4.5.10.4 Fuzzy Matching (Future Improvement)

##### *4.5.10.4.1 Possibility Measures and Necessity Measures*

If the value of a variable is described using a possibility distribution, it often needs to determine to what degree it matches a given condition. Such needs arise in fuzzy logic control as well as in fuzzy databases. In fuzzy database systems, imprecise information may be represented using possibility distributions. Retrieving information from a database involves posting a question (called a query) to the database management system. The question typically involves conditions about the kind of information we are interested in retrieving. Hence, processing these database queries for retrieving imprecise information requires the capability to match conditions in a query with possibility distributions in a fuzzy database.

In general, matching a possibility distribution of a variable $x$ with a condition "$x$ *is* $A$" involves asking the following two related questions:

(Q1) Given the possibility distribution of $x$ (denoted $\Pi(x)$), is it possible for $x$ to be $A$?

(Q2) Given the possibility distribution $\Pi(x)$, is it necessary for $x$ to be $A$?

These two questions will be first illustrated using a non-fuzzy example. Suppose we know the following:

Jack's GPA is between 2.00 and 2.50

Consider the following condition:

Whether a student's GPA exceeds 2.20

Does the possibility distribution of Jack's GPA match the condition? Due to the imprecise information about Jack's GPA, a black-and-white answer to the question cannot be found. More specifically, it is possible for Jack's GPA to exceed 2.20; however, this is not necessarily the case. Hence, the answer to the first question listed above (Q1) is "yes," but the answer to the

113

**Original source visual: printed page 113, PDF page 123.**

second question (Q2) is “no.” Obviously, if we are interested in the condition that “whether a student’s GPA is between 2.10 and 2.30,” our answers to both questions are positive for Jack (i.e., it is possible as well as necessary for Jack’s GPA to be in the interval [2.10, 2.30]).

In the example above, the answers to Q1 and Q2 are either “yes” or “no” because neither Jack’s GPA nor the conditions about a student’s GPA are fuzzy. In general, however, the answer to each question is a matter of degree. The answer to Q1 is called a “possibility”, and the answer to Q2 a “necessity.”

#### 4.5.10.4.2 Example

If we are given a possibility distribution of a student’s marks, and we want to find all poor students, we can calculate the possibility degree and necessity degree§ as follows:

$$Poss(Poor/\Pi_x) = \sup_x [min(\Pi_x, \mu_{poor}(x))]$$

**Equation 1**

$$Nec(Poor/\Pi_x) = \inf_x [max(\Pi_x, 1 - \mu_{poor}(x))]$$

**Equation 2**

§ From http://www.dbai.tuwien.ac.at/marchives/fuzzy-mail/1233.html

Necessity is the counterpart of belief and possibility is the counterpart of plausibility in possibility theory. Possibility theory is a special branch of evidence theory that deals only with bodies of evidence whose focal elements are nested.

The following are true:

Nec(A intersection B) = min(Nec(A), Nec(B))

Pos(A union B) = max(Pos(A), Pos(B))

This is a simple way to think of them:

Necessity for A is the sum of probabilities for every element that is contained in A.

Possibility for A is the sum of probabilities for every element that contains A.

For more information, please read Klir G. and Bo Yuan, Fuzzy Sets and Fuzzy Logic, Theory and Applications, New Jersey, Prentice Hall, 1995

second question (Q2) is "no." Obviously, if we are interested in the condition that "whether a student's GPA is between 2.10 and 2.30," our answers to both questions are positive for Jack (i.e., it is possible as well as necessary for Jack's GPA to be in the interval [2.10, 2.30]).

In the example above, the answers to Q1 and Q2 are either "yes" or "no" because neither Jack's GPA nor the conditions about a student's GPA are fuzzy. In general, however, the answer to each question is a matter of degree. The answer to Q1 is called a "possibility", and the answer to Q2 a "necessity."

#### *4.5.10.4.2 Example*

If we are given a possibility distribution of a student's marks, and we want to find all *poor* students, we can calculate the possibility degree and necessity degree[§] as follows:

$$Poss(Poor \mid \Pi x) = \sup_x[\min(\Pi_x, \mu_{poor}(x))] \qquad \text{Equation 1}$$

$$Nec(Poor \mid \Pi_x) = \inf_x[\max(\Pi_x, 1\text{-}\mu_{poor}(x))] \qquad \text{Equation 2}$$

---

§ From http://www.dbai.tuwien.ac.at/marchives/fuzzy-mail/1233.html

Necessity is the counterpart of belief and possibility is the counterpart of plausibility in possibility theory. Possibility theory is a special branch of evidence theory that deals only with bodies of evidence whose focal elements are nested.

The following are true:
Nec(A intersection B) = min(Nec(A), Nec(B))
Pos(A union B) = max(Pos(A), Pos(B))

This is a simple way to think of them:
Necessity for A is the sum of probabilities for every element that is contained in A.
Possibility for A is the sum of probabilities for every element that contains A.

For more information, please read Klir G. and Bo Yuan, Fuzzy Sets and Fuzzy Logic, Theory and Applications, New Jersey, Prentice Hall, 1995

114

**Original source visual: printed page 114, PDF page 124.**

**Original printed page 115 · source PDF page 125 ·** Open source page

Let the universe of discourse of a student's marks be { 20, 30, 40, 50, 60, 70, 80, 90, 95 }, and the marks possibility distribution of a student (denoted J) be:

$$\Pi_{marks}(J) = 0.2/30 + 0.5/40 + 1/50 + 0.8/60$$

Suppose that the membership function for the linguistic term Poor is defined as a discrete fuzzy set as follows:

$$Poor = 1/20 + 1/30 + 1/40 + 0.8/50 + 0.4/60 + 0.2/70$$

Using equation 1, we calculate the possibility degree fro student J to satisfy the condition “poor student,”:

$$\begin{aligned} &Poss(Poor / \Pi_{marks} J) = max\{0.2 \wedge 1, 0.5 \wedge 1, 1 \wedge 0.8, 0.8 \wedge 0.4\} \\ &= max\{0.2, 0.5, 0.8, 0.4\} \\ &= 0.8 \end{aligned}$$

To calculate the necessity measure, we first calculate the complement of the possibility distribution of a student J’s marks:

$$\begin{aligned} &1 - \Pi_{marks}(J) = 1/20 + 0.8/30 + 0.5/40 + 0/50 + 0.2/60 \\ &+ 1/70 + 1/80 + 1/90 + 1/95 \end{aligned}$$

The necessity measure is obtained using equation 2

$$\begin{aligned} &Nec(Poor / \Pi_{marks} J) = min\{1 \vee 1, 1 \vee 0.8, 1 \vee 0.5, 0.8 \vee 0, 0.4 \vee 0.2, \\ &\qquad 0.2 \vee 1, 0 \vee 1, 0 \vee 1, 0 \vee 1\} \\ &= min\{1, 1, 1, 0.8, 0.4, 1, 1, 1, 1\} \\ &= 0.4 \end{aligned}$$

Therefore, the possibility that student J is academically poor is 0.8, while the necessity that he/she is academically poor is 0.4 .

Let the universe of discourse of a student's marks be { 20, 30, 40, 50, 60, 70, 80, 90, 95 }, and the marks possibility distribution of a student (denoted *J*) be:

$$\Pi_{marks}(J) = 0.2/30 + 0.5/40 + 1/50 + 0.8/60$$

Suppose that the membership function for the linguistic term Poor is defined as a discrete fuzzy set as follows:

$$Poor = 1/20 + 1/30 + 1/40 + 0.8/50 + 0.4/60 + 0.2/70$$

Using equation 1, we calculate the possibility degree fro student *J* to satisfy the condition "poor student,":

$$Poss(\ Poor \mid \Pi_{marks}, J\ ) = \max \{0.2 \wedge 1, 0.5 \wedge 1, 1 \wedge 0.8, 0.8 \wedge 0.4\}$$

$$= \max \{0.2, 0.5, 0.8, 0.4\}$$

$$= 0.8$$

To calculate the necessity measure, we first calculate the complement of the possibility distribution of a student *J*'s marks:

$$1\text{-}\Pi marks(J) = 1/20 + 0.8/30 + 0.5/40 + 0/50 + 0.2/60 + 1/70 + 1/80 + 1/90 + 1/95$$

The necessity measure is obtained using equation 2

$$Nec(Poor \mid \Pi marks, J) = \min\{1 \vee 1, 1 \vee 0.8, 1 \vee 0.5, 0.8 \vee 0, 0.4 \vee 0.2, 0.2 \vee 1, 0 \vee 1, 0 \vee 1, 0 \vee 1\}$$

$$= \min \{1, 1, 1, 0.8, 0.4, 1, 1, 1, 1\}$$

$$= 0.4$$

Therefore, the possibility that student *J* is academically poor is 0.8, while the necessity that he/she is academically poor is 0.4 .

**Original source visual: printed page 115, PDF page 125.**

**Original printed page 116 · source PDF page 126 · Open source page**

### 4.6 Case Refining (Filtering of retrieved solutions)

At this point, the three algorithms for Nearest Neighbour matching are:

1. The combination of Correlation of Interest vector and the distance of non-Interest attributes to their mean values

2. Distance of two vectors (i.e., using one vector as a reference point and find the distance of the other vector of this point)

3. Trivial Information Retrieval (IR) algorithm, utilising correlation only

It is worth noting that all three algorithms tend to return the same ranking for the top three ranks and, in many cases, the top five. The variation seems to show itself from Rank no. 6 onwards.

The matching percentage is also used to compare the cases within the same algorithm only. A comparison using the matching percentage across two or more algorithms would be deemed meaningless.

Once the top X number of cases are ranked (X = 10 at this stage), it is logical to filter out some courses to refine the solution further. In the example of the prototype academic advisement, PRFM (Performance attribute) is used to recommend whether or not each student should undertake courses which are deemed more academically demanding by matching with the DIFF (Difficulty attribute) of each course. This process is essentially equivalent to the Case Refining process in INRECA methodology.

The algorithm is simple. Loop through the array of the top X ranks. For each of the recommended courses, check to see whether the student's PRFM exceeds, or is at least equal to, the course's DIFF. If so, this course is recommended. If the advice does not yet contain enough courses (i.e., number of recommended courses is below number of courses to be taken), then proceed to the next rank and perform the same PRFM vs. DIFF checking. In the prototype system, number of courses to be taken is three.

**Original printed page 117 · source PDF page 127 · Open source page**

An issue arises when the advice set is filled with two courses so far and is proceeding to the next rank to find the next and last possible best course. What if there are two or more courses found to be suitable? Which one is to be chosen? This issue is illustrated in the Problem no. 6 (index = 0130223). More filters (i.e., more rules) may need to come into play here in order to determine a plausible recommendation of this tie.

Given that the proposed algorithm in the previous page was flawed in the sense that the ranking percentage was not utilised, the improved version is as follows:

The advice matrix consists of three categories: the number of occurrences of each course in the top ten ranks, highest percentage obtained from the best rank that each course was a solution, and the weighted count of occurrences.

The last item is based on the following ideas:

- The ten ranks should have their own weights
- For each weight, there should be a based value, say X, and a step value, say Y
- The lowest rank will have a weight of X+Y and the highest rank: X+10Y

Suppose that courses that appear in the top five ranks would be TWICE as significant as the one appearing in the last five ranks.

That is,

$$2[(X+Y)+(X+2Y)+(X+3Y)+(X+4Y)+(X+5Y)]=[(X+6Y)+(X+7Y)+(X+8Y)+(X+9Y)+(X+10Y)].$$

This equates to $X=2Y$.

To make things simple, the sum of weights $(X+Y)$ up until $(X+10Y)$ is 1.

Solve these equations and we get $X=2/75$ and $Y=1/75$.

An issue arises when the advice set is filled with two courses so far and is proceeding to the next rank to find the next and last possible best course. What if there are two or more courses found to be suitable? Which one is to be chosen? This issue is illustrated in the Problem no. 6 (index = 0130223). More filters (i.e., more rules) may need to come into play here in order to determine a plausible recommendation of this tie.

Given that the proposed algorithm in the previous page was flawed in the sense that the ranking percentage was not utilised, the improved version is as follows:

The advice matrix consists of three categories: the number of occurrences of each course in the top ten ranks, highest percentage obtained from the best rank that each course was a solution, and the weighted count of occurrences.

The last item is based on the following ideas:

- The ten ranks should have their own weights
- For each weight, there should be a based value, say X, and a step value, say Y
- The lowest rank will have a weight of X+Y and the highest rank: X+10Y

Suppose that courses that appear in the top five ranks would be TWICE as significant as the one appearing in the last five ranks.

That is,
$2[(X+Y)+(X+2Y)+(X+3Y)+(X+4Y)+(X+5Y)]=[(X+6Y)+(X+7Y)+(X+8Y)+(X+9Y)+(X+10Y)]$.
This equates to $X=2Y$.

To make things simple, the sum of weights $(X+Y)$ up until $(X+10Y)$ is 1.

Solve these equations and we get $X=2/75$ and $Y=1/75$.

117

**Original source visual: printed page 117, PDF page 127.**

**Original printed page 118 · source PDF page 128 ·** Open source page

Then, for each time the course appears in each rank, we would add these weights to get the weighted count of occurrences.

The prototype system has been improved to provide the advice along with basic reasons for each advice.

The advice matrix is now sorted by the ranked percentage as the first sort key. If there is a tie in any ranks, then the weighted count is used to break this tie. This is to ensure that each ranking contains only one distinct element.

To make the advice slightly more readable, each advice is colour-coded as follows:

- RED denotes courses whose prerequisites have not yet been fulfilled
- ORANGE denotes courses that are not recommended due to its perceived difficulty being greater than the student's demonstrated academic performance
- GREEN denotes recommended courses with ranked order
- BLUE denotes other possible courses for the student’s consideration

#### 4.6.1 Comparison-Based Recommendation

Feedback strategies can be classified in terms of the type of feedback they aim to capture (feature-level vs. case-level feedback) and the cost to the student (low-cost vs. high-cost) [29]. Four different feedback strategies have been incorporated into recommender systems to a lesser or greater extent. Value elicitation and tweaking are feature level techniques in the sense that the student is asked to provide information about the features of a recommended case. In the former, they must provide a specific value for a specific feature (e.g., show the electives whose prerequisite is DS207), while the latter requests that they simply express a directional preference for a particular feature value (e.g., show the electives that belong to higher year than the current recommendation or electives that have and increased perceived difficulty level). In contrast, ratings-based and preference-based feedback methods operate at the case level. In the former, the student is asked to rate or grade the recommended cases according to their suitability, while in the latter the student is simply asked to select one of the current set of recommendations that is

**Original printed page 119 · source PDF page 129 · Open source page**

closest to their requirements and circumstances (or farthest from their requirements/circumstances).

### 4.7 Validation

#### 4.7.1 Stability under change

The prototype system can now take existing cases from three different knowledge bases. The difference in each knowledge base is on the number of the permutations of the INT profiles. For example, Knowledge 15 has a number of 15 permutations of INT profile out of all 60 possible permutations ($_5P_3$).

When reducing the number of these permutations from 20 down to 15, the system still ranks the cases with exactly the same order as in Knowledge 20, excluding any cases that are no longer present in Knowledge 15.

Unfortunately, in order to test this effectively, the problem set should be developed with the highest number of INT permutations in mind. Upon reducing these permutations, the stability under change can then be tested for this criterion.

The current problem set was developed for Knowledge 20, and it can show the stability of the system when working with Knowledge 15. Another problem set is to be devised for Knowledge 47. This set can be used to test the system, given Knowledge 20 or Knowledge 15.

#### 4.7.2 Plausibility when the student’s perceived performance is lower than the course’s difficulty

To test this, the performance or PRFM attribute is set to either 2 or 1 to see the recommendation it affects against courses with difficulty level of 3.

**Original printed page 120 · source PDF page 130 · Open source page**

Prerequisites to all courses are assumed fulfilled. That is, no blocking on unfulfilled prerequisites will occur in this test.

**Original printed page 121 · source PDF page 131 · Open source page**

The difficulty vector (D1, O2, OM3, M4, NM5, N6, NS7, S8) is set to (1, 2, 1, 3, 2, 3, 1, 2).

The format of the attributes vector is (DSP, OPTIC, MWAVE, NTWRK, SATEL, PRFM, CHRT).

#### 4.7.2.1 Test case (0, 2, 0, 3, 1, 2, 3)

Translation of the attributes vector:

- Interests: Networking, Optics, Satellite
- Performance: 2
- Absolute specialisation

A logical advice would be to recommend N6, NS7, NM5, initially. But, N6 has a difficulty of 3. The final logical advice should be NS7, NM5, O2. The CBR system recommends NS7, OM3, and O2, with NM5 available as the fourth choice.

#### 4.7.2.2 Test case (0, 3, 0, 1, 2, 1, 1)

Translation of the attributes vector:

- Interests: Optics, Satellite, Networking
- Performance: 1
- Minimum specialisation and more diversity across the fields

The initial logical advice is O2, S8, and N6. But, all of these are more difficult than the student's perceived performance, so the next advice is OM3, NS7, NM5. NM5 has a difficulty level higher than 1, so the final advice can be OM3, NS7 and D1 (as the only other course, M4, is level 3 difficult). The prototype recommends NS7, OM3, and D1, which is a very close match to the logics.

**Original printed page 122 · source PDF page 132 · Open source page**

#### 4.7.2.3 Test case (3, 2, 0, 0, 1, 2, 2)

Translation of the attributes vector:

- Interests: DSP, Optics, Satellite
- Performance: 2
- Some specialisation

The first logical advice would be D1, O2, and OM3. Given that all of these courses are within the student's perceived capability, this advice is also final. The system gives very similar advice of D1, OM3, and O2 as well.

#### 4.7.2.4 Test case (0, 3, 1, 2, 0, 1, 3)

Translation of the attributes vector:

- Interests: Optics, Networking, Microwave
- Performance: 1
- Absolute specialisation

The logical advice begins with O2, OM3, and N6. But, O2 and N6 are deemed too difficult for the student, so the next advice is OM3, NM5, and M4. But, NM5 and M4 are still too difficult. The final advice would be OM3, NS7, and D1 (as the other remaining course, S8, is also too difficult). The system also recommends a perfectly matched advice of OM3, NS7, and D1 in the same ranking as well.

#### 4.7.2.5 Test case (0, 1, 0, 3, 2, 1, 3)

Translation of the attributes vector:

- Interests: Networking, Satellite, Optics

**Original printed page 123 · source PDF page 133 · Open source page**

- Performance: 1
- Absolute specialisation

The first human advice would be N6, NS7, and NM5. Because N6 and NM5 are too difficult, the second advice is NS7, S8, and O2. But, S8 and O2 are also deemed too difficult. The last advice would be NS7, and any order between D1 and OM3 (as the other course, M4, is also too difficult). The system ranks NS7, OM3, and D1, which is very reasonable.

#### 4.7.2.6 Test case (0, 1, 3, 0, 2, 2, 3)

Translation of the attributes vector:

- Interests: Microwave, Satellite, Optics
- Performance: 2
- Absolute specialisation

Initially, the human advice would be M4, OM3, and NM5. But, M4 has a difficulty level of 3. The next and final advice would be OM3, NM5, and S8. The prototype system recommends OM3, NM5, and NS7 (with S8 as a tie with NS7, but it has a lower sub-ranking score).

#### 4.7.2.7 Test case (0, 0, 1, 2, 3, 2, 1)

Translation of the attributes vector:

- Interests: Satellite, Networking, Microwave
- Performance: 2
- Minimal specialisation with more diversity across fields

The best logical advice would be NS7, NM5, and OM3. The CBR system recommends NS7, NM5, and S8 (with OM3 as the next available choice).

**Original printed page 124 · source PDF page 134 · Open source page**

#### 4.7.2.8 Test case (1, 2, 0, 0, 3, 1, 2)

Translation of the attributes vector:

- Interests: Satellite, Optics, DSP
- Performance: 1
- Some specialisation

The first advice would be S8, O2, and D1. Because S8 and O2 have higher difficulty level than the student's performance, the next advice would be NS7, OM3, and D1. The system also gives exactly the same advice in both choices and their ranking order.

From the comparison table D-1 (Appendix D), the improvement version has performed reasonably better than the other two algorithms, in both overall matching percentage and the ranking, as seen in table 4-5.

### 4.7.3 Plausibility when the prerequisites of the recommended courses have not yet been fulfilled.

To test this, the prerequisites of certain courses will have the status of unfulfilled (i.e., it has a value 0 in the prerequisite vector). Out of the test cases below, the first four cases will have the prerequisites vector (D1, O2, OM3, M4, NM5, N6, NS7, S8) set to (1, 0, 1, 1, 1, 1, 1, 0). The remaining four cases will be tested against the prerequisites vector of (1, 1, 0, 1, 0, 1, 0, 1). The PRFM attribute will be 3 so the recommended courses are not affected by the level of difficulty. The format of the attributes vector is (DSP, OPTIC, MWAVE, NTWRK, SATEL, PRFM, CHRT).

#### 4.7.3.1 Test case (0, 2, 0, 3, 1, 3, 3)

Translation of the attributes vector:

- Interests: Networking, Optics, Satellite

**Original printed page 125 · source PDF page 135 · Open source page**

- Performance: 3
- Absolute specialisation

The logical advice would be to recommend N6, NS7, NM5. The CBR system recommends NS7, OM3, and N6, with NM5 available as the fourth choice.

#### 4.7.3.2 Test case (0, 3, 0, 1, 2, 3, 1)

Translation of attributes vector:

- Interests: Optics, Satellite, Networking
- Performance: 3
- Minimum specialisation and more diversity across the fields

The initial logical advice is O2, S8, and NM5. But, O2 and S8 are blocked by unfulfilled prerequisites. The next advice would be OM3, NS7, and NM5. The system recommends NS7, OM3, and NM5, which is sensible under the circumstances.

#### 4.7.3.3 Test case (3, 2, 0, 0, 1, 3, 2)

Translation of the attributes vector:

- Interests: DSP, Optics, Satellite
- Performance: 3
- Some specialisation

The first logical advice would be D1, O2, and OM3. But, O2 is blocked by prerequisite incompletion. The second advice would then be D1, OM3, and NS7. (Remember that S8 is also blocked.) The system gives exactly the same advice of D1, OM3, and NS7.

**Original printed page 126 · source PDF page 136 · Open source page**

#### 4.7.3.4 Test case (0, 3, 1, 2, 0, 3, 3)

Translation of the attributes vector:

- Interests: Optics, Networking, Microwave
- Performance: 3
- Absolute specialisation

The logical advice begins with O2, OM3, and N6. But, the prerequisite of O2 has not been fulfilled yet. The next and final advice would OM3, N6, and NM5. The system recommends OM3, M4, and NS7, with D1, NM5 and N6 as subsequent choices. The discrepancy may be result from the fact that the knowledge base makes up less than 10% of the overall possible cases.

#### 4.7.3.5 Test case (0, 1, 0, 3, 2, 3, 3)

Translation of the attributes vector:

- Interests: Networking, Satellite, Optics
- Performance: 3
- Absolute specialisation

The first human advice would be N6, NS7, and NM5. Because the prerequisites of NS7 and NM5 have not been fulfilled, the second and final advice is then N6, S8, and O2. The system gives exactly the same advice of N6, S8, and O2 as well.

#### 4.7.3.6 Test case (0, 1, 3, 0, 2, 3, 3)

Translation of the attributes vector:

- Interests: Microwave, Satellite, Optics
- Performance: 3

**Original printed page 127 · source PDF page 137 · Open source page**

- Absolute specialisation

Initially, the human advice would be M4, OM3, and NM5. But, OM3 and NM5 are blocked due to incomplete prerequisites. The subsequent advice would be M4, S8, and NS7. Given that NS7 is also blocked, the final advice would be M4, S8, and O2. The prototype system recommends M4, S8, and D1 (with M4 available as the next ranked choice). Diversity (or lack of it) within the knowledge base may have caused this difference.

#### 4.7.3.7 Test case (0, 0, 1, 2, 3, 3, 1)

Translation of the attributes vector:

- Interests: Satellite, Networking, Microwave
- Performance: 3
- Minimal specialisation with more diversity across fields

The best logical advice would be NS7, NM5, and OM3. But, all of these courses are blocked because none of the prerequisites have been fulfilled. The next and final advice would be S8, N6, and M4. The CBR system recommends S8, N6, and D1 (with M4 available as the next ranked choice). Diversity (or lack of it) within the knowledge base may have caused this difference.

#### 4.7.3.8 Test case (1, 2, 0, 0, 3, 3, 2)

Translation of the attributes vector:

- Interests: Satellite, Optics, DSP
- Performance: 3
- Some specialisation

The first advice would be S8, O2, and OM3. Because OM3 is blocked for not having the prerequisite fulfilled, the next advice would be S8, O2, and D1. The system also recommends S8, O2, and D1 in exactly the same order.

From the comparison table D-2 (Appendix D), the improvement version has clearly outperformed the other two algorithms, in both overall matching percentage and the ranking as seen in table 4-5.

Table D-3 uses the trivial Information Retrieval algorithm to calculate the similarity percentage between the output of each algorithm and the expected recommendation. For each of the case, each algorithm is given a discrete value between 1 and 3, where 3 denotes the best match for that particular case. The matching percentage is then totalled to yield to overall matching percentage. The individual case ranking is also summed up to show the final ranking among three algorithms, where the higher the value is, the more efficient that algorithm can make recommendation.

**Figure 4-5: Correlation and Ranking table to show the efficiency of improvement algorithm over earlier versions — comparison between algorithm when performance is lower than difficulty**

| Correlation: Trivial | Correlation: Euclid | Correlation: Improve | Ranking: Trivial | Ranking: Euclid | Ranking: Improve |
|---|---|---|---|---|---|
| 0.845 | 0.655 | 0.845 | 3 | 1 | 3 |
| 1.000 | 1.000 | 1.000 | 3 | 3 | 3 |
| 1.000 | 1.000 | 1.000 | 3 | 3 | 3 |
| 1.000 | 1.000 | 1.000 | 3 | 3 | 3 |
| 0.964 | 0.964 | 1.000 | 2 | 2 | 3 |
| 0.964 | 0.964 | 0.964 | 3 | 3 | 3 |
| 0.802 | 0.926 | 0.964 | 1 | 2 | 3 |
| 1.000 | 1.000 | 1.000 | 3 | 3 | 3 |
| 7.574 | 7.508 | 7.772 | 21.000 | 20.000 | 24.000 |

**The last row above is the printed total.**

**Figure 4-5: Correlation and Ranking table to show the efficiency of improvement algorithm over earlier versions — comparison between algorithm when prerequisites are not yet fulfilled**

| Correlation: Trivial | Correlation: Euclid | Correlation: Improve | Ranking: Trivial | Ranking: Euclid | Ranking: Improve |
|---|---|---|---|---|---|
| 0.886 | 0.756 | 0.802 | 3 | 1 | 2 |
| 0.756 | 0.886 | 0.964 | 1 | 2 | 3 |
| 1.000 | 1.000 | 1.000 | 3 | 3 | 3 |
| 0.655 | 0.655 | 0.802 | 2 | 2 | 3 |
| 0.802 | 0.655 | 1.000 | 2 | 1 | 3 |
| 0.655 | 0.655 | 0.964 | 2 | 2 | 3 |

**Original printed page 129 · source PDF page 139 · Open source page**

#### 4.7.4 Plausibility when both the student's performance is lower than the course's difficulty AND the prerequisites of the recommended courses have not been fulfilled

Both tests above are combined here to see the effect on the overall plausibility and sensibility of the given academic advices.

##### 4.7.4.1 Test case (3, 0, 1, 2, 0, 2, 3)

The first test case is when the problem profile (DSP, OPTIC, MWAVE, NTWRK, SATEL, PRFM, CHRT) of (3, 0, 1, 2, 0, 2, 3).

The advisor will recommend D1, N6, and NM5. Now, N6 is deemed too difficult (difficulty level 3) for this student (PRFM = 2). The next three courses NS7 (prerequisite not fulfilled), M4 (also deemed to be too difficult), are OM3 (prerequisite not fulfilled) are not recommended. This leaves either O2 or S8 as the final choice. Using the improvement algorithm, the prototype system recommended NM5, D1, and S8, which corresponds to the logic of an academic advisor.

##### 4.7.4.2 Test case (0, 0, 3, 2, 1, 2, 2)

Another test for stability under small change is to assign a prerequisite vector for the given courses. For testing purposes, make the prerequisites for courses associated with multiple disciplines (e.g., OM3 and NS7) not yet fulfilled. The prerequisite vector of (D1, O2, OM3, M4, NM5, N6, NS7, S8) will have the values (1, 1, 0, 1, 1, 1, 0, 1).

Here is a test problem. (See http://cbr.tanat.net/match.php?table=3&DSP=0&OPTIC=0&MWAVE=3&NTWRK=2&SATEL=1&PRFM=2&CHRT=2)

**Original printed page 130 · source PDF page 140 · Open source page**

A student who is interested in Microwave, Networking, and Satellite (in order of preference), has an average academic performance, and wants some specialisation in his fields will have a profile (DSP, OPTIC, MWAVE, NTWRK, SATEL, PRFM, CHRT) as (0, 0, 3, 2, 1, 2, 2).

Now, as an academic advisor, the instant advice given will be the courses M4, NM5, and N6. Taking into account that the perceived difficulties (difficulty level 3) of M4 and N6 are probably beyond the student's academic performance (PRFM = 2), the advisor will skip these two courses and recommend NM5, OM3, and NS7. Because the prerequisites of OM3 and NS7 have not yet been fulfilled, the advisor will find another two courses to substitute these. Thus, the final recommendation will be NM5, S8, and either D1 or O2.

The prototype system using the improvement matching algorithm actually recommends a very similar set of courses, which contain NM5, O2 and D1, with a possibility of S8. Now, using commonsense, S8 should have a higher precedent over D1 and O2. However, keep in mind that pure CBR depends entirely on the quality of cases in its knowledge base. Currently, the number of cases is less than 10% of the overall number of every possible permutation. In addition, no rules are in place to double check the final advice against the preference in the fields of interest. We can use such rules to detect that S8 is within the student's field of interests and D1 and O2 are not, thus adjusting the ranking of S8. This last process can be integrated into the refine phase in the CBR cycle.

### 4.8 Framework

#### 4.8.1 Characteristics of Case-Based Academic Advisement Recommender Systems

The increasing scarcity of human academic advisement resources makes it difficult for students to readily access academic advisement services regardless of time and place. Academic advisors use their own experiences to solve most of the student cases that are queried to them. However, as student cases become more complex, the advising decision experiences that academic advisors are experts in tend to diverge - academic advising experience is distributed among experts and

**Original printed page 131 · source PDF page 141 · Open source page**

the advising experiences do not necessarily overlap. Nevertheless, when a student has a problem, he or she wants it solved as soon as possible. If that expert advisor is not available, the student has to wait, which is annoying and creates a negative image for the school or department. Advising experience must be available to every student at all times. Since academic advising experience can be deemed an educational institution asset, it has to be collected, preserved, and used as efficiently and effectively as possible. The best way to achieve this is to use CBR to develop academic advisement systems.

The goal of developing a case-based academic advisement recommender system is to create a knowledge repository that contains academic advising experiences for a complex advising domain that changes over time. This knowledge repository will be used in an institution by a group of people with varying levels of expertise. It is obvious that the development and use of such a system does not only involve technical processes, but also raises managerial and organisational issues as well. The tasks that should be performed to develop a case-based academic advisement recommender system and the processes that have to be put into place to make such a system operational are described.

Suppose that students are thought of as customers and academic advisors as salespeople. Academic advice or "recommendations" by academic advisors are always at least based on knowledge about the availability of courses, interest in the fields of study, the student's academic performance, and specialisation behaviour. A concept of "recommender agents" was developed in an attempt to model the behaviour of salespeople. Such agents draw on previous results from CBR (among machine learning and other advances in AI technology) to develop user models, and anticipate and predict user preferences. Given its characteristics, this concept can likely be applied to model the behaviour of academic advisors [30].

**Original printed page 132 · source PDF page 142 · Open source page**

## 4.8.2 Development and Use of Case-Based Academic Advisement Recommender Systems

### 4.8.2.1 Process Types

The table 4-1 below outlines the different processes, namely managerial processes, organisational processes, and technical processes, over the two phases, namely system development and system use.

**Table 4-1: Process Types**

| | | **System Development** | **System Use** |
|---|---|---|---|
| Managerial Processes | | Goal Definition<br>Awareness Creation<br>CBR-Tool Selection | Progress Verification and Controlling |
| Organisational Processes | | Project Team Selection<br>Initial Domain Selection<br>Project Team Training<br>Knowledge Acquisition Process Development<br>Utilisation Process Development | End-User Training<br>Continuous Knowledge Acquisition<br>Utilisation Process |
| Technical Processes | General IT-System Related | System Specification<br>System Implementation<br>System Integration<br>System Verification | Continuous System Maintenance |
| | Knowledge Repository Related | Initial Knowledge Acquisition<br>Core Knowledge Acquisition | Continuous Knowledge Acquisition and Maintenance |

Original printed page 133 · source PDF page 143 · Open source page

#### 4.8.2.2 Managerial Processes during System Development

##### 4.8.2.2.1 Goal Definition

For a case-based academic advisement recommender system project to be successful, precise goals must be determined at the outset. This enables management to fix the direction in which the project should develop and to measure the success of the project upon completion. Hard (quantitative) and soft (qualitative) success criteria should be identified (Stolpmann & Wess, 1999).

Hard criteria are measurable quantities and cover aspects like:

- Problem solution quality (advice soundness, consistency)
- Process quality (average time needed to solve a student case)
- Organisational quality (speed up in new academic advisor training, flexibility of staffing, cost per interaction)

In contrast, soft criteria measure the subjective quality of the academic advisement:

- Student satisfaction (availability of the academic advisors, perceived competence, friendliness)
- Academic advisor satisfaction (workload, work atmosphere, repetitiveness of tasks, intellectual stimulation)
- Institution aspects (preservation of knowledge, publicity)

The goal must be communicated to the project team, and the team has to be motivated to achieve them. When project goals are selected, it is important that these goals are realistic in terms of their time frame and whether they can be achieved with an acceptable amount of resources. The case-based academic advisement recommender project must be seen as part of the long-term knowledge management strategy for the institution. Since knowledge increases and evolves, the experience in a CBR system must be maintained continuously. System development is only the initial phase in any CBR project.

Original printed page 134 · source PDF page 144 · Open source page

##### 4.8.2.2.2 Awareness Creation and Motivation

The case-based academic advisement recommender system project targets the most precious asset of human academic advisors: their experience. The project's goal is to collect the academic advising experience of each relevant academic advisor and make it available to whoever needs it in the institution.

Obviously, the academic advisors will have a motivational barrier to giving away their experience. Every academic advisor knows that "knowledge is power". In academic advisement environments or domains where experience is being used to solve problems, having experience translates into being superior and indispensable, whereas giving away the knowledge can be perceived as becoming obsolete.

However, as soon as academic advisors become part of a project team and understand that sharing knowledge means that they will get back much more than they invest, most barriers disappear. It has to be made clear that the user and beneficiary of the developed system will be themselves and not an anonymous party. They will be able to access the experiences of their colleagues and advise on cases they might not have been able to advise on before, as well as end situations in which colleagues constantly pester them for advice. The resulting academic advisement system will enable them to work with increased efficiency and effectiveness.

Apart from the academic advisors, management and administration has to be motivated as well. CBR is perceived to be "academic" by most managers and administrators. While investing resources into a database project seems to be no problem for them, investing into CBR is investing into a venture with an uncertain outcome. It has to be clarified that CBR is an established technology and by no means only an academic “playground”. It also has to be clarified that the initial installation of the case-based academic advisement recommender system is only the beginning of a process that will enable the institute to capture and reuse experience. Management and administrators must be prepared to invest resources on a continuous basis while the system is operational. A CBR system is useful only if it contains knowledge and it is maintained on a continuous basis. Without continuous management support and advisors who are willing to fill and use the system, any CBR project is bound to fail.

**Original printed page 136 · source PDF page 145 · Open source page**

#### 4.8.2.2.3 CBR Tool Selection

Earlier, the author described the prototyping of a proof of concept case-based academic advising system. However, each institute has the choice of using the existing tools with the cost of significant customisation. Based on the project, domain, and user-group specifications, a suitable tool to develop the case-based academic advisement recommender system may be chosen instead. Criteria to be taken into account include:

- The operating environment in which the system is going to be used (hardware and software, network architecture, database type, etc.)
- The complexity of the technical domain (home appliances or networked workstations)
- The level of experience of the students and the academic advisors
- The organisation of the academic advisors (study programs under responsibility, physical locations, etc.)
- The project goals that have been defined

Since the case-based academic advisement recommender system is going to serve as a long-term knowledge repository for the institution, this selection should be based not only on technical criteria, but also should take into account economic and institutional considerations, as well as the strategic decisions of the institute.

### 4.8.3 Tweaking

There is a mode of operation, namely navigation by browsing or navigation by proposing, whereby a student can be presented with a group of cases during each recommendation cycle and the next cycle can be based on the feedback provided [29]. The student can also be asked to express feedback in the form of a tweak or a set of tweaks. In short, the student can implicitly select one of the presented cases and express directional preferences for certain features. For instance, in the Academic Advisement system, the student can be presented with the single best case during each recommendation cycle and introduce tweaks over features such as subject

**Original printed page 137 · source PDF page 146 · Open source page**

difficulty (find me easier or more difficult subjects) or teaching style (find me subjects with lectures, workshops, or online delivery).

In a sense, tweaking is a combination of navigation by asking and navigation by proposing – sequences of cases can be proposed and the student can be asked for specific information about particular features. However, the cost to or involvement of the student should be lower than value elicitation because the student is not expected to provide specific feature values of which they may have little or no knowledge. This strategy can also combine elements of preference-based feedback since the student must explicitly identify a preferred case over which the tweaks should be applied.

### 4.8.4 Ratings-based

Recently, ratings-based feedback has become a popular form of feedback in recommender systems. The basic idea is to ask the student to rate individual recommendation proposals - the student provides case-level feedback. For example, we can design an Academic Advisement recommender system to compile personalised subject lists for students based on their learned fields of interest and, as such, it is involved in recommending subjects to students. We can probably encourage students to rate the system's subject recommendations (using a 5-point scale from strong-negative to strong-positive) as a means of fine-tuning subsequent recommendation cycles.

Generally, ratings-feedback is a relatively low-cost form of feedback since the student need only express a qualitative (or perhaps quantitative) indication of interest at the level of an individual case. The level of effort naturally increases if the student needs to rate many or all of the recommended items, but we should allow students to choose to rate as few or as many of the subjects as they desire.

**Original printed page 138 · source PDF page 147 · Open source page**

#### 4.8.5 Preference-Based

Perhaps, the lowest cost form of feedback may be for the student to express a simple positive or negative preference for one of the recommended cases. This form of feedback can be viewed as a subset of tweaking in the sense that the student is just asked to express an individual positive or negative preference without needing to indicate specific feature tweaks. Similarly, it has a lower cost than ratings-based feedback because the student is not expected to distinguish between different levels of preference.

# Chapter 5 Conclusion and Future Works

**Original printed page 139 · source PDF page 148 · Open source page**

## 5.1 Conclusion

Case-based reasoning (CBR) has been an active area of research for the past two decades. It was proposed and implemented as a cognitive model by artificial intelligence researchers and then applied as a model for building expert systems. Since the early 1990s, several popular educational applications and innovations have been developed from the lessons learned from CBR research, although none were directly related to academic advising.

This thesis designs, develops, and validates a prototype case-based academic advisement recommender system that can be used to conduct online academic advising. A summary of the features of the prototype academic advisement recommender system, including its features, is presented in this chapter. It concludes with a list of future enhancements for the further development of the case-based academic advisement recommender system.

Although the prototype system is small, it serves its purpose well as a proof of concept. The prototype system was built for the normal case of students finishing one academic year and advancing to the next. The cases only take into account “clear pass” students (i.e., students not failing any courses). It also only considers a single stream of a program with multiple streams (majors).

None of these are serious limitations. The test cases are quite general and valid for students advancing from any year (n) of the program to the subsequent year (n+1). As the system can also enforce prerequisites, it can handle students who have failed one or more courses in the current year. The extension to multiple streams or majors is also quite straight forward when, as for Electrical and Computer Engineering at RMIT, the program has a common structure and majors are based simply on the selection of a subset of core courses for that major in each year level.

**Original printed page 140 · source PDF page 149 · Open source page**

Expanding the system to cover all possible student situations in such a program primarily involves adding sufficient cases to cover all of these student situations, to give an academic advisement recommender system with only one universal case base.

To develop case-based academic advisement recommender systems that will be used in dynamic, institution environments by a large group of users, one must take into account the managerial (or administrative), organisational (or institutional), and technical processes. It has to be remembered that once a CBR system is in place, continuous knowledge acquisition and maintenance is necessary. Processes for knowledge acquisition and maintenance have to be developed and executed, and personnel have to be dedicated to performing these tasks. A non-technical framework that defines these processes can be found in the appendix.

Some findings are briefly outlined below. Two distinct expert system techniques should be used. Rule-based reasoning should be used for the selection of core (non-elective) courses and filtering out the courses whose prerequisites have not yet been fulfilled. However, CBR should be used for the advisory functions, in particular the selection of electives. Given the set of attributes of the problem student case or profile, sound and reasonable academic advice will be given in the form of recommended courses for that particular student to enrol in.

The standard CBR methodology should be used as follows:

- Retrieve similar cases to the current student profile in question
- Reuse a set of courses recommended by a similar case

- Revise or adapt that solution to better fit the new student profile in question by collating the ranked results of similar cases
- Retain the new set of recommended courses

The most common set of attributes in each student case are the fields of interest, popularity of the course, pass rate and high marks rate, suitability with the current industry trend, reputation of the lecturer, type of assessment, and specialisation preference.

**Original printed page 141 · source PDF page 150 · Open source page**

For the purpose of prototyping, only the fields of interest, the student's academic performance, and the student's specialisation preference are used to demonstrate the underlying ideas and concepts.

Some exceptional cases that could happen to students:

- Student is in 3rd year and has already chosen a major. This student performed well in the minor stream courses, but did not perform well in the core major courses. The student would then like to change their major half way. This can be addressed by utilising the minimum-time rule to determine what courses they must take to complete in the minimum time under the new major.
- Student failed one of the core courses, which is a prerequisite to many subsequent courses. The student also progresses normally in other series of courses. At one point, the student may have to enrol in courses that span more than three different academic years. If desired, a maximum-span rule could be added to limit the number of academic years spanned in a single enrolment year.

To efficiently index a case and retrieve it in a timely manner, the prototype case-based academic advisement recommender system has a small case base. At this point, as a proof of concept, a 1-dimensional flat structure of case indexes is used.

Features in the prototype case-based academic advisement recommender system:

- Advise student which courses to take next semester
- Input and update information on course prerequisites by editing the backend scripts
- Input and update information on the academic year requirements, also by editing the backend scripts

Advantages:

- Easy access to the system from anywhere using a Web browser
- Reduces advising traffic to academic advisors
- Saves time for both staff (academic advisors) and students (advisees)

**Original printed page 142 · source PDF page 151 · Open source page**

Disadvantages:

- Even when this system is eventually developed as a live, fully-fledged production version, no system is a perfect system. Therefore, the results should not be taken as final. They should be crosschecked with an academic advisor
- Minimises student-advisor interaction, which may in turn affect the opportunity of advisors to keep themselves up-to-date with individual progress at a personal level
- Key decisions cannot be taken using the system
- System may not provide answers to all possible questions
- Systems may not provide support in special or exceptional cases like drop outs, incomplete, and failures. Need to contact academic advisors for decision making in these extraordinary cases

• System does not take into consideration students’ strengths and weaknesses while advising courses

### 5.2 Future Work

A number of improvements necessary to take the system from a prototype to a working system have been identified in the previous section. Further improvements to the system would require more research into existing and emerging variations in CBR methodology, as well as exploring variations in the content and structure of the cases for this system. Some of these are outlined below.

#### 5.2.1 Case Representation

There are many ways of representing a case, including predicate representations, frames, or even entries resembling database records. The prototype system developed here uses a simple representation, as only a small set of attributes are used for a case. As more attributes are added to cases, an alternative representation may provide better performance.

Three general features need to be captured: the problem-situation description, the solution, and the outcome. The problem-situation is essentially a description of the characteristics of the

**Original printed page 143 · source PDF page 152 · Open source page**

problem in the context in which it occurs. The representation must match the reasoning goals of the system. However, it is not always obvious what a case should be. A problem description may need to include features that are intermediary, or a solution-in-progress. A description, after some reasoning, may need to be redescribed to reflect the situation more accurately (for example, the condition of a patient as the disease progresses). Most of the difficulty comes when information is time-based, or changes over time. Current methods of representing cases are mostly static - how should the case representation be made dynamic in order to reflect changes? Should a situation be represented as one large monolithic case, or distributed cases with portions related to each other? How can such distributed cases be retrieved and judged similar?

An important problem to be addressed is the question of when to admit a case as a new case and not as an adaptation of an existing case. In other words, what should be the extent of dissimilarity between two cases before they are considered two separate cases? Is it when they differ in a sufficient number of features, or is it when they differ only in significant features but differ significantly, or is it some combination or the two? Should there be a “metric of goodness” developed to decide whether a case is sufficiently different to be treated as a new case?

Another issue worth addressing is to understand whether it is important to know that a case was derived from another case. If such knowledge is important, should such derived cases be handled as versions of the initial case? Furthermore, should comparisons be made with the original cases only, or should similarities be tested against the derived ones as well? One can argue that it may be sufficient to test similarities against the most recent cases, as is normal in the fields of medicine and law. For such a position, one needs to address the potential pitfall of reaching the old case by modifying the most recent ones. In other words, how much of past cases need to be considered without worrying about reinventing some really old cases?

#### 5.2.2 Case Indexing

Indexing allows the cases to be retrieved selectively, quickly, and accurately. Identifying the relevant features to be indexed is not trivial, especially for indexes that are not just surface or

**Original printed page 144 · source PDF page 153 · Open source page**

shallow features. Retrieving the correct or relevant case is crucial to the success of CBR in solving new problems. Once cases are represented and indexed, they need to be stored and organised in an efficient structure for fast and accurate retrieval. Memory may be organised as a simple list or a discrimination or dependency graph in order to reduce the search effort, and the most efficient retrieval algorithm determined.

Another problem is the uncontrolled growth of the case base. For example in the main undergraduate program for the School of Electrical and Computer Engineering at RMIT, about 300 new students are enrolled each year. If only 10% of them result in new cases, that is 30 new cases each year. The total for all programs at RMIT would result in hundreds of new cases each year. How do we determine which cases to store and which are to be weeded out? Essentially, what makes a “good” case? Currently, maintaining the case library is carried out manually, with a case base administrator browsing through each case and deciding whether to keep it or remove it. While there are simple procedures (such as time stamping) to assist the administrator, it is still a time consuming process, especially with thousands of cases in the library.

### 5.2.3 Improving Case-Based Recommendation

Treating student profiles as transactions and the rated courses therein as item sets, the Apriori algorithm can be used to derive a set of course-course association rules and confidence levels. The resulting confidence values are taken as probabilities and used to fill in a course-course similarity matrix, which provides the additional similarity knowledge necessary to compare non-identical profile items [31].

Since the matter of additional similarity coverage rests in populating the matrix as densely as possible, two natural extensions suggest themselves. First, the directly generated rules can be changed together ($A \rightarrow B$ and $B \rightarrow C$ imply $A \rightarrow C$) to provide indirect course relationships (absolutely useful for inter-disciplinary electives).

Building such course-course similarity knowledge to address the sparsity/maintenance problem is similar in spirit to item-based collaborative techniques. The item-based collaborative

**Original printed page 145 · source PDF page 154 · Open source page**

approach uses rating overlaps to build course-course similarities, and suggested courses are then retrieved in direct comparison to the elements/measures/attributes that comprise a student profile. The direct nature of such item retrieval recalls direct content-based item recommendation, as well as the potential cost in terms of diversity.

The availability of course similarity knowledge facilitates a new type of similarity-based recommendation strategy that combines measures/attributes from case-based and collaborative recommendation techniques. It facilitates the use of more sophisticated CBR-like similarity metrics on ratings-based profile data, which in turn make it possible to leverage indirect similarities between profile cases, and so generate improved recommendation lists. This new recommendation strategy consists of two basic steps:

- The target profile, t, is compared to each profile case, $s_i$ is an element of S, to select the k most similar cases
- The courses contained within these selected cases (but absent in the target profile) are ranked according to the relevance to the target, and the r most similar courses are returned as recommendations

### 5.2.4 Case Adaptation and Learning

A CBR system learns through adaptation (i.e., modifying a retrieved case that partially matches the current problem) to solve a new problem. This means that over time new cases are acquired and the system’s problem solving performance should improve as it “learns”. Learning may be from positive examples given by the domain expert or from its own success (or even from its failure) in achieving the problem-solving goal. However, adaptation formulas are not always easy to define, even by a domain expert. How does the case-based reasoner know what to adapt and when? Can adaptation rules that are sufficiently generic (for all situations) be built so that the case-based reasoner is able to detect the differences between the new and retrieved case, decide what needs to be adapted, determine which adaptation strategy to choose (from several appropriate strategies), and project the likelihood of succeeding?

**Original printed page 146 · source PDF page 155 · Open source page**

In summary, some of the open issues and research challenges are in the areas of case representation, similarity matching, selecting indices and generating new indices dynamically, and case library organisation (as new cases are encountered and entered into the library), as well as mixing CBR with other paradigms to build hybrid systems.

### 5.2.5 Emerging Hybrid Systems

CBR hybrid systems capitalise on the inherent strengths of the other reasoning paradigms to counter the weaknesses in CBR and complement its strengths. The most common combination attempted at this point is combining rule-based reasoning (RBR) with CBR and model-based reasoning.

Academic Advice CBR systems solve new academic profiles/preferences by retrieving stored prior cases and adapting their advices to fit new student circumstances [25]. Traditionally, CBR systems draw their cases from a single local case base tailored to their task. However, when a system's own set of cases is limited, it may be beneficial to supplement the local case base with cases drawn from external case bases for related tasks. The effective use of external case bases requires strategies for multi-case base reasoning (MCBR):

- For deciding when to dispatch problems to an external case-base
- For performing cross-case base adaptation to compensate for differences in the tasks and environments that each case base reflects

### 5.2.6 CBR and Rule-Based Reasoning (RBR)

Rules of the form IF-THEN-ELSE cover a single aspect of knowledge, whereas a case covers a particular problem-solving situation. It is sometimes difficult for a domain expert to come up with rules to problem solving, whereas cases are actually examples of a particular situation. Hence, acquiring cases is generally easier than rules.

**Original printed page 147 · source PDF page 156 · Open source page**

RBR has some inherent weaknesses:

- Past experiences are not "learned"
- New, novel problems that deviate from the norm cannot be solved (because no rules are applicable)
- Maintaining a large rule base is difficult and time consuming

The wrong information in rules must be individually edited, and every time a new rule is added, there is the risk of redundancy or contradiction. Deleting a rule may interrupt the reasoning process, whereas removing an individual case does not have the same extensive consequences. Hence, RBR systems are more difficult to maintain and update to keep up with the rapidly changing domains.

Thus, most hybrid systems have a combination of rules and cases, and may be according to whether rules and cases are independent of each other, or whether one was derived from the other (i.e., derive cases from rules, or rules from a set of sample cases). Rules may be used to capture broad trends in the domain, while cases are used to support exceptions to the rules or violations of the rules. Cases may be kept as positive examples to support, explain, or justify rules, as counter-examples to rules, or when rules fail because they are ill defined. However, there are problems integrating cases and rules, and these have to do with control – which mode of reasoning to use, and when. To date, there is still a lack of cognitive models of how human experts integrate rules and cases.

The system proposed in this thesis assumes the use of RBR for the identification of core courses and perquisites, as these are easily defined by rules. However, the thesis concentrates on the use of CBR to help students select electives, which is not easily defined by rules. Essentially, RBR and CBR are used to solve different parts of the advising problem.

Original printed page 148 · source PDF page 157 · Open source page

### 5.2.7 Machine Induction

The academic advisement can exploit the approach to machine learning known as induction. Other case-based systems and data mining applications have made machine induction commercially useful (and viable).

ID3 [32] is a non-incremental algorithm, meaning it derives its classes from a fixed set of training instances. An incremental algorithm revises the current concept definition, if necessary, with a new sample. The classes created by ID3 are inductive, that is, given a small set of training instances, the specific classes created by ID3 are expected to work for all future instances. The distribution of the unknowns must be the same as the test cases. Induction classes cannot be proven to work in every case since they may classify an infinite number of instances. Note that ID3 (or any inductive algorithm) may misclassify data.

#### 5.2.7.1 Data Description

The sample data used by ID3 has certain requirements, which are:

- Attribute-value description - the same attributes must describe each example and have a fixed number of values.
- Predefined classes - an example's attributes must already be defined, that is, they are not learned by ID3.
- Discrete classes - classes must be sharply delineated. Continuous classes broken up into vague categories such as a metal being "hard, quite hard, flexible, soft, quite soft" are suspect.
- Sufficient examples - since inductive generalization is used (i.e. not provable) there must be enough test cases to distinguish valid patterns from chance occurrences.

Original printed page 149 · source PDF page 158 · Open source page

#### 5.2.7.2 Attribute Selection

How does ID3 decide which attribute is the best? A statistical property, called information gain, is used. Gain measures how well a given attribute separates training examples into targeted classes. The one with the highest information (information being the most useful for classification) is selected. In order to define gain, we first borrow an idea from information theory called entropy. Entropy measures the amount of information in an attribute.

Given a collection S of c outcomes

Entropy(S) = S -p(I) log2 p(I)

where p(I) is the proportion of S belonging to class I. S is over c. Log2 is log base 2.

Note that S is not an attribute but the entire sample set.

Machine learning can be applied to automate index assignment. Case attributes can be improved to assume ordered symbol sets. This can then be used with a symbolic learning algorithm, one that learns classification rules, which is more appropriate for comparatively small numbers of training examples of the attributes. ID3 algorithm can be implemented here, as it can learn decision trees in which rules are implicit. The improvement of the prototype system can employ ID3 to induce decision trees for case indexing. ID3 learns a decision tree by recursively partitioning the training set according to the feature that best discriminates instances of each case attribute. Each learned decision tree represents a number of rules for assigning an attribute in knowledge cases.

#### 5.2.7.2 Attribute Selection

How does ID3 decide which attribute is the best? A statistical property, called information gain, is used. Gain measures how well a given attribute separates training examples into targeted classes. The one with the highest information (information being the most useful for classification) is selected. In order to define gain, we first borrow an idea from information theory called entropy. Entropy measures the amount of information in an attribute.

Given a collection S of c outcomes

$$\text{Entropy}(S) = S\ -p(I)\ \log2\ p(I)$$

where p(I) is the proportion of S belonging to class I. S is over c. Log2 is log base 2.

Note that S is not an attribute but the entire sample set.

Machine learning can be applied to automate index assignment. Case attributes can be improved to assume ordered symbol sets. This can then be used with a symbolic learning algorithm, one that learns classification rules, which is more appropriate for comparatively small numbers of training examples of the attributes. ID3 algorithm can be implemented here, as it can learn decision trees in which rules are implicit. The improvement of the prototype system can employ ID3 to induce decision trees for case indexing. ID3 learns a decision tree by recursively partitioning the training set according to the feature that best discriminates instances of each case attribute. Each learned decision tree represents a number of rules for assigning an attribute in knowledge cases.

**Original source visual: printed page 149, PDF page 158.**

**Original printed page 150 · source PDF page 159 · Open source page**

### 5.3 Implications

Finally, the work done in this research (i.e., the design of cases) and the development of the prototype case-based academic advisement recommender system is no doubt an important building block towards the design and implementation of a full production system. Ultimately, a fully developed system could eliminate the need for human academic advisors. In turn, this would enable ubiquitous academic advising for distance online students who define the changing face of the higher education in the developed regions.

# Bibliography

**Original printed page 151 · source PDF page 160 · Open source page**

**Original printed page 152 · source PDF page 161 · Open source page**

**Original printed page 153 · source PDF page 162 · Open source page**

**Original printed page 154 · source PDF page 163 · Open source page**

**Original printed page 155 · source PDF page 164 · Open source page**

**Original printed page 156 · source PDF page 165 · Open source page**

# Appendix A Non-technical Framework

**Original printed page 157 · source PDF page 166 · Open source page**

## A.1 Organisational Processes during System Development

### Project Team and Initial Domain Selection

The creation of a project team to serve as the "knowledge engineers", and the selection of a group to serve as the initial test users of the system, are the first organisational steps that must be taken.

Apart from the person implementing the case-based academic advisement recommender system (CBR consultant), the project team should contain academic advising personnel that are very experienced in the relevant sub domain (e.g., different program of study) to be modelled and well respected by the academic advisors outside the project group. Personnel that are interested in "technology" rather than implementing useful systems are very common in academic environments, but definitely not helpful in creating a case-based system and should be avoided. Once selected, the members of the group should be kept constant (i.e., fluctuations should be avoided).

The group of initial users should comprise two types of academic advising personnel and students. The first type of academic personnel should be on a comparable level of expertise with the project team with respect to the selected sub domain (i.e., expert users), and the other should be academic advising personnel who are less familiar with the specific problem area (i.e., intermediate users). Finally, the students represent the users of the academic advice given (i.e., novice users). While the expert test users can communicate to the project group in their language, the intermediate users and the novice users will represent the target group for which the system is being implemented. Feedback from both types of users is required for a successful project. After a first "rapid prototype" has been implemented, the expert users can provide advice regarding problems with the knowledge modelled in the system. In contrast, the members

**Original printed page 158 · source PDF page 167 · Open source page**

of the intermediate user group will serve as models of the academic advisors that will use the system, while the members of the novice user group will test the user-friendliness of the self-service system. The vocabulary used to represent the cases and the knowledge contained within them has to be adjusted to the intermediate and novice user groups.

The domain selected for the initial knowledge acquisition is of the utmost importance. The domain should be representative of the student cases that are queried to the academic advisors in terms of complexity and frequency. It should also be a problem area that accounts for a considerable amount of the workload, as well as an area the academic advisors are interested in sharing (obtaining) knowledge about.

### Training the Project Team

Training the project team is an organisational process that has a major impact on the success of the help desk project. At the beginning of the project, the project team is usually inexperienced with respect to CBR and knowledge acquisition. Since the project group will be responsible for system maintenance and continuous case acquisition after the development has finished, it is very important that they are trained in CBR as well as in knowledge acquisition and modelling during the initial knowledge acquisition.

While the project team should also get advanced training to be able to model, fill, and maintain the knowledge in the system, the test users only need to be trained in using the resulting case-based academic advisement recommender system.

### Development of the Knowledge Acquisition and Utilisation Processes

The introduction and use of a case-based academic advisement recommender system usually causes a re-evaluation and modification of the existing knowledge and information management processes in an academic advising environment. When the system is used, it must be integrated into the operating environment of the academic advisors and become part of the standard business process. Existing processes must be altered to facilitate the flow of information to and

**Original printed page 159 · source PDF page 168 · Open source page**

from the CBR system. When the organisational processes are defined, the tasks to be performed, the personnel or roles to perform these tasks, and the communication among the groups/roles have to be fixed.

After the development of the case-based academic advisement recommender system is complete, it will serve as the central source of information for the academic advisors. To ensure a smooth flow of information, the knowledge sources and formats as well as the qualifications of the personnel that require the knowledge have to be analysed, and processes that allow the efficient and effective acquisition and use of knowledge have to be developed. One should keep in mind that while the group enacting the initial knowledge acquisition process is the project team and rather experienced, the users who use the system (both in terms of knowledge retrieval and continuous acquisition) may be less qualified.

It may useful to define four roles for the organisational processes during the use of the academic advisement system:

- The academic advisor
- The student
- The CBR author
- The CBR administrator

Academic advisors are the users from the target group. Their duty is to use the implemented academic advisement system in their daily work. If they cannot find an appropriate solution with the system, they will have to solve the student case on their own and generate a new case. Depending on the domain and managerial decisions, this new case may or may not be made immediately available as an “unconfirmed” case to the other academic advisors. For maintenance purposes, the academic advisors are also encouraged to comment on the quality and applicability of the cases in the case base.

The students are also the users from the target group. They should have access only to the self-service system, which gives advice based on their individual circumstances and preferences.

**Original printed page 160 · source PDF page 169 · Open source page**

They may be asked to provide feedback and ratings of the given advice, which in turn can be captured into the system as a “refined but unconfirmed” case.

The unconfirmed (new or refined) cases have to be verified in terms of their correctness and suitability for the case base by the CBR author(s). The CBR author is a person with experience both in the domain and in using the CBR system. While the CBR author can decide on the quality and inclusion of a case in the case base, he or she is not allowed to perform modifications on the vocabulary, similarity, or adaptation knowledge. These can only be performed by the CBR administrator.

The personnel enacting the roles of the CBR author(s) and the CBR administrator should be included in the project group from the start of the project. It should be noted that both these roles require a considerable amount of resources and should be performed by dedicated personnel. If the institution does not permit dedicating more than one person to these tasks, the duties of the CBR author and CBR administrator should be performed by one person.

## A.2 Technical Processes during System Development

### General IT-System Development Related Processes

The development of a case-based academic advisement recommender system is similar to any other IT project in most aspects. As usual, the system has to be specified, implemented, integrated, and verified. The definition of the requirements, implementation, and the testing and revision of the prototype and the actual case-based academic advisement recommender system are steps that have to be performed in accordance with standard software engineering techniques. However, the user interfaces and the connection to supporting programs (integration) are two features that require additional attention.

The essential task in developing a user interface is to present the relevant data at the right moment, in the right representation, and on a level of abstraction that is suitable for the current users of the system. If the available data is presented to the users in a way that they do not

**Original printed page 161 · source PDF page 170 · Open source page**

understand, in a representation they are not familiar with, or at a moment when it is irrelevant, it will only cause confusion and be of no use. The user interface of the case-based academic advisement recommender system has to be developed in accordance with the user group (i.e., academic advisor and student), the specific domain (e.g., program of study), and the institution's academic policies (i.e., who is allowed to see what kind of data). It has to present the right data, at the right moment, on the right level of abstraction, and in accordance with institution policies.

A case-based academic advisement recommender system cannot operate in isolation. While the CBR system will store experience, it will not contain data regarding student academic records, structures of programs of study, or other academic rules. The student information is stored in a student records database. The program structures information is usually stored in separate database. Since this information is needed during academic advising, the system must have interfaces to these databases.

### Initial Knowledge Acquisition for the Case-Based Academic Advisement Recommender System

A CBR system is useless without cases. When the case-based academic advisement recommender system is handed over to the academic advisors, it has to contain enough cases to cover at least part of the relevant present student cases. Otherwise, the system will be considered useless and the project will fail. Initial knowledge acquisition serves three major goals:

- Training the project team in knowledge acquisition
- Initialising the knowledge in the system
- Collecting enough help desk cases to bootstrap the system

During initial knowledge acquisition, the knowledge in the system can be distributed among the domain model (vocabulary), similarity measure, adaptation knowledge, and the case base. These knowledge containers (Richter, 1998) have to be created and filled. While the similarity measures, adaptation knowledge, and vocabulary are compiled knowledge, the case is interpreted at runtime. In principle, each container could be used to represent most of the knowledge. However, this is obviously not very feasible, and the CBR consultant should carefully decide on

**Original printed page 162 · source PDF page 171 · Open source page**

the distribution of knowledge into the containers. After the initial knowledge acquisition is completed, the distribution is more or less fixed and should only be changed with caution.

The processes for the acquisition of knowledge for each container is run in parallel and cannot be easily separated during the initial knowledge acquisition. Since the vocabulary lays the ground for entering the cases and describing the similarity measures and adaptation knowledge, it has to be available first.

However, to be able to create a domain model (i.e., the vocabulary) one has to understand how the domain is structured, and this can only be done by looking at the cases, similarities, and adaptation rules.

A good way to approach this problem is to create and use standardised forms to acquire an initial amount of cases from the project team [HOMER system]. The form should be developed in cooperation with the project team. The first thing that must be done is to ask the project team to fill out as many case acquisition forms as they can. By looking at the elements of the forms, the vocabulary (i.e., the phrases that have to be used and the domain structure) can be derived and a vocabulary that is capable of describing the cases on the forms can be modelled.

By asking the project team to define the range of possible values for each attribute on the forms, and inquiring what would have happened if one of the values on a form were different, a broad range of cases can be created and the vocabulary expanded in a short time. Discussion among the project team members raises the level of understanding of both the approach and the problems, and it should be encouraged in this early phase. One should also keep in mind that the goal is not to model the domain in every detail, but on a level that helps the system's user solve a problem (i.e., the system does not have to solve the problem autonomously). Especially during initial knowledge acquisition, it is advisable to have more cases on an "everyday" level, rather than having a few extremely specific ones.

**Original printed page 163 · source PDF page 172 · Open source page**

While the initial vocabulary is being created and value ranges fixed, questions regarding adaptation rules and similarities should be posed and the results entered into the system.

One of the major challenges one must face when creating a system to capture and represent the experience of domain experts is determining the level of abstraction with which the domain and the knowledge will be modelled. If the model is too simplistic, it will cause problems while the experience is being captured and miss important details. If, however, the domain model is too specific, the user will get lost quickly in useless details and knowledge acquisition will be very tedious and time consuming. Maintenance is very difficult for both a too simplistic and a too complex model.

The decision to use a structured domain model approach as opposed to a textual query-answer-based approach also depends on the system's intended users. For inexperienced academic advisors or students, a tool with which simple student cases can be given advice by answering a limited number of questions is of great value (Thomas et al., 1997). However, for experienced academic advisors who would not bother to use a system for (subjectively) trivial problems, a structured domain model approach may be more appealing. The system will be able to present the not-so-obvious similar solutions that the academic advisors could not find. Since knowledge contained in the domain model is used in similarity calculation, the retrieved solutions will be similar in a semantic and structural manner. The domain model allows the solutions in the case base to be applicable to a broader range of problems.

# Appendix B Knowledge Base Contents

**Original printed page 164 · source PDF page 173 · Open source page**

```
DSP:OPTIC:MWAVE:NTWRK:SATEL:PRFM:CHRT:D1:O2:OM3:M4:NM5:N6:NS7:S8
```

```
0001231300000111
0001232200001011
0002131100001011
0003122100101010
0003122300111000
0003123200011010
0010231300000111
0010322100101010
0012033200100011
0012301200101100
0013202100101010
0020133301000011
0020311301000110
0020312200100110
0021033100101010
0021303100101010
0023102301110000
0030211201100010
0032013301110000
0032102301110000
0100323110001010
0102031300010011
0102302200101100
0103023110001010
0120032301000011
0120301300001110
0130022201100010
```

## Appendix B: Knowledge Base Contents

DSP:OPTIC:MWAVE:NTWRK:SATEL:PRFM:CHRT:D1:O2:OM3:M4:NM5:N6:NS7:S8

0 0 0 1 2 3 1 3 0 0 0 0 0 1 1 1
0 0 0 1 2 3 2 2 0 0 0 0 1 0 1 1
0 0 0 2 1 3 1 1 0 0 0 0 1 0 1 1
0 0 0 3 1 2 2 1 0 0 1 0 1 0 1 0
0 0 0 3 1 2 2 3 0 0 1 1 1 0 0 0
0 0 0 3 1 2 3 2 0 0 0 1 1 0 1 0
0 0 1 0 2 3 1 3 0 0 0 0 0 1 1 1
0 0 1 0 3 2 2 1 0 0 1 0 1 0 1 0
0 0 1 2 0 3 3 2 0 0 1 0 0 0 1 1
0 0 1 2 3 0 1 2 0 0 1 0 1 1 0 0
0 0 1 3 2 0 2 1 0 0 1 0 1 0 1 0
0 0 2 0 1 3 3 3 0 1 0 0 0 0 1 1
0 0 2 0 3 1 1 3 0 1 0 0 0 1 1 0
0 0 2 0 3 1 2 2 0 0 1 0 0 1 1 0
0 0 2 1 0 3 3 1 0 0 1 0 1 0 1 0
0 0 2 1 3 0 3 1 0 0 1 0 1 0 1 0
0 0 2 3 1 0 2 3 0 1 1 1 0 0 0 0
0 0 3 0 2 1 1 2 0 1 1 0 0 0 1 0
0 0 3 2 0 1 3 3 0 1 1 1 0 0 0 0
0 0 3 2 1 0 2 3 0 1 1 1 0 0 0 0
0 1 0 0 3 2 3 1 1 0 0 0 1 0 1 0
0 1 0 2 0 3 1 3 0 0 0 1 0 0 1 1
0 1 0 2 3 0 2 2 0 0 1 0 1 1 0 0
0 1 0 3 0 2 3 1 1 0 0 0 1 0 1 0
0 1 2 0 0 3 2 3 0 1 0 0 0 0 1 1
0 1 2 0 3 0 1 3 0 0 0 0 1 1 1 0
0 1 3 0 0 2 2 2 0 1 1 0 0 0 1 0

164

**Original source visual: printed page 164, PDF page 173.**

**Original printed page 165 · source PDF page 174 · Open source page**

0130201110100010
0130203110100010
0132003201101000
0201032110100010
0201303300001110

```
0203011210011000
0203012300111000
0203101110100010
0210303210001100
0213001300111000
0230012110100010
0230103311100000
0231001211100000
0300212110001010
0301021210000011
0301023110100010
0301202310001100
0302103210011000
0310023210000011
0310201110100010
0312002110101000
0320011211100000
0320012311100000
```

0 1 3 0 2 0 1 1 1 0 1 0 0 0 1 0

0 1 3 0 2 0 3 1 1 0 1 0 0 0 1 0

0 1 3 2 0 0 3 2 0 1 1 0 1 0 0 0

0 2 0 1 0 3 2 1 1 0 1 0 0 0 1 0

0 2 0 1 3 0 3 3 0 0 0 0 1 1 1 0

0 2 0 3 0 1 1 2 1 0 0 1 1 0 0 0

0 2 0 3 0 1 2 3 0 0 1 1 1 0 0 0

0 2 0 3 1 0 1 1 1 0 1 0 0 0 1 0

0 2 1 0 3 0 3 2 1 0 0 0 1 1 0 0

0 2 1 3 0 0 1 3 0 0 1 1 1 0 0 0

0 2 3 0 0 1 2 1 1 0 1 0 0 0 1 0

0 2 3 0 1 0 3 3 1 1 1 0 0 0 0 0

0 2 3 1 0 0 1 2 1 1 1 0 0 0 0 0

0 3 0 0 2 1 2 1 1 0 0 0 1 0 1 0

0 3 0 1 0 2 1 2 1 0 0 0 0 0 1 1

0 3 0 1 0 2 3 1 1 0 1 0 0 0 1 0

0 3 0 1 2 0 2 3 1 0 0 0 1 1 0 0

0 3 0 2 1 0 3 2 1 0 0 1 1 0 0 0

0 3 1 0 0 2 3 2 1 0 0 0 0 0 1 1

0 3 1 0 2 0 1 1 1 0 1 0 0 0 1 0

0 3 1 2 0 0 2 1 1 0 1 0 1 0 0 0

0 3 2 0 0 1 1 2 1 1 1 0 0 0 0 0

0 3 2 0 0 1 2 3 1 1 1 0 0 0 0 0

**Original source visual: printed page 165, PDF page 174.**

# Appendix C PHP scripts

**Original printed page 166 · source PDF page 175 · Open source page**

## Functions.inc

```
<?
define("RANK_NUM", 10); // number of cases to be ranked
define("ENROL_COURSE_NUM", 3); // number of courses to be enrolled in
$tables = array('1' => 15, '2' => 20, '3' => 47);
// build case from given values
function build_case($dsp=0, $optic=0, $mwave=0, $ntwrk=0, $satel=0, $prfm=0, $chrt=0) {
return array(
'DSP' => $dsp,
'OPTIC' => $optic,
'MWAVE' => $mwave,
'NTWRK' => $ntwrk,
'SATEL' => $satel,
'PRFM' => $prfm,
'CHRT' => $chrt
);
}
// calculate the size of given case
function case_size($case) {
$sum = 0;
foreach ($case as $value) {
$sum += pow($value, 2);
}
```

**Original printed page 167 · source PDF page 176 · Open source page**

```
return sqrt($sum);
}
// multiply the corresponding pair and sum up the values
function sumproduct($v1, $v2) {
$sum = 0;
foreach ($v1 as $key => $value) {
$sum += $value * $v2[$key];
}
return $sum;
}
// exhaustive search through the case base
function found_new_solution(&$match,$pos,$id,$percent,
$DSP, $OPTIC, $MWAVE, $NTWRK, $SATEL, $PRFM, $CHRT,
$D1,$O2,$OM3,$M4,$NM5,$N6,$NS7,$S8) {
for ($j=RANK_NUM; $j>$pos; $j--) {
$match[$j] = $match[$j-1];
}
$match[$pos] = array('id' => $id, 'percent' => $percent,
'DSP' => $DSP, 'OPTIC' => $OPTIC,
'MWAVE' => $MWAVE, 'NTWRK' => $NTWRK,
'SATEL' => $SATEL, 'PRFM' => $PRFM,
'CHRT' => $CHRT, 'D1' => $D1,
'O2' => $O2, 'OM3' => $OM3, 'M4' => $M4,
'NM5' => $NM5, 'N6' => $N6, 'NS7' => $NS7,
'S8' => $S8);
}
// display standard table header showing attributes
function attr_tab_header() {
```

**Original printed page 168 · source PDF page 177 · Open source page**

```
print "<th>DSP</th>";
print "<th>OPTIC</th>";
print "<th>MWAVE</th>";
print "<th>NTWRK</th>";
print "<th>SATEL</th>";
print "<th>PRFM</th>";
print "<th>CHRT</th>";
}
// display standard table header showing solutions
function solu_tab_header() {
print "<th>D1</th>";
print "<th>O2</th>";
print "<th>OM3</th>";
print "<th>M4</th>";
```

```
print "<th>NM5</th>";
print "<th>N6</th>";
print "<th>NS7</th>";
print "<th>S8</th>";
}
// display INT vector
function show_interest($i) {
print "<td>" . $i['DSP'] . "</td>";
print "<td>" . $i['OPTIC'] . "</td>";
print "<td>" . $i['MWAVE'] . "</td>";
print "<td>" . $i['NTWRK'] . "</td>";
print "<td>" . $i['SATEL'] . "</td>";
print "<td>" . $i['PRFM'] . "</td>";
print "<td>" . $i['CHRT'] . "</td>";
}
```

**Original printed page 169 · source PDF page 178 · Open source page**

```
// display solutions in table
function show_solution($s, $mode = "") {
if (strlen($mode) > 0) {
printf("<td>%2.3f</td>", $s['D1']);
printf("<td>%2.3f</td>", $s['O2']);
printf("<td>%2.3f</td>", $s['OM3']);
printf("<td>%2.3f</td>", $s['M4']);
printf("<td>%2.3f</td>", $s['NM5']);
printf("<td>%2.3f</td>", $s['N6']);
printf("<td>%2.3f</td>", $s['NS7']);
printf("<td>%2.3f</td>", $s['S8']);
} else {
print "<td>" . $s['D1'] . "</td>";
print "<td>" . $s['O2'] . "</td>";
print "<td>" . $s['OM3']. "</td>";
print "<td>" . $s['M4'] . "</td>";
print "<td>" . $s['NM5'] . "</td>";
print "<td>" . $s['N6']. "</td>";
print "<td>" . $s['NS7'] . "</td>";
print "<td>" . $s['S8']. "</td>";
}
}
// initialize solution vectors
function init_solu() {
return array('D1' => 0, 'O2' => 0, 'OM3' => 0, 'M4' => 0, 'NM5' => 0, 'N6' => 0, 'NS7' =>
0, 'S8' => 0);
}
// initialize advice vector
```

**Original printed page 170 · source PDF page 179 · Open source page**

```
function init_advice() {
return array( 'COUNT' => init_solu(), 'RANK_PCENT' => init_solu(), 'WEIGHT_SUM'
=> init_solu());
}
// show ranked cases in table
function process_ranked_cases($match) {
print "<table border=\"1\"><tr>";
attr_tab_header();
print "<td>case</td>";
print "<td>percent</td>";
solu_tab_header();
print "</tr>";
$advice = init_advice();
for ($i=1; $i<=RANK_NUM; $i++) {
print "<tr>";
show_interest($match[$i]);
print "<td>" . $match[$i]['id'] . "</td>";
printf("<td>%2.3f</td>", $match[$i]['percent']*100);
show_solution($match[$i]);
print "</tr>";
foreach ($advice['RANK_PCENT'] as $course => $rank_percentage) {
// if this course is recommended, ie solution vector shows 1
if ($match[$i][$course] > 0) {
$advice['COUNT'][$course]++;
$advice['WEIGHT_SUM'][$course] += (2/75 + (RANK_NUM-
$i+1)/75) * 100;
if ($match[$i]['percent'] > $rank_percentage) {
```

**Original printed page 171 · source PDF page 180 · Open source page**

```
$advice['RANK_PCENT'][$course] = $match[$i]['percent']
```

```
* 100;
}
}
}
}
print "</table><p>";
return $advice;
}
// show advice in table format
function show_advice_table($advice, $prerequisites, $difficulties) {
print "<table border=\"1\">";
print "<tr>";
print "<td>Description</td>";
solu_tab_header();
print "</tr>";
print "<tr>";
print "<td>Prequisite</td>";
show_solution($prerequisites);
print "</tr>";
print "<tr>";
print "<td>Difficulty</td>";
show_solution($difficulties);
print "</tr>";
print "<tr>";
print "<td>Count</td>";
show_solution($advice['COUNT']);
print "</tr>";
print "<tr>";
```

**Original printed page 172 · source PDF page 181 · Open source page**

```
print "<td>Ranked Percentage</td>";
show_solution($advice['RANK_PCENT'], "f");
print "</tr>";
print "<tr>";
print "<td>Weighted Sum according to Counts</td>";
show_solution($advice['WEIGHT_SUM'], "f");
print "</tr>";
print "</table><p>";
}
// give out advice in plain English
function give_advice($advice, $prerequisites, $difficulties, $performance) {
$ordinal = 0;
foreach ($advice['RANK_PCENT'] as $course => $rank_percentage) {
$relevance = $advice['WEIGHT_SUM'][$course];
if ($prerequisites[$course] == 0) {
$colour = "red";
$reason = "<br>This course is not available because its pre-requisites have
not been fulfilled.";
} elseif ($difficulties[$course] > $performance) {
$colour = "orange";
$reason = "<br>This course is not recommended because its perceived
difficulty is greater than the student's demonstrated academic performance.";
} elseif ($ordinal < ENROL_COURSE_NUM) {
$colour = "green";
$ordinal++;
$reason = "<br>This is recommended as the choice # $ordinal.";
} else {
$colour = "blue";
$reason = "";
```

**Original printed page 173 · source PDF page 182 · Open source page**

```
}
if ($rank_percentage > 0) {
printf("<p><font color=\"$colour\">The course $course has a rating of
%2.3f%% (matched with cases in the past) and a subrating of %2.3f%% (based on frequency of
occurrances in ranked cases). %s</font>",
$rank_percentage, $relevance, $reason);
}
}
}
?>
```

## Index.php

```
<html>
<head><title>CBR</title></head>
<body>
<h3>Number of Different INT permutations out of 60 (5P3)
```

```
</h3>
<?
include("prepend.inc");
foreach ($tables as $key => $value) {
print "<a href=\"knowledge.php?table=$key\" target=\"_blank\">knowledge
$value</a><br>";
}
print "<h1>Problem</h1>\n";
print "<form method=\"GET\" action=\"match.php\">";
print "<table border=\"1\">\n";
print "<tr><td>id</td>\n";
attr_tab_header();
```

**Original printed page 174 · source PDF page 183 · Open source page**

```
print "<td>knowledge15</td><td>knowledge20</td><td>knowledge47</td>\n";
print "</tr>\n";
print "<tr><td><input type=\"Submit\" value=\"Submit\"></td>";
print "<td><input type=\"text\" name=\"DSP\" size=\"1\"></td>";
print "<td><input type=\"text\" name=\"OPTIC\" size=\"1\"></td>";
print "<td><input type=\"text\" name=\"MWAVE\" size=\"1\"></td>";
print "<td><input type=\"text\" name=\"NTWRK\" size=\"1\"></td>";
print "<td><input type=\"text\" name=\"SATEL\" size=\"1\"></td>";
print "<td><input type=\"text\" name=\"PRFM\" size=\"1\"></td>";
print "<td><input type=\"text\" name=\"CHRT\" size=\"1\"></td>";
print "<td><input type=\"radio\" name=\"table\" value=\"1\"></td>";
print "<td><input type=\"radio\" name=\"table\" value=\"2\"></td>";
print "<td><input type=\"radio\" name=\"table\" value=\"3\"></td>";
print "</tr>";
// display a list of predefined test cases in table
$lines = file ('http://www.tanat.net/cbr/problems.txt');
foreach ($lines as $line_num => $line) {
$problem = build_case();
list($problem['DSP'],
$problem['OPTIC'],$problem['MWAVE'],$problem['NTWRK'],$problem['SATEL'],$problem['P
RFM'],$problem['CHRT'])=explode("\t",$line);
print "<tr>\n";
$query_str =
"DSP=".$problem['DSP']."&OPTIC=".$problem['OPTIC']."&MWAVE=".$problem['MWAVE'].
"&NTWRK=".$problem['NTWRK']."&SATEL=".$problem['SATEL']."&PRFM=".$problem['P
RFM']."&CHRT=".$problem['CHRT'];
$problem_id = strval($line_num+1);
```

**Original printed page 175 · source PDF page 184 · Open source page**

```
?>
<td><?=$problem_id?></td>
<?
show_interest($problem);
foreach ($tables as $key => $value) {
print "<td><a href=\"match.php?table=$key&$query_str\"
target=\"_blank\">$problem_id - $value</a></td>";
}
print "</tr>\n";
}
print "</table>\n";
print "</form>\n";
include("append.inc");
?>
</body>
</html>
```

**Original printed page 176 · source PDF page 185 · Open source page**

## Knowledge.php

```
<?
include("prepend.inc");
$knowledge_size = $tables[$_GET['table']];
print "<h2>Knowledge $knowledge_size</h2>";
$query = "SELECT * FROM knowledge$knowledge_size";
$result = mysql_query($query) or die ("Query failed");
print "<table border=\"1\">\n";
print "<tr><td>id</td>\n";
attr_tab_header();
solu_tab_header();
print "<td>case size</td></tr>\n";
while ($row = mysql_fetch_array($result, MYSQL_ASSOC)) {
print "<tr>\n";
print "<td>" . $row['id'] . "</td>\n";
```

```
show_interest($row);
show_solution($row);
$case = build_case($row['DSP'], $row['OPTIC'], $row['MWAVE'], $row['NTWRK'],
$row['SATEL'], $row['PRFM'], $row['CHRT']);
printf ("<td>%2.3f</td>\n",case_size($case));
print "</tr>\n";
}
print "</table><p>\n";
include("append.inc");
?>
```

**Original printed page 177 · source PDF page 186 · Open source page**

## Match.php

```
<?
include("prepend.inc");
$knowledge_size = $tables[$_GET['table']];
print "<h2>Problem (using Knowledge $knowledge_size)";
print "</h2>";
print "<table border=\"1\">";
attr_tab_header();
print "<td>problem size</td>";
print "</tr>";
// create a problem vector
$problem =
build_case($_GET['DSP'],$_GET['OPTIC'],$_GET['MWAVE'],$_GET['NTWRK'],$_GET['SA
TEL'],$_GET['PRFM'],$_GET['CHRT']);
print "<tr>";
show_interest($problem);
printf ("<td>%2.3f</td>",case_size($problem));
print "</tr>";
print "</table><p>";
/***************************************Correlaton & Diff of Means
*****************************************/
for ($i=1; $i<=RANK_NUM; $i++) {
$match[$i] =
array('id'=>"0",'percent'=>0,'DSP'=>"0",'OPTIC'=>"0",'MWAVE'=>"0",'NTWRK'=>"0",'SATE
L'=>"0",'PRFM'=>"0",'CHRT'=>"0",'D1'=>"0",'O2'=>"0",'OM3'=>"0",'M4'=>"0",'NM5'=>"0",'
N6'=>"0",'NS7'=>"0",'S8'=>"0");
```

**Original printed page 178 · source PDF page 187 · Open source page**

```
}
$query = "SELECT * FROM knowledge$knowledge_size";
$result = mysql_query($query) or die ("Query failed");
while ($row = mysql_fetch_array($result, MYSQL_ASSOC)) {
$case = build_case($row['DSP'], $row['OPTIC'], $row['MWAVE'],
$row['NTWRK'], $row['SATEL'], $row['PRFM'], $row['CHRT']);
$case_int = array_slice($case, 0, 5);
$case_char = array_slice($case, 5);
$problem_int = array_slice($problem, 0, 5);
$problem_char = array_slice($problem, 5);
$mean = $problem_char;
foreach ($case_char as $key => $value) {
$mean[$key] = ($value + $problem_char[$key]) / 2;
}
$diff = $problem_char;
foreach ($problem_char as $key => $value) {
$diff[$key] = $value - $mean[$key];
}
// calculate matching percentage, using vector product divided by the product of
both case sizes. Assign a weight of 5 out of 7
$match_percent = sumproduct($problem_int, $case_int) /
case_size($problem_int) / case_size($case_int) * 5/7;
$match_percent += (1- pow(case_size($diff) / case_size($problem_char), 2)) *
2/7;
for ($j=1; $j<=RANK_NUM; $j++) {
if ($match_percent > $match[$j]['percent']) {
```

**Original printed page 179 · source PDF page 188 · Open source page**

```
found_new_solution($match, $j, $row['id'], $match_percent,
$row['DSP'], $row['OPTIC'], $row['MWAVE'],
$row['NTWRK'], $row['SATEL'],
$row['PRFM'], $row['CHRT'], $row['D1'], $row['O2'],
$row['OM3'], $row['M4'], $row['NM5'],
$row['N6'], $row['NS7'], $row['S8']);
break;
}
```

```
}
}
$difficulty = array('D1' => 1, 'O2' => 2, 'OM3' => 1, 'M4' => 3, 'NM5' => 2, 'N6' => 3,
'NS7' => 1,'S8'=>2);
$prereq = array('D1' => 1, 'O2' => 1, 'OM3' => 0, 'M4' => 1, 'NM5' => 0, 'N6' => 1, 'NS7'
=> 0, 'S8' => 1);
print "<h2>Correlation of Interests (INT) and Difference of Means</h2><br>";
$advice1 = process_ranked_cases($match);
/***************************************distance of 2 vectors
*****************************************/
for ($i=1; $i<=RANK_NUM; $i++) {
$match[$i] =
array('id'=>"0",'percent'=>0,'DSP'=>"0",'OPTIC'=>"0",'MWAVE'=>"0",'NTWRK'=>"0",'SATE
L'=>"0",'PRFM'=>"0",'CHRT'=>"0",'D1'=>"0",'O2'=>"0",'OM3'=>"0",'M4'=>"0",'NM5'=>"0",'
N6'=>"0",'NS7'=>"0",'S8'=>"0");
}
$query = "SELECT * FROM knowledge$knowledge_size";
$result = mysql_query($query) or die ("Query failed");
while ($row = mysql_fetch_array($result, MYSQL_ASSOC)) {
```

**Original printed page 180 · source PDF page 189 · Open source page**

```
$case = build_case($row['DSP'], $row['OPTIC'], $row['MWAVE'],
$row['NTWRK'], $row['SATEL'], $row['PRFM'], $row['CHRT']);
$diff = build_case();
foreach ($case as $key => $value) {
$diff[$key] = $value - $problem[$key];
}
// calculate matching percentage, the further the distance from the knowledge
vector, the lower the matching percentage would be.
$match_percent = 1- pow(case_size($diff) / case_size($problem), 2);
for ($j=1; $j<=RANK_NUM; $j++) {
if ($match_percent > $match[$j]['percent']) {
found_new_solution($match, $j, $row['id'], $match_percent,
$row['DSP'], $row['OPTIC'], $row['MWAVE'],
$row['NTWRK'], $row['SATEL'],
$row['PRFM'], $row['CHRT'], $row['D1'], $row['O2'],
$row['OM3'], $row['M4'], $row['NM5'],
$row['N6'], $row['NS7'], $row['S8']);
break;
}
}
}
print "<h2>Distance of 2 vectors</h2><br>";
$advice2 = process_ranked_cases($match);
/***************************************trivial IR
*****************************************/
for ($i=1; $i<=RANK_NUM; $i++) {
```

**Original printed page 181 · source PDF page 190 · Open source page**

```
$match[$i] =
array('id'=>"0",'percent'=>0,'DSP'=>"0",'OPTIC'=>"0",'MWAVE'=>"0",'NTWRK'=>"0",'SATE
L'=>"0",'PRFM'=>"0",'CHRT'=>"0",'D1'=>"0",'O2'=>"0",'OM3'=>"0",'M4'=>"0",'NM5'=>"0",'
N6'=>"0",'NS7'=>"0",'S8'=>"0");
}
$query = "SELECT * FROM knowledge$knowledge_size";
$result = mysql_query($query) or die ("Query failed");
while ($row = mysql_fetch_array($result, MYSQL_ASSOC)) {
$case = build_case($row['DSP'], $row['OPTIC'], $row['MWAVE'],
$row['NTWRK'], $row['SATEL'], $row['PRFM'], $row['CHRT']);
// matching percentage is calculated purely using trivial IR algorithm
$match_percent = sumproduct($problem,$case) / case_size($problem) /
case_size($case);
for ($j=1; $j<=RANK_NUM; $j++) {
if ($match_percent > $match[$j]['percent']) {
found_new_solution($match, $j, $row['id'], $match_percent,
$row['DSP'], $row['OPTIC'], $row['MWAVE'],
$row['NTWRK'], $row['SATEL'],
$row['PRFM'], $row['CHRT'], $row['D1'], $row['O2'],
$row['OM3'], $row['M4'], $row['NM5'],
$row['N6'], $row['NS7'], $row['S8']);
break;
}
}
}
print "<h2>Trivial IR algorithm</h2><br>";
$advice3 = process_ranked_cases($match);
/******************************************************/
print "<h1>Problem</h1>";
```

**Original printed page 182 · source PDF page 191 · Open source page**

```
print "<table border=\"1\">";
attr_tab_header();
print "</tr>";
print "<tr>";
show_interest($problem);
print "</tr>";
print "</table><p>";
print "<h1>Advice Table</h1>";
print "<h3>Correlation of Interests (INT) and Difference of Means</h3><br>";
show_advice_table($advice1, $prereq, $difficulty);
print "<h3>Distance of 2 vectors</h3><br>";
show_advice_table($advice2, $prereq, $difficulty);
print "<h3>Trivial IR algorithm</h3><br>";
show_advice_table($advice3, $prereq, $difficulty);
array_multisort( $advice1['RANK_PCENT'], SORT_NUMERIC,
SORT_DESC,$advice3['WEIGHT_SUM'],SORT_NUMERIC, SORT_DESC);
array_multisort( $advice2['RANK_PCENT'], SORT_NUMERIC,
SORT_DESC,$advice3['WEIGHT_SUM'],SORT_NUMERIC, SORT_DESC);
array_multisort( $advice3['RANK_PCENT'], SORT_NUMERIC,
SORT_DESC,$advice3['WEIGHT_SUM'],SORT_NUMERIC, SORT_DESC);
/******************************************************/
print "<h1>Problem</h1>";
print "<table border=\"1\"><tr>";
attr_tab_header();
print "</tr>";
print "<tr>";
show_interest($problem);
```

**Original printed page 183 · source PDF page 192 · Open source page**

```
print "</tr>";
print "</table><p>";
print "<h1>Advice with Reasons</h1>";
print "<p><font color=\"red\">RED</font> denotes courses that are blocked due to its
unfulfilled pre-requisites<br>";
print "<font color=\"orange\">ORANGE</font> denotes courses that are not recommended due
to its perceived difficulty being greater than the student's demonstrated academic
performance<br>";
print "<font color=\"green\">GREEN</font> denotes the recommended course with a ranked
order<br>";
print "<font color=\"blue\">BLUE</font> denotes other possible courses student may consider
taking</p>";
print "<h3>Correlation of Interests (INT) and Difference of Means</h3><br>";
give_advice($advice1, $prereq, $difficulty, $problem['PRFM']);
print "<h3>Distance of 2 vectors</h3><br>";
give_advice($advice2, $prereq, $difficulty, $problem['PRFM']);
print "<h3>Trivial IR algorithm</h3><br>";
give_advice($advice3, $prereq, $difficulty, $problem['PRFM']);
include("append.inc");
?>
```

# Appendix D Algorithm Comparison and Validation

**Original printed page 184 · source PDF page 193 · Open source page**

**Table D-1: comparison between algorithms when performance is lower than difficulty**

| Test case | Expected Result | Trivial IR | Euclidean Distance | Improvement |
|---|---|---|---|---|
| 0, 2, 0, 3, 1, 2, 3 | NS7, NM5, O2 | NS7, OM3, O2 | O2, OM3, NS7 | NS7, OM3, O2 |
| 0, 3, 0, 1, 2, 1, 1 | OM3, NS7, D1 | OM3, NS7, D1 | OM3, NS7, D1 | OM3, NS7, D1 |
| 3, 2, 0, 0, 1, 2, 2 | D1, O2, OM3 | D1, O2, OM3 | D1, O2, OM3 | D1, O2, OM3 |
| 0, 3, 1, 2, 0, 1, 3 | OM3, NS7, D1 | OM3, NS7, D1 | OM3, NS7, D1 | OM3, NS7, D1 |
| 0, 1, 0, 3, 2, 1, 3 | NS7, OM3, D1 | NS7, OM3 (2) | NS7, OM3 (2) | NS7, OM3, D1 |
| 0, 1, 3, 0, 2, 2, 3 | OM3, NM5, S8 | OM3, NM5, NS7 | OM3, NM5, NS7 | OM3, NM5, NS7 |
| 0, 0, 1, 2, 3, 2, 1 | NS7, NM5, OM3 | NM5, S8, NS7 | NM5, NS7, S8 | NS7, NM5, S8 |
| 1, 2, 0, 0, 3, 1, 2 | NS7, OM3, D1 | NS7, OM3, D1 | NS7, OM3, D1 | NS7, OM3, D1 |

**Table D-2: comparison between algorithms when prerequisites are not yet fulfilled**

| Test case | Expected Result | Trivial IR | Euclidean | Improvement |
|---|---|---|---|---|
| 0, 2, 0, 3, 1, 3, 3 | N6, NS7, NM5 | N6, OM3, NS7 | OM3, N6, NS7 | NS7, OM3, N6 |
| 0, 3, 0, 1, 2, 3, 1 | OM3, NS7, NM5 | N6, OM3, NS7 | OM3, N6, NS7 | NS7, OM3, NM5 |
| 3, 2, 0, 0, 1, 3, 2 | D1, OM3, NS7 | D1, OM3, NS7 | D1, OM3, NS7 | D1, OM3, NS7 |
| 0, 3, 1, 2, 0, 3, 3 | OM3, N6, NM5 | M4, OM3, D1 | M4, OM3, D1 | OM3, M4, NS7 |
| 0, 1, 0, 3, 2, 3, 3 | N6, S8, O2 | N6, OM3, NS7 | OM3, N6, NS7 | N6, S8, O2 |
| 0, 1, 3, 0, 2, 3, 3 | M4, S8, O2 | OM3, M4, NS7 | OM3, M4, NM5 | M4, S8, D1 |
| 0, 0, 1, 2, 3, 3, 1 | S8, N6, M4 | NM5, NS7, D1 | NM5, NS7, D1 | S8, N6, D1 |
| 1, 2, 0, 0, 3, 3, 2 | S8, O2, D1 | NS7, OM3, NM5 | NS7, OM3, NM5 | S8, O2, D1 |

# Appendix E Example Screen Output

**Original printed page 185 · source PDF page 194 · Open source page**

**Problem (using Knowledge 47)**

| DSP | OPTIC | MWAVE | NTWRK | SATEL | PRFM | CHRT | problem size |
|---|---|---|---|---|---|---|---|
| 0 | 2 | 0 | 3 | 1 | 3 | 3 | 5.657 |

**Correlation of Interests (INT) and Difference of Means**

| DSP | OPTIC | MWAVE | NTWRK | SATEL | PRFM | CHRT | case | percent | D1 | O2 | OM3 | M4 | NM5 | N6 | NS7 | |
|---|---|---|---|---|---|---|---|---|---|---|---|---|---|---|---|---|
| 0 | 2 | 0 | 3 | 1 | 2 | 2 | 14 | 99.206 | 0 | 0 | 1 | 0 | 0 | 1 | 1 | |
| 0 | 2 | 0 | 3 | 1 | 1 | 3 | 13 | 98.413 | 0 | 1 | 0 | 0 | 0 | 1 | 1 | |
| 0 | 2 | 1 | 3 | 0 | 3 | 1 | 16 | 93.311 | 0 | 0 | 1 | 0 | 1 | 0 | 1 | |
| 1 | 2 | 0 | 3 | 0 | 1 | 3 | 26 | 93.311 | 0 | 0 | 0 | 0 | 1 | 1 | 1 | |
| 0 | 1 | 0 | 3 | 2 | 2 | 1 | 8 | 92.914 | 0 | 0 | 1 | 0 | 1 | 0 | 1 | |
| 0 | 3 | 0 | 2 | 1 | 1 | 2 | 18 | 92.914 | 0 | 1 | 1 | 0 | 0 | 0 | 1 | |
| 1 | 3 | 0 | 2 | 0 | 3 | 1 | 29 | 88.209 | 1 | 0 | 1 | 0 | 0 | 0 | 1 | |
| 1 | 3 | 0 | 2 | 0 | 1 | 1 | 28 | 86.621 | 1 | 0 | 1 | 0 | 0 | 0 | 1 | |
| 2 | 1 | 0 | 3 | 0 | 3 | 2 | 36 | 84.297 | 1 | 0 | 0 | 0 | 1 | 1 | 0 | |
| 0 | 1 | 0 | 2 | 3 | 1 | 3 | 7 | 83.107 | 0 | 0 | 0 | 0 | 0 | 1 | 1 | 1 |

**Distance of 2 vectors**

| DSP | OPTIC | MWAVE | NTWRK | SATEL | PRFM | CHRT | case | percent | D1 | O2 | OM3 | M4 | NM5 | N6 | NS7 | |
|---|---|---|---|---|---|---|---|---|---|---|---|---|---|---|---|---|
| 0 | 2 | 0 | 3 | 1 | 2 | 2 | 14 | 93.750 | 0 | 0 | 1 | 0 | 0 | 1 | 1 | 0 |

| DSP | OPTIC | MWAVE | NTWRK | SATEL | PRFM | CHRT | case | percent | D1 | O2 | OM3 | M4 | NM5 | N6 | NS7 | |
|---|---|---|---|---|---|---|---|---|---|---|---|---|---|---|---|---|
| 0 | 2 | 0 | 3 | 1 | 1 | 3 | 13 | 87.500 | 0 | 1 | 0 | 0 | 0 | 1 | 1 | 0 |
| 0 | 2 | 1 | 3 | 0 | 3 | 1 | 16 | 81.250 | 0 | 0 | 1 | 0 | 1 | 0 | 1 | 0 |
| 1 | 2 | 0 | 3 | 0 | 1 | 3 | 26 | 81.250 | 0 | 0 | 0 | 0 | 1 | 1 | 1 | 0 |

## *Appendix E: Example Screen Output*

Problem (using Knowledge 47)

| DSP | OPTIC | MWAVE | NTWRK | SATEL | PRFM | CHRT | problem size |
|---|---|---|---|---|---|---|---|
| 0 | 2 | 0 | 3 | 1 | 3 | 3 | 5.657 |

Correlation of Interests (INT) and Difference of Means

| DSP | OPTIC | MWAVE | NTWRK | SATEL | PRFM | CHRT | case | percent | D1 | O2 | OM3 | M4 | NM5 | N6 | NS7 | [illegible] |
|---|---|---|---|---|---|---|---|---|---|---|---|---|---|---|---|---|
| 0 | 2 | 0 | 3 | 1 | 2 | 2 | 14 | 99.206 | 0 | 0 | 1 | 0 | 0 | 1 | 1 | [illegible] |
| 0 | 2 | 0 | 3 | 1 | 1 | 3 | 13 | 98.413 | 0 | 1 | 0 | 0 | 0 | 1 | 1 | [illegible] |
| 0 | 2 | 1 | 3 | 0 | 3 | 1 | 16 | 93.311 | 0 | 0 | 1 | 0 | 1 | 0 | 1 | [illegible] |
| 1 | 2 | 0 | 3 | 0 | 1 | 3 | 26 | 93.311 | 0 | 0 | 0 | 0 | 1 | 1 | 1 | [illegible] |
| 0 | 1 | 0 | 3 | 2 | 2 | 1 | 8 | 92.914 | 0 | 0 | 1 | 0 | 1 | 0 | 1 | [illegible] |
| 0 | 3 | 0 | 2 | 1 | 1 | 2 | 18 | 92.914 | 0 | 1 | 1 | 0 | 0 | 0 | 1 | [illegible] |
| 1 | 3 | 0 | 2 | 0 | 3 | 1 | 29 | 88.209 | 1 | 0 | 1 | 0 | 0 | 0 | 1 | [illegible] |
| 1 | 3 | 0 | 2 | 0 | 1 | 1 | 28 | 86.621 | 1 | 0 | 1 | 0 | 0 | 0 | 1 | [illegible] |
| 2 | 1 | 0 | 3 | 0 | 3 | 2 | 36 | 84.297 | 1 | 0 | 0 | 0 | 1 | 1 | 0 | [illegible] |
| 0 | 1 | 0 | 2 | 3 | 1 | 3 | 7 | 83.107 | 0 | 0 | 0 | 0 | 0 | 1 | 1 | 1 |

Distance of 2 vectors

| DSP | OPTIC | MWAVE | NTWRK | SATEL | PRFM | CHRT | case | percent | D1 | O2 | OM3 | M4 | NM5 | N6 | NS7 | S |
|---|---|---|---|---|---|---|---|---|---|---|---|---|---|---|---|---|
| 0 | 2 | 0 | 3 | 1 | 2 | 2 | 14 | 93.750 | 0 | 0 | 1 | 0 | 0 | 1 | 1 | 0 |
| 0 | 2 | 0 | 3 | 1 | 1 | 3 | 13 | 87.500 | 0 | 1 | 0 | 0 | 0 | 1 | 1 | 0 |
| 0 | 2 | 1 | 3 | 0 | 3 | 1 | 16 | 81.250 | 0 | 0 | 1 | 0 | 1 | 0 | 1 | 0 |
| 1 | 2 | 0 | 3 | 0 | 1 | 3 | 26 | 81.250 | 0 | 0 | 0 | 0 | 1 | 1 | 1 | 0 |

**Original source visual: printed page 185, PDF page 194.**

**Distance of 2 vectors (continued from PDF194)**

| DSP | OPTIC | MWAVE | NTWRK | SATEL | PRFM | CHRT | case | percent | D1 | O2 | OM3 | M4 | NM5 | N6 | NS7 | |
|---|---|---|---|---|---|---|---|---|---|---|---|---|---|---|---|---|
| 0 | 1 | 0 | 3 | 2 | 2 | 1 | 8 | 78.125 | 0 | 0 | 1 | 0 | 1 | 0 | 1 | |
| 0 | 3 | 0 | 2 | 1 | 1 | 2 | 18 | 78.125 | 0 | 1 | 1 | 0 | 0 | 0 | 1 | |
| 2 | 1 | 0 | 3 | 0 | 3 | 2 | 36 | 78.125 | 1 | 0 | 0 | 0 | 1 | 1 | 0 | |
| 0 | 2 | 0 | 1 | 3 | 3 | 3 | 12 | 75.000 | 0 | 1 | 0 | 0 | 0 | 0 | 1 | |
| 1 | 3 | 0 | 2 | 0 | 3 | 1 | 29 | 75.000 | 1 | 0 | 1 | 0 | 0 | 0 | 1 | |
| 0 | 1 | 0 | 2 | 3 | 1 | 3 | 7 | 68.750 | 0 | 0 | 0 | 0 | 0 | 1 | 1 | |

**Trivial IR algorithm**

| DSP | OPTIC | MWAVE | NTWRK | SATEL | PRFM | CHRT | case | percent | D1 | O2 | OM3 | M4 | NM5 | N6 | NS7 | |
|---|---|---|---|---|---|---|---|---|---|---|---|---|---|---|---|---|
| 0 | 2 | 0 | 3 | 1 | 2 | 2 | 14 | 97.991 | 0 | 0 | 1 | 0 | 0 | 1 | 1 | |
| 0 | 2 | 0 | 3 | 1 | 1 | 3 | 13 | 93.819 | 0 | 1 | 0 | 0 | 0 | 1 | 1 | |
| 0 | 2 | 1 | 3 | 0 | 3 | 1 | 16 | 90.211 | 0 | 0 | 1 | 0 | 1 | 0 | 1 | |
| 1 | 2 | 0 | 3 | 0 | 1 | 3 | 26 | 90.211 | 0 | 0 | 0 | 0 | 1 | 1 | 1 | |
| 0 | 1 | 0 | 3 | 2 | 2 | 1 | 8 | 89.222 | 0 | 0 | 1 | 0 | 1 | 0 | 1 | |
| 0 | 3 | 0 | 2 | 1 | 1 | 2 | 18 | 89.222 | 0 | 1 | 1 | 0 | 0 | 0 | 1 | |
| 2 | 1 | 0 | 3 | 0 | 3 | 2 | 36 | 88.454 | 1 | 0 | 0 | 0 | 1 | 1 | 0 | |
| 0 | 2 | 0 | 1 | 3 | 3 | 3 | 12 | 87.500 | 0 | 1 | 0 | 0 | 0 | 0 | 1 | |
| 1 | 3 | 0 | 2 | 0 | 3 | 1 | 29 | 86.603 | 1 | 0 | 1 | 0 | 0 | 0 | 1 | |
| 2 | 0 | 1 | 3 | 0 | 3 | 3 | 32 | 84.375 | 0 | 0 | 0 | 0 | 1 | 1 | 1 | |

**Problem**

| DSP | OPTIC | MWAVE | NTWRK | SATEL | PRFM | CHRT |
| --- | --- | --- | --- | --- | --- | --- |
| 0 | 2 | 0 | 3 | 1 | 3 | 3 |

## Advice Table

## Correlation of Interests (INT) and Difference of Means

Description D1 O2 OM3 M4 NM5 N6 NS7 S8

| | | | | | | | | | | | | | | | |
|---|---|---|---|---|---|---|---|---|---|---|---|---|---|---|---|
| 0 | 1 | 0 | 3 | 2 | 2 | 1 | 8 | 78.125 | 0 | 0 | 1 | 0 | 1 | 0 | 1 |
| 0 | 3 | 0 | 2 | 1 | 1 | 2 | 18 | 78.125 | 0 | 1 | 1 | 0 | 0 | 0 | 1 |
| 2 | 1 | 0 | 3 | 0 | 3 | 2 | 36 | 78.125 | 1 | 0 | 0 | 0 | 1 | 1 | 0 |
| 0 | 2 | 0 | 1 | 3 | 3 | 3 | 12 | 75.000 | 0 | 1 | 0 | 0 | 0 | 0 | 1 |
| 1 | 3 | 0 | 2 | 0 | 3 | 1 | 29 | 75.000 | 1 | 0 | 1 | 0 | 0 | 0 | 1 |
| 0 | 1 | 0 | 2 | 3 | 1 | 3 | 7 | 68.750 | 0 | 0 | 0 | 0 | 0 | 1 | 1 |

Trivial IR algorithm

| DSP | OPTIC | MWAVE | NTWRK | SATEL | PRFM | CHRT | case | percent | D1 | O2 | OM3 | M4 | NM5 | N6 | NS7 |
|---|---|---|---|---|---|---|---|---|---|---|---|---|---|---|---|
| 0 | 2 | 0 | 3 | 1 | 2 | 2 | 14 | 97.991 | 0 | 0 | 1 | 0 | 0 | 1 | 1 |
| 0 | 2 | 0 | 3 | 1 | 1 | 3 | 13 | 93.819 | 0 | 1 | 0 | 0 | 0 | 1 | 1 |
| 0 | 2 | 1 | 3 | 0 | 3 | 1 | 16 | 90.211 | 0 | 0 | 1 | 0 | 1 | 0 | 1 |
| 1 | 2 | 0 | 3 | 0 | 1 | 3 | 26 | 90.211 | 0 | 0 | 0 | 0 | 1 | 1 | 1 |
| 0 | 1 | 0 | 3 | 2 | 2 | 1 | 8 | 89.222 | 0 | 0 | 1 | 0 | 1 | 0 | 1 |
| 0 | 3 | 0 | 2 | 1 | 1 | 2 | 18 | 89.222 | 0 | 1 | 1 | 0 | 0 | 0 | 1 |
| 2 | 1 | 0 | 3 | 0 | 3 | 2 | 36 | 88.454 | 1 | 0 | 0 | 0 | 1 | 1 | 0 |
| 0 | 2 | 0 | 1 | 3 | 3 | 3 | 12 | 87.500 | 0 | 1 | 0 | 0 | 0 | 0 | 1 |
| 1 | 3 | 0 | 2 | 0 | 3 | 1 | 29 | 86.603 | 1 | 0 | 1 | 0 | 0 | 0 | 1 |
| 2 | 0 | 1 | 3 | 0 | 3 | 3 | 32 | 84.375 | 0 | 0 | 0 | 0 | 1 | 1 | 1 |

Problem

| DSP | OPTIC | MWAVE | NTWRK | SATEL | PRFM | CHRT |
|---|---|---|---|---|---|---|
| 0 | 2 | 0 | 3 | 1 | 3 | 3 |

Advice Table

Correlation of Interests (INT) and Difference of Means

| Description | D1 | O2 | OM3 | M4 | NM5 | N6 | NS7 | S8 |
|---|---|---|---|---|---|---|---|---|

Original source visual: printed page 186, PDF page 195.

**Original printed page 187 · source PDF page 196 · Open source page**

**Correlation of Interests (INT) and Difference of Means — source screen advice table**

| Description | D1 | O2 | OM3 | M4 | NM5 | N6 | NS7 | S8 |
|---|---|---|---|---|---|---|---|---|
| Prequisite | 1 | 1 | 0 | 1 | 0 | 1 | 0 | 1 |
| Difficulty | 1 | 2 | 1 | 3 | 2 | 3 | 1 | 2 |
| Count | 3 | 2 | 6 | 0 | 4 | 5 | 9 | 1 |
| Ranked Percentage | 88.209 | 98.413 | 99.206 | 0.000 | 93.311 | 99.206 | 99.206 | 83.107 |
| Weighted Sum according to Counts | 20.000 | 24.000 | 64.000 | 0.000 | 41.333 | 52.000 | 94.667 | 4.000 |

**Distance of 2 vectors — source screen advice table**

| Description | D1 | O2 | OM3 | M4 | NM5 | N6 | NS7 | S8 |
|---|---|---|---|---|---|---|---|---|
| Prequisite | 1 | 1 | 0 | 1 | 0 | 1 | 0 | 1 |
| Difficulty | 1 | 2 | 1 | 3 | 2 | 3 | 1 | 2 |
| Count | 2 | 3 | 5 | 0 | 4 | 5 | 9 | 2 |
| Ranked Percentage | 78.125 | 87.500 | 93.750 | 0.000 | 81.250 | 93.750 | 93.750 | 75.000 |
| Weighted Sum according to Counts | 13.333 | 30.667 | 54.667 | 0.000 | 44.000 | 54.667 | 92.000 | 10.667 |

**Trivial IR algorithm — source screen advice table**

| Description | D1 | O2 | OM3 | M4 | NM5 | N6 | NS7 | S8 |
|---|---|---|---|---|---|---|---|---|
| Prequisite | 1 | 1 | 0 | 1 | 0 | 1 | 0 | 1 |
| Difficulty | 1 | 2 | 1 | 3 | 2 | 3 | 1 | 2 |
| Count | 2 | 3 | 5 | 0 | 5 | 5 | 9 | 1 |
| Ranked Percentage | 88.454 | 93.819 | 97.991 | 0.000 | 90.211 | 97.991 | 97.991 | 87.500 |
| Weighted Sum according to Counts | 13.333 | 30.667 | 54.667 | 0.000 | 48.000 | 54.667 | 92.000 | 6.667 |

**Problem**

| DSP | OPTIC | MWAVE | NTWRK | SATEL | PRFM | CHRT |
|---|---|---|---|---|---|---|
| 0 | 2 | 0 | 3 | 1 | 3 | 3 |

## Advice with Reasons

**Original printed page 188 · source PDF page 197 · Open source page**

RED denotes courses that are blocked due to its unfulfilled pre-requisites denotes courses that are not recommended due to its perceived difficulty being greater than the student's demonstrated academic performance

GREEN denotes the recommended course with a ranked order

BLUE denotes other possible courses student may consider taking

### Correlation of Interests (INT) and Difference of Means

The course NS7 has a rating of 99.206% (matched with cases in the past) and a subrating of 94.667% (based on frequency of occurrances in ranked cases). This course is not available because its pre-requisites have not been fulfilled.

The course OM3 has a rating of 99.206% (matched with cases in the past) and a subrating of 64.000% (based on frequency of occurrances in ranked cases). This course is not available because its pre-requisites have not been fulfilled.

The course N6 has a rating of 99.206% (matched with cases in the past) and a subrating of 52.000% (based on frequency of occurrances in ranked cases). This is recommended as the choice # 1.

The course O2 has a rating of 98.413% (matched with cases in the past) and a subrating of 24.000% (based on frequency of occurrances in ranked cases). This is recommended as the choice # 2.

The course NM5 has a rating of 93.311% (matched with cases in the past) and a subrating of 41.333% (based on frequency of occurrances in ranked cases). This course is not available because its pre-requisites have not been fulfilled.

The course D1 has a rating of 88.209% (matched with cases in the past) and a subrating of 20.000% (based on frequency of occurrances in ranked cases). This is recommended as the choice # 3.

The course S8 has a rating of 83.107% (matched with cases in the past) and a subrating of 4.000% (based on frequency of occurrances in ranked cases).

### Distance of 2 vectors

The course OM3 has a rating of 93.750% (matched with cases in the past) and a subrating of 54.667% (based on frequency of occurrances in ranked cases). This course is not available because its pre-requisites have not been fulfilled.

**Original printed page 189 · source PDF page 198 · Open source page**

The course N6 has a rating of 93.750% (matched with cases in the past) and a subrating of 54.667% (based on frequency of occurrances in ranked cases). This is recommended as the choice # 1.

The course NS7 has a rating of 93.750% (matched with cases in the past) and a subrating of 92.000% (based on frequency of occurrances in ranked cases). This course is not available because its pre-requisites have not been fulfilled.

The course O2 has a rating of 87.500% (matched with cases in the past) and a subrating of 30.667% (based on frequency of occurrances in ranked cases). This is recommended as the choice # 2.

The course NM5 has a rating of 81.250% (matched with cases in the past) and a subrating of 44.000% (based on frequency of occurrances in ranked cases). This course is not available because its pre-requisites have not been fulfilled.

The course D1 has a rating of 78.125% (matched with cases in the past) and a subrating of 13.333% (based on frequency of occurrances in ranked cases). This is recommended as the choice # 3.

The course S8 has a rating of 75.000% (matched with cases in the past) and a subrating of 10.667% (based on frequency of occurrances in ranked cases).

### Trivial IR algorithm

The course N6 has a rating of 97.991% (matched with cases in the past) and a subrating of 54.667% (based on frequency of occurrances in ranked cases). This is recommended as the choice # 1.

The course OM3 has a rating of 97.991% (matched with cases in the past) and a subrating of 54.667% (based on frequency of occurrances in ranked cases). This course is not available because its pre-requisites have not been fulfilled.

The course NS7 has a rating of 97.991% (matched with cases in the past) and a subrating of 92.000% (based on frequency of occurrances in ranked cases). This course is not available because its pre-requisites have not been fulfilled.

The course O2 has a rating of 93.819% (matched with cases in the past) and a subrating of 30.667% (based on frequency of occurrances in ranked cases). This is recommended as the choice # 2.

**Original printed page 190 · source PDF page 199 · Open source page**

The course NM5 has a rating of 90.211% (matched with cases in the past) and a subrating of 48.000% (based on frequency of occurrances in ranked cases). This course is not available because its pre-requisites have not been fulfilled.

The course D1 has a rating of 88.454% (matched with cases in the past) and a subrating of 13.333% (based on frequency of occurrances in ranked cases). This is recommended as the choice # 3.

The course S8 has a rating of 87.500% (matched with cases in the past) and a subrating of 6.667% (based on frequency of occurrances in ranked cases).

## Closing source PDF page 200

**Original closing source page 200**

## Closing source PDF page 201

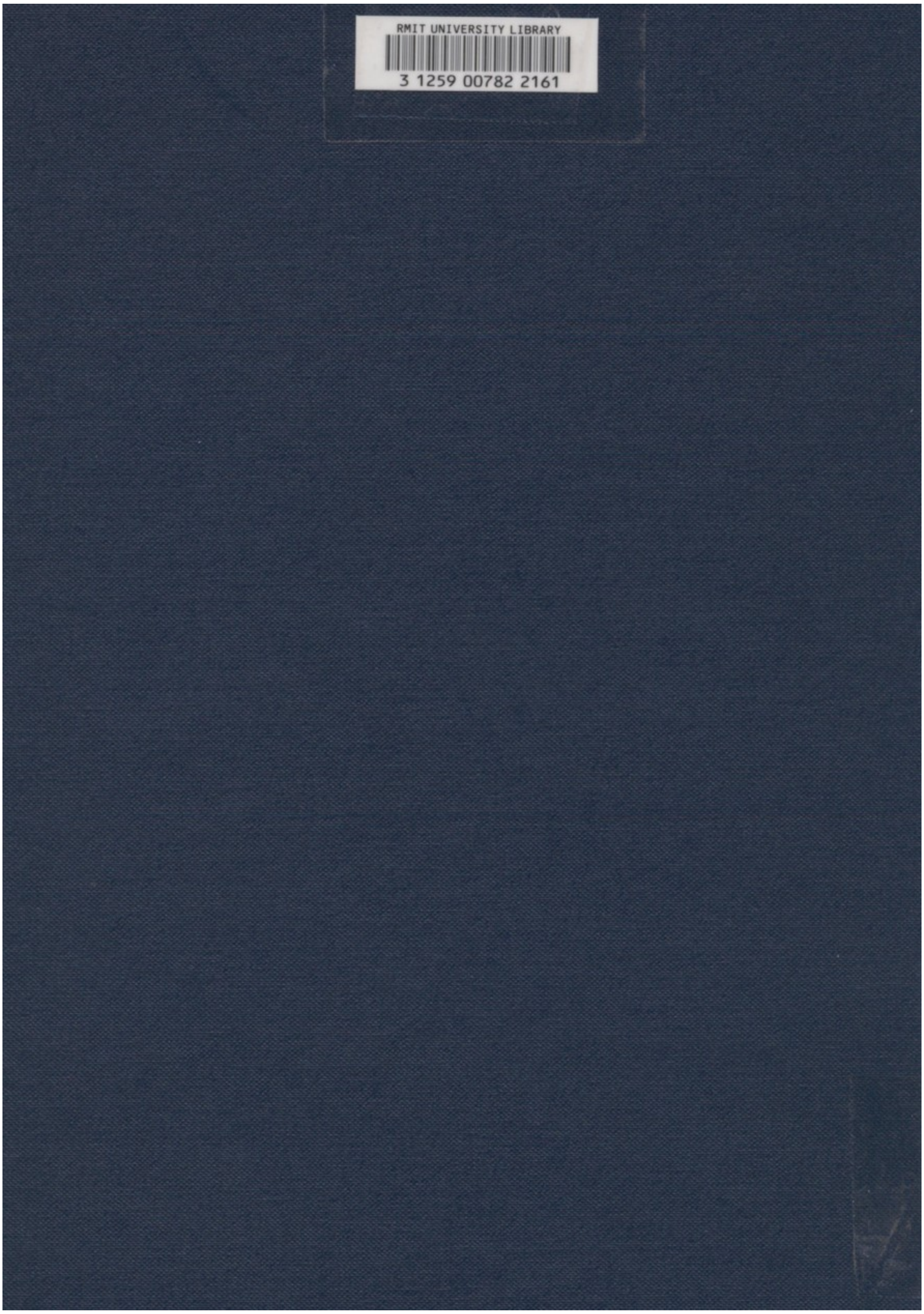

**Original closing source page 201**